\documentclass[11pt]{article}
\usepackage{amsmath,amsthm,amssymb,physics,mathtools}
\usepackage{courier}
\usepackage{hyperref}
\usepackage{todonotes}
\usepackage{datetime}
\usepackage{bbm}
\usepackage{algorithm}
\usepackage{algpseudocode}
\usepackage{natbib}
\usepackage[bottom]{footmisc}
\usepackage{xcolor}
\usepackage{enumitem}

\theoremstyle{plain}

\newtheorem*{theorem*}{Theorem}

\newtheorem*{definition*}{Definition}

\newtheorem*{assumption*}{Assumption}

\definecolor{customLink}{HTML}{0077B6}

\setcitestyle{authoryear,round}

\title{Introducing {\cogent{COGENT$^{\textbf{3}}$}} \\ An AI Architecture for Emergent Cognition}
\author{\vspace{-1em} Eduardo Salazar \\ \\ Nebula Technology Lab \\ \color{customLink}{\small{\href{mailto:cogent.three@gmail.com}{cogent.three@gmail.com}}}}
\date{}

\begin{document}

\newcommand{\SubItem}[1]{
    {\setlength\itemindent{15pt} \item[-] #1}
}

\newenvironment{cogent}{\fontfamily{lmss}\selectfont}{\par}

\maketitle

\begin{center}
    \normalsize{Rev. 2 -- January 24, 2026}
\end{center}
\vspace{0.5em}


\begin{abstract}
This paper presents {\cogent{COGENT$^{\textbf{3}}$}} (or \underline{\cogent{Co}}llective \underline{\cogent{G}}rowth and \underline{\cogent{Ent}}ropy-modulated 
\underline{\cogent{Triads}} System), a novel approach for emergent cognition integrating pattern formation networks with group influence dynamics. Contrasting with 
traditional strategies that rely on predetermined architectures, computational structures emerge dynamically in our framework through agent interactions. This enables 
a more flexible and adaptive system exhibiting characteristics reminiscent of human cognitive processes. The incorporation of temperature modulation and memory effects 
in {\cogent{COGENT$^{\textbf{3}}$}} closely integrates statistical mechanics, machine learning, and cognitive science.
\end{abstract}

\vspace{0.5em}
\noindent \normalsize{\textbf{AMS Classification:} \texttt{68T05}, \texttt{37A60}, \texttt{05C82}, \texttt{82C20}.}


\section{Introduction}  
\label{sec:introduction}

Understanding how cognitive capabilities emerge remains a central challenge in artificial intelligence (AI), cognitive science, and computational neuroscience. 
Historically, research on cognition has been sharply divided between symbolic approaches which rely on stable, rule-based representations as in \cite{newell1972human} 
or \cite{pylyshyn1984computation}, and connectionist (or neural network) models which emphasize distributed, adaptive processing structures, as in 
\cite{rumelhart1986parallel}. While both traditions have yielded valuable insights, neither fully captures the complex interplay between relatively enduring cognitive 
architectures and the dynamic, context-dependent processes that induce their usage, e.g., in \cite{chemero2009radical} or \cite{clark2013whatever}.

Advances in statistical physics, dynamical systems theory, and machine learning suggest a promising path for reconciling these perspectives, framing cognition 
as a combination of dynamic pattern formation, the interplay of multiple interacting elements across different scales, and the contextual modulation of stable 
structures by variable parameters.\footnote{The work by \cite{kelso1995dynamic} and \cite{thelen1994dynamic}, among others, is particularly relevant. See also 
\cite{buzsaki2010neural}.} This notion aligns with mounting evidence that human cognitive processes are neither exclusively symbolic nor purely connectionist. 

Cognition is therefore seen as emerging from continuous interactions among stable representational substrates (e.g., linguistic rules or mathematical operations), 
flexible functional groupings (such as neural assemblies that form and dissolve on-demand), environmental factors, and internal states that adaptively modulate their 
access and expression \citep{friston2010free}. This perspective not only aligns with early theories of script-based understanding and dynamic 
memory \citep{schank1983dynamic, schank2020scripts} but reflects attempts to computationally model cognition in artificial systems, also explored by 
\cite{schank1984cognitive} or more recently by \cite{lake2017machines} where language and learning mechanisms are integrated within a cognitive computational 
framework.


\subsection{Motivation and Challenges}  
\label{subsec:paperMotivation}

Three observations motivate our approach. Firstly, while traditional computational models often rely on fixed processing units, be they symbol-manipulating 
modules or static neural architectures, biological cognition demonstrates remarkable flexibility. Evidence from neurodynamics shows that human brains form transient 
functional assemblies that adapt fluidly to ongoing tasks, as discussed, e.g., by \cite{buzsaki2006rythms} and \cite{thivierge2008neural}. Although everyone shares similar stable cognitive 
building blocks\footnote{Basic sensory-motor patterns, linguistic rules or mathematical concepts, what could be viewed as the substrate or toolkit every human being
has, however basic they may be.} the actual formation, coordination, and dissolution of functional groups depends on context, skill, training, and even transient mental states. 

This helps explain differences among individuals. While most can run, and some can do it quite fast, there has been one Usain Bolt; while many learn algebra and physics, Hilbert or Einstein were unique; 
and although many of us can write, only a few, like Hemingway or Orwell, gained acclaim and recognition.

Such outcomes, although extraordinary, arise not from distinct fundamental tools but from their dynamic, context-sensitive application. What separates an
average runner from Usain Bolt doesn't stem from the running mechanism itself (basic biomechanics are, actually, the same across individuals) but rather from the dynamic optimization,
continuous feedback loops with the environment (training regimes, competition, personal physiology) and the adaptive fine-tuning of what is essentially a stable
pattern (the general human gait). Likewise, Einstein's symbolic reasoning didn't rely on radically different algebraic symbols than everyone else had been taught, but 
emerged from how he combined physical insight, spatial reasoning, plus (it must be said) daring conceptual leaps, to reconfigure these stable components in novel ways.

Secondly, cognition operates across multiple timescales and modes: from rapid, intuitive processes that act almost reflexively, to slower, more deliberative 
reasoning; see \cite{cisek2007cortical} and the landmark work by \cite{kahneman2011thinking}. Even stable symbolic knowledge, such as algebraic rules, can be deployed 
differently under pressure, in imaginative problem-solving, or in creative tasks that push the boundaries of known structures. This temporal layering suggests that 
cognition isn't reducible to static representations alone; the same stable tools can yield remarkably different outcomes depending on when and how they are engaged, 
and the emergent patterns of interaction that unfold during problem-solving or creative exploration.

In other words, small contextual differences\footnote{Like exposure to certain experiences, changes in arousal or focus, or subtle variations in perspective.} might lead to 
disproportionately large differences in how the human toolkit gets deployed. This is akin to a complex system in physics: the underlying rules (like the basic 
equations of motion) don't change, but slight differences in initial conditions can lead to vastly different trajectories. The cognitive landscape likewise might 
remain anchored by stable elements, yet unfold in highly individual ways.

Finally, the cognitive system appears sensitive to parameters that modulate its operational regime, akin to temperature effects in physical systems. This was noted 
long ago, e.g., by \cite{hopfield1982neural} or \cite{ackley1985learning}. Small shifts in factors such as stress, arousal, or cognitive load can produce substantial 
changes in how stable structures are utilized, reorganized, or sidelined. It all hints at a deeper principle of cognitive flexibility: stable rules and representations 
exist, but their effective expression is contingent on context-dependent, dynamically tuned parameters \citep{kirkpatrick1983optimization}. 


\subsection{An Integrated Perspective}
\label{subsec:towardsIntegrated}

To capture this interplay,\footnote{That is, the continuity of stable representations and the fluidity of dynamic assemblies observed in biological cognition.} we 
view cognitive architectures as \textbf{dynamic pattern formation networks} operating over a substrate of \textbf{stable representational elements}. This principle is 
at the core of our framework, inspired by concepts from statistical physics, where emergent global patterns arise from local interactions yet remain sensitive to 
parameters that can shift the system between different regimes of activity \citep[e.g.,][]{haken1983synergetics, ashby1962principles}. In the cognitive domain, this 
translates to \textbf{stable building blocks} (such as symbolic rules, learned motor programs, and lexical items) whose usage and combination are neither rigidly fixed 
nor entirely unconstrained, but rather \textbf{guided by dynamically changing conditions}.

Contemporary machine learning and neural network research can incorporate these insights by blending stable symbolic structures\footnote{For an example, 
see \cite{garcez2019neurosymbolic} on neuro-symbolic AI.} with flexible, context-sensitive pattern formation mechanisms. In other words, recognizing that 
\textbf{cognition is not just a matter of stable representations or dynamic interactions taken alone}. By fusing both elements, {\cogent{COGENT$^{\textbf{3}}$}} 
represents an artificial computational system that aims to approximate the nuanced interplay underlying human intelligence.


\subsection{Theoretical Framework}
\label{subsec:theoryFramework}

To do so, our framework integrates \textbf{three powerful paradigms} in a way that was never attempted before.
\begin{itemize}
    \item \textbf{Pattern Formation Networks} which extend and generalize the concepts presented by \cite{goodfellow2014generative} when formalizing adversarial 
    networks.\footnote{Building on the learning strategy proposed by \cite{Schmidhuber1991one, Schmidhuber1991two} using the notion of artificial curiosity.}
    \item \textbf{Collective influence models} from statistical physics, developed by \cite{castellano2009statistical} when presenting the q-voter model as a 
    generalization of the well-studied voter model of \cite{clifford1973sudbury}.
    \item \textbf{Memory kernels} as discussed in \cite{zwanzig2001nonequilibrium} in the context of non-equilibrium dynamics.
\end{itemize}

\noindent Pattern formation provides a mechanism for generative learning through evaluation interaction, while collective influence models capture the essence of group 
opinion dynamics. Memory kernels allow us to incorporate history-dependent effects crucial for cognitive processing.

The key innovation in our approach is introducing \textbf{triads} (a group of three agents) where \textbf{pattern formation emerges naturally from local 
interactions}.\footnote{The relevance of triads in graphs is well known; see, e.g., \cite{felmlee2021dyads}, \cite{peixoto2002disentangling} or \cite{cirigliano2024triad}.} 
Unlike traditional architectures with fixed mechanisms, triads \textbf{emerge and dissolve} based on context. Consequently, triad groupings follow a \textbf{hypergraph} 
structure in which roles are \textbf{dynamically assigned}.

Temperature modulation plays a central role, controlling both stochasticity and coupling strength between agents. This provides a natural mechanism for transitioning 
between different processing regimes, analogous to the \textbf{trade-off between exploration and exploitation} in human decision-making.\footnote{Exploitation involves 
selecting the currently presumed best option based on existing (although potentially incomplete, or biased) information. Exploration, on the other hand, is probing 
new alternatives that might produce superior outcomes later, even though doing so might forgo the immediate advantage of exploiting known options. See, e.g., 
\cite{cohen2007should}.}


\subsection{Paper Organization}
\label{subsec:paperOrganization}

The remainder of this paper is structured as follows. 
\begin{itemize}
    \item Section \ref{sec:architecture} (Architectural Framework) outlines the structural foundations of the proposed system, integrating key mathematical elements 
    such as hypergraph topology, agent properties, role dynamics, and hierarchical organization. It bridges local agent interactions with emergent global phenomena, 
    proposing a comprehensive model for multi-scale emergent behavior.
    \item Section \ref{sec:core} (Core Mathematical Framework) develops the mathematical backbone of our system, integrating state spaces, operators, temperature 
    dynamics, memory effects, and system Hamiltonian.
    \item Section \ref{sec:formation} (Formation Dynamics) defines our approach for modeling emergent pattern formation, focusing on the mechanisms by which local 
    interactions, memory effects, and temperature-dependent dynamics can lead to global computational structures and the formation of coherent, collective states.
    \item Section \ref{sec:dynamics} (Dynamic Processes) examines the temporal evolution of the system, looking at aspects such as the interaction between agents over 
    time, mechanisms for system-wide convergence and emergent cognitive states, and the role played by stochasticity, memory, and temperature in driving such dynamics. 
    \item Section \ref{sec:phase} (Phase Space Analysis) and Section \ref{sec:critical} (Critical Phenomena) investigate the conceptual and mathematical foundation 
    of how our system’s computational states are organized, evolve, and transition. In Section \ref{sec:phase} the focus centers on energy landscapes and the dynamics 
    in phase space. In turn, Section \ref{sec:critical} explores how the system undergoes transitions between different computational phases, characterized by symmetry 
    breaking and critical behaviors. Together, these sections provide a rigorous framework to support complex, memory-dependent, and symmetry-driven computational 
    systems, such as the one proposed in {\cogent{COGENT$^{\textbf{3}}$}}.
    \item Section \ref{sec:collective} (Collective States) investigates the emergent behaviors and computational capabilities of the system as it transitions from 
    localized individual processes to large-scale coordinated dynamics. It delves into the structural, informational, and dynamical aspects of collective states, 
    emphasizing their computational and cognitive relevance.
    \item Finally, Section \ref{sec:stability} (Stability Analysis) examines the conditions under which the system maintains reliable computational behavior in the 
    face of internal dynamics, memory effects, environmental perturbations, and structural uncertainties.
\end{itemize}

To summarize, this paper opens new directions for developing \textbf{artificial systems that better capture the flexibility and context-sensitivity of human cognition}. 
Our framework offers a principled way to bridge pattern formation and thermodynamic concepts, while maintaining mathematical rigor throughout.


\subsection{Related Work}
\label{subsec:relatedWork}

Our work builds on several lines of research. Sociophysics models, for example, have demonstrated the value of statistical physics approaches in understanding collective 
behavior; see \cite{galam1996fragmentation, galam2002stability, galam2008sociophysics} or \cite{salazar2002Coalitions} as examples. The importance of transient dynamics in neural 
computation has been established in various frameworks, as in \cite{maass2002real}, though rarely integrated with pattern formation models.

The role of temperature-like parameters in neural computation also has a rich history, from Hopfield networks to Boltzmann machines. That said, the integration of such 
temperature modulation with modern pattern formation architectures and memory effects, as pursued in this paper, is a novel direction in cognitive modeling. The aim is 
to provide a coherent mathematical framework for understanding such temperature-dependent transitions.


\section{Notation and Conventions}
\label{sec:notation}
We model {\cogent{COGENT$^{\textbf{3}}$}} as a classical stochastic dynamical system on a \emph{dynamic triadic hypergraph}. All spatial operators 
(gradients/Laplacians) acting on node- or triad-indexed fields are interpreted in the \emph{graph and hypergraph-discrete} sense. When we introduce a 
continuous coordinate sector $Y$ of $\Gamma$ (as in Section \ref{subsec:coupling}) we use standard Euclidean operators $\nabla_Y$ and $\nabla_Y\cdot$ on that sector.

\begin{itemize}
    \item \textbf{Feature/knowledge space}
    \begin{itemize}
        \item $\Theta\subset\mathbb{R}^d$ is a compact feature space.
        \item $\mathcal{K}\coloneqq\mathcal{P}_2(\Theta)$ is the space of Borel probability measures on $\Theta$ with finite second moment, endowed with the 
        Wasserstein-2 metric
        $W_2^2(\mu,\nu)\coloneqq \inf_{\pi\in\Pi(\mu,\nu)}\int_{\Theta\times\Theta}\|x-y\|^2\,d\pi(x,y)$
        \item Initial ignorance is uniform: $\mathcal{K}_i(0)=\mathrm{Unif}(\Theta)$ for all agents $i$.
    \end{itemize}

    \item \textbf{Hypergraph structure}
    \begin{itemize}
        \item $V$ is the set of agents, $|V|=N$.
        \item $E^{(3)}(t)\subset\binom{V}{3}$ is the set of active triads at time $t$ (dynamic).
        \item The induced (2-section) graph is $E^{(2)}(t)\coloneqq \big\{\{i,j\}\in\tbinom{V}{2}:\exists\,\{i,j,k\}\in E^{(3)}(t)\big\}$.
    \end{itemize}

    \item \textbf{Agent variables and roles}
    \begin{itemize}
        \item Opinion state: $\mathbf{s}_i(t)\in\mathcal{S}\coloneqq\{-1,1\}^m$.
        \item Temperature: $T_i(t)\in\mathbb{R}_+$.
        \item Roles are \emph{triad-incidence variables}: for each active triad $\tau\in E^{(3)}(t)$ and each $i\in \tau$, $r(i,\tau)\in\{G_1,G_2,D\}$.
        \item Incidence set at time $t$ defined by 
        $\mathcal{I}(t)\coloneqq\{(i,\tau):\ \tau\in E^{(3)}(t),\ i\in\tau\}$.
        \item Triad temperature (used for triad-local stochastic learning terms) as 
        $T_\tau(t)\coloneqq \frac{1}{3}\sum_{i\in\tau}T_i(t)$.
        \item Node formation field: $\phi_i(t)\in\mathbb{R}$,
        \item Local agent state (excluding roles, which live on triads)
        $x_i(t)\coloneqq(\mathcal{K}_i(t),\mathbf{s}_i(t),T_i(t),\phi_i(t))\in \mathcal{P}_2(\Theta)\times\{-1,1\}^m\times\mathbb{R}_+\times\mathbb{R}$.
    \end{itemize}

    \item \textbf{Formation networks and parameter coordinates (per triad)}
    \begin{itemize}
        \item For each active triad $\tau$ and $m\in\{1,2\}$, $G_\tau^{(m)}(\cdot;\theta_{G,\tau}^{(m)})$ is a generator network and $D_\tau(\cdot;\theta_{D,\tau})$ 
        is a discriminator.
        \item When norms or inner-products are applied to formation variables, they are taken in \emph{parameter space} after vectorization, that is
        $\mathbf{G}_\tau^{(m)}\coloneqq \mathrm{vec}(\theta_{G,\tau}^{(m)})$ and  $\mathbf{D}_\tau\coloneqq \mathrm{vec}(\theta_{D,\tau})$.
    \end{itemize}

    \item \textbf{Configuration space and observables}
        \begin{itemize}
        \item Global agent state (random variable): $X(t)\coloneqq (x_i(t))_{i\in V}\in \prod_{i\in V}\Omega_i$.
        \item Global agent state space: $\Omega\coloneqq \prod_{i\in V}\Omega_i$.
        \item Structural state: $\mathfrak{g}(t)\coloneqq(E^{(3)}(t),r(t))$.
        \item Full configuration: $\Gamma(t)\coloneqq(X(t),\mathfrak{g}(t),M(t))$ where $M(t)$ denotes memory variables.
        \item Observables: $\mathcal{O}\coloneqq L^2(\Gamma,\mu)$ for a reference (e.g., stationary) measure $\mu$ when it exists.
    \end{itemize}

    \item \textbf{Discrete calculus conventions}
    \begin{itemize}
        \item On the induced graph $G^{(2)}(t)=(V,E^{(2)}(t))$ with symmetric weights $w_{ij}(t)\ge 0$, the (weighted) graph Laplacian acting on node 
        fields $f:V\to\mathbb{R}$ is $(Lf)_i\coloneqq \sum_{j:\{i,j\}\in E^{(2)}(t)} w_{ij}(t)\big(f_i-f_j\big)$. When $\nabla^2 f$ or $\Delta f$ 
        appears for node-indexed fields it denotes $Lf$.
        \item Discrete integrals satisfy $\int_V g(\mathbf{r})\,d\mathbf{r}\equiv \sum_{i\in V} g_i$.
    \end{itemize}

    \item \textbf{Units and constants}\\[0.5em]
        \noindent We keep $k_B$ explicit when convenient. When $k_B$ is omitted, $T$ is understood in energy units (e.g., $k_BT$ has been absorbed into $T$).

    \item \textbf{Dynamics operator (classical Markov semigroup)}
    \begin{itemize}
        \item The time evolution is a continuous-time Markov process with generator $\mathcal{L}$ acting on observables and adjoint $\mathcal{L}^{\ast}$ acting 
        on densities, hence $\partial_t \rho_t = \mathcal{L}^{\ast}\rho_t$, with $P_t = e^{t\mathcal{L}}$.
        \item We will keep the symbol $\hat{\mathfrak{L}}$ to denote this generator, e.g., $\hat{\mathfrak{L}}\equiv\mathcal{L}$.
    \end{itemize}
\end{itemize}


\section{Architectural Framework}
\label{sec:architecture}

The mathematical foundation of {\cogent{COGENT$^{\textbf{3}}$}} rests on the integration of three key elements: \textbf{pattern formation dynamics}, \textbf{group influence processes} 
and \textbf{memory effects}. As noted above, this requires careful consideration of the underlying mathematical structures to ensure consistency across 
different scales and types of dynamics.

That would not be too relevant, however, if we were unable to provide some of the \textbf{cognitive intuition} behind our model. The simplest way is to consider how 
biological neural systems dynamically reorganize, based on task demands. As discussed earlier, just as neural assemblies form and dissolve adaptively our framework 
allows computational structures to emerge through \textbf{dynamic role assignments and interactions}. The hypergraph structure (where each edge connects nodes assembled 
in triads) delivers the \textbf{minimal topology needed} to capture both \textbf{local pattern formation} and \textbf{global information integration} processes, 
mirroring how biological systems simultaneously generate and assess computational patterns.


\subsection{Network Structure}
\label{subsec:networkStructure}

The system is composed of $N$ agents forming a \textbf{dynamic hypergraph} $\mathcal{G} = (V, E^{(3)})$ where
    \begin{itemize}
        \item $V$ is the set of agents; and
        \item $E^{(3)}(t)\subset \dbinom{V}{3}$ represents \textbf{unordered triads} (3-uniform hyperedges).
    \end{itemize}

\noindent Given the dynamic role assignment between formation agents and evaluator, each interaction potentially leads to a \textbf{pattern formation structure} when 
required. 

The choice of a hypergraph structure is motivated by both theoretical and practical considerations. While traditional neural architectures rely on pairwise interactions, 
cognitive systems often exhibit higher-order dependencies that cannot be reduced to simple dyadic relationships. Hence our use of triad groupings.

As also mentioned earlier, the other key innovation where we also notably depart from traditional architectures is allowing for \textbf{dynamic role assignment}. 

Putting it all together, the proposed system is characterized by the \textbf{topological measures} outlined below.
    \begin{itemize}
        \item \textbf{Local Density}\\[0.5em]
        \begin{equation}
            \rho_i(t) \coloneqq \frac{|E_i^{(3)}(t)|}{\binom{|V|-1}{2}} \quad \text{where} \quad E_i^{(3)}(t)\coloneqq\{\tau\in E^{(3)}(t): i\in\tau\}
        \end{equation}

        \item \textbf{Clustering coefficient}\\[0.5em]
        \begin{equation}
        C_i(t) \coloneqq \frac{\big|\{(\tau,\tau'):\ \tau,\tau'\in E_i^{(3)}(t),\ \tau\neq\tau',\ |\tau\cap\tau'|=2\}\big|}
        {\binom{|E_i^{(3)}(t)|}{2}}
        \end{equation}

        \item \textbf{Path length}\\[0.5em]
        \begin{equation}
            L_{ij}(t)\coloneqq \mathrm{dist}_{G^{(2)}(t)}(i,j) \quad \text{with} \quad G^{(2)}(t)=(V,E^{(2)}(t))
        \end{equation}      
    \end{itemize}


\subsection{Agent Properties}
\label{subsec:agentProperties}

Each agent $i\in V$ maintains a local state
    \begin{equation}
        x_i(t)=(\mathcal{K}_i(t),\mathbf{s}_i(t),T_i(t),\phi_i(t))\in \mathcal{P}_2(\Theta)\times\{-1,1\}^m\times\mathbb{R}_+\times\mathbb{R}
    \end{equation}

Roles are not stored as a per-agent Cartesian factor, because roles are defined \emph{per triad-incidence}. Instead, for each active triad $\tau\in E^{(3)}(t)$ and 
each $i\in \tau$, the role map assigns
\begin{equation}
    r(i,\tau)\in\{G_1,G_2,D\}
\end{equation}

\noindent At $t=0$ all agents are uniformly ignorant, hence $\mathcal{K}_i(0)=\mathrm{Unif}(\Theta)$.


\subsection{Role Assignment Dynamics}
\label{subsec:assignDyn}

Role updates are defined on \textbf{triad incidences}. For an active triad $\tau=\{i,j,k\}\in E^{(3)}(t)$ and an agent $i\in\tau$, the incidence-role variable is 
$r(i,\tau)\in\{G_1,G_2,D\}$. The transition probability for updating the role on the incidence $(i,\tau)$ is taken to be temperature-dependent and of heat-bath form
    \begin{equation}
        P\!\big(r(i,\tau)\to r'(i,\tau)\,\big|\,\Gamma_t\big)
        =\frac{\exp\!\big(-\Delta E_{i,\tau}(r')/(k_B T_i(t))\big)}
        {\sum_{\tilde r\in\{G_1,G_2,D\}}\exp\!\big(-\Delta E_{i,\tau}(\tilde r)/(k_B T_i(t))\big)}
    \end{equation}

\noindent Here, $\Delta E_{i,\tau}(r')$ is the energy cost of assigning role $r'$ to incidence $(i,\tau)$, defined by
    \begin{equation}
        \begin{split}
            \Delta E_{i,\tau}(r') \coloneqq&\; E_{local}(i,\tau;r') \;+\; \gamma\sum_{j\in\tau\setminus\{i\}} J_{ij}\big(r',\,r(j,\tau)\big)\\
            +&\; \lambda\int_0^t K(t-s)\,M_{r'}(s)\,ds
        \end{split}
    \end{equation}

The local term $E_{local}(i,\tau;r')$ quantifies compatibility of agent $i$ with the other two agents in $\tau$ under the proposed role. 
Let $(j,k)=\tau\setminus\{i\}$ and define the two-point Wasserstein barycenter
    \begin{equation}
        \mathrm{Bar}(\mathcal{K}_j,\mathcal{K}_k)\in\arg\min_{\mu\in\mathcal{P}_2(\Theta)}\Big(W_2^2(\mu,\mathcal{K}_j)+W_2^2(\mu,\mathcal{K}_k)\Big)
    \end{equation}

Define also a discrete \emph{consensus} opinion (componentwise majority) as \linebreak $\mathrm{Med}(\mathbf{s}_j,\mathbf{s}_k)\in\{-1,1\}^m$. Then
    \begin{equation}
        E_{local}(i,\tau;r') \coloneqq \alpha\, W_2^2\!\Big(\mathcal{K}_i,\mathrm{Bar}(\mathcal{K}_j,\mathcal{K}_k)\Big)
        +\beta\, d_H\!\Big(\mathbf{s}_i,\mathrm{Med}(\mathbf{s}_j,\mathbf{s}_k)\Big)
    \end{equation}

The memory signal entering the role update is defined from the conserved node-memory field $(M_i(t))_{i\in V}$ by incidence-averaging over the currently 
active incidence set, that is
    \begin{equation}
        \begin{gathered}
            M_{r}(t)\coloneqq \frac{1}{|\mathcal{I}_{r}(t)|}\sum_{(a,\sigma)\in\mathcal{I}_{r}(t)} M_a(t)\\
            \text{with } \mathcal{I}_{r}(t)\coloneqq\{(a,\sigma):\ \sigma\in E^{(3)}(t),\ a\in\sigma,\ r(a,\sigma)=r\}
        \end{gathered}
    \end{equation}
\noindent ensuring that role assignment is defined consistently as a triad-incidence process while remaining compatible with the conserved total memory 
$Q_3=\sum_{i\in V}M_i$.


\subsection{Cognitive Interpretation}
\label{subsec:cognitiveInterp}

{\cogent{COGENT$^{\textbf{3}}$}} components map onto cognitive processes, but the mapping is structural rather than merely analogical.

Its formulation as a Markov process reflects the inherently stochastic nature of neural computation. Transitions between states are probabilistic rather than 
deterministic. The graph Laplacian governs how information diffuses through discrete connectivity, matching how signals propagate through actual neural architectures 
rather than continuous fields.

Conservation, in turn, mirrors balanced excitation and inhibition, as one region's activation gain is another's loss, preserving total activity. Temperature modulation 
provides a mechanism for balancing exploration and exploitation, while the Markovian embedding of memory kernels offers a model of working memory based on finite state, 
not stored trajectories.

This mapping reflects shared constraints: any system processing information through discrete, stochastic, resource-limited substrates will face the same mathematical 
requirements.


\subsection{State Space Topology}
\label{subsec:ssTopology}

Drawing from methods of statistical physics and dynamical systems, we endow the state space with the product topology induced by the agent-level factors. The local 
agent state space is
    \begin{equation*}
        \Omega_i \coloneqq \mathcal{P}_2(\Theta)\times\{-1,1\}^m\times\mathbb{R}_+\times\mathbb{R}
    \end{equation*}

\noindent so that the global agent state space is $\Omega=\prod_{i\in V}\Omega_i$. Roles are \emph{not} included as a per-agent factor because roles are triad-incidence 
variables. Instead, they are encoded by the incidence map $r(t)$ on the active triad set $E^{(3)}(t)$.

Accordingly, the full configuration at time $t$ is
    \begin{equation}
    \Gamma(t)\coloneqq\big(X(t),\mathfrak{g}(t),M(t)\big) \quad \text{with} \quad \mathfrak{g}(t)\coloneqq\big(E^{(3)}(t),r(t)\big)
    \end{equation}

\noindent where $M(t)$ denotes the memory variables.

For the knowledge component, $\mathcal{P}_2(\Theta)$ carries the Wasserstein metric topology induced by $W_2$ as defined in Section \ref{sec:notation}. The opinion space 
$\{-1,1\}^m$ carries the discrete topology; and $\mathbb{R}_+$ carries the standard Euclidean topology.


\subsection{Structural Organization}
\label{subsec:structOrg}

{\cogent{COGENT$^{\textbf{3}}$}} operates across \textbf{three primary scales}, inspired by the hierarchical models of cognitive modeling in \cite{friston2005cortical}, 
\cite{lee2013bayesian}, and the applied research in \cite{diuk2013hierarchical} and others.

\subsubsection{Microscopic Level}
\textbf{Individual agents} with the properties outlined below.
\begin{itemize}
    \setlength\itemsep{0.8em}
    \item Local state spaces: $\Omega_i$.
    \item Neighborhood structure: $\mathcal{N}_i(t)\coloneqq \{j\in V:\exists\,\tau\in E^{(3)}(t),\ \{i,j\}\subset\tau\}$.
    \item Local triadic connectivity: $\kappa_i(t)\coloneqq |\{\tau\in E^{(3)}(t): i\in\tau\}|$.
\end{itemize}

\subsubsection{Mesoscopic Level}
\textbf{Triad structures} characterized by the following properties.
\begin{itemize}
    \setlength\itemsep{0.8em}
    \item Role distribution on a triad: $\tau=\{i,j,k\}$: $\mathcal{R}_\tau(t)\coloneqq \big(r(i,\tau),r(j,\tau),r(k,\tau)\big)$.
    \item Local triad-neighborhood topology: $\mathcal{T}_\tau(t)\coloneqq \{\tau'\in E^{(3)}(t): \tau\cap\tau'\neq\emptyset\}$.
    \item Triad interaction strength (model-dependent): $J_\tau(t)\coloneqq J\big(\mathcal{K}_i(t),\mathcal{K}_j(t),\mathcal{K}_k(t)\big)$\\
    for $\tau=\{i,j,k\}$.
\end{itemize}

\subsubsection{Macroscopic Level}
\textbf{Global network} properties, as defined below.
\begin{itemize}
    \setlength\itemsep{0.8em}
    \item Network density: $\rho(t)\coloneqq |E^{(3)}(t)|\,/\,\binom{|V|}{3}$.
    \item Global clustering: $C(t)\coloneqq \frac{1}{|V|}\sum_{i\in V} C_i(t)$.
    \item Average path length: $\overline{d}(t)\coloneqq \frac{1}{|V|(|V|-1)}\sum_{i\neq j} L_{ij}(t)$.
\end{itemize}

\newpage

\noindent Among the various metrics characterizing the system structure, we will focus on those most relevant to cognitive processing, which we enumerate below.
\begin{itemize}
    \item The clustering coefficient $C_i$ capturing \textbf{local processing coherence}.
    \item Order parameters $\Psi_\alpha$ measuring the \textbf{emergence of collective computational states}.
    \item Hierarchical entropy $S_{hier}$ quantifying \textbf{genuine emergence beyond simple aggregation}.
\end{itemize}

\noindent This multi-scale organization enables the \textbf{emergence of collective computational states} while maintaining \textbf{local processing capabilities}, 
a feature deemed crucial for cognitive systems; see, e.g., \cite{buzsaki2010neural}.


\subsection{Local-Global Network Relations}
\label{subsec:localGlobal}

The relationship between local and global network properties is characterized through several key mappings, following the principles from statistical physics outlined 
by \cite{castellano2009statistical} and others.

\subsection{State and Configuration Spaces}
\label{subsec:confSpaces}

The state space $\Omega=\prod_{i\in V}\Omega_i$ collects all agent variables (knowledge, opinions, temperatures, and node formation field) while the configuration 
space augments $\Omega$ by the structural degrees of freedom (active triads and their incidence-role assignments) and memory variables. Hence,
    \begin{equation}
        \Gamma(t)=\big(X(t),\mathfrak{g}(t),M(t)\big) \quad \text{with} \quad \mathfrak{g}(t)=\big(E^{(3)}(t),r(t)\big)
    \end{equation}

\noindent This separation matches the model convention that roles live on triad incidences rather than as per-agent coordinates.

\subsubsection{Scale Bridging Functions}
\label{subsec:scaleBridge}

Define the triad-local agent state space for $\tau=\{i,j,k\}$ by
    \begin{equation*}
        \Omega_\tau \coloneqq \Omega_i\times\Omega_j\times\Omega_k
    \end{equation*}

\noindent We use scale transition operators that map the global agent state to local, mesoscopic, and macroscopic representations defined by
    \begin{align*}
        \pi_{local}: \Omega(t) &\to \bigoplus_{i\in V} \Omega_i && \text{(Local projection)}\\
        \pi_{meso}: \Omega(t) &\to \bigoplus_{\tau\in E^{(3)}(t)} \Omega_\tau && \text{(Mesoscopic projection)}\\
        \pi_{global}: \Omega(t) &\to \Omega(t)/{\sim} && \text{(Global quotient)}
    \end{align*}

\noindent Here $\sim$ identifies microstates sharing the same chosen macroscopic summary statistics. For any observable $\mathcal{O}$, a consistent cross-scale 
representation is
    \begin{equation*}
        \mathcal{O}_{macro} = \pi_{global}\circ\pi_{meso}\circ\pi_{local}[\mathcal{O}]
    \end{equation*}

\subsubsection{Hierarchical Consistency Relations}
The consistency between scales is ensured by noting the following relationships. 
\begin{enumerate}
    \item \textbf{Local-Mesoscopic coupling}
        \begin{equation*}
            \kappa_{i,\tau}(t) \coloneqq
            \frac{\big|\{\tau' \in E^{(3)}(t):\ i\in\tau',\ \tau'\cap\tau \neq \emptyset\}\big|}
            {|E^{(3)}(t)|}
        \end{equation*}

    \item \textbf{Mesoscopic-Global bridging}
        \begin{equation*}
            \Phi_{\tau,g}(t) \coloneqq \frac{1}{|E^{(3)}(t)|}\sum_{\tau'\in E^{(3)}(t)} \mathbbm{1}[\tau' \text{ contributes to } g]
        \end{equation*}

    \item \textbf{Scale-dependent correlation functions}
        \begin{align*}
            C_{local}(i,j;t) &= \langle \mathbf{s}_i(t)\cdot\mathbf{s}_j(t)\rangle - \langle \mathbf{s}_i(t)\rangle\cdot\langle \mathbf{s}_j(t)\rangle \\[1em]
            C_{meso}(\tau,\tau';t) &= \langle \phi_{align,\tau}(t)\,\phi_{align,\tau'}(t)\rangle - \langle \phi_{align,\tau}(t)\rangle\langle \phi_{align,\tau'}(t)\rangle \\[0.8em]
            C_{global}(t) &= \frac{1}{|V|^2}\sum_{i,j\in V} C_{local}(i,j;t)
        \end{align*}
\end{enumerate}

\subsubsection{Emergence Characterization}
The emergence of \textbf{global properties} is represented by the metrics depicted below.
    \begin{enumerate}
        \item \textbf{Order parameters}
            \begin{equation*}
                \Psi_\alpha = \frac{1}{|V|}\sum_{i\in V} \psi_{\alpha,i}
            \end{equation*}
        \item \textbf{Fluctuation measures}
            \begin{equation*}
                \chi_\alpha = \frac{1}{|V|k_BT}\left(\langle \Psi_\alpha^2 \rangle - \langle \Psi_\alpha \rangle^2\right)
            \end{equation*}
        \item \textbf{Hierarchical entropy}
            \begin{equation*}
                S_{hier}(t) = S_{global}(t) - \frac{1}{|V|}\sum_{i\in V} S_{local,i}(t) - \frac{1}{|E^{(3)}(t)|}\sum_{\tau\in E^{(3)}(t)} S_{meso,\tau}(t)
            \end{equation*}
    \end{enumerate}

\noindent The quantities above \textbf{measure genuine emergence} beyond the simple aggregation of local properties.

\subsubsection{Topological Invariants}
The network structure preserves certain invariants across scales, providing robust measures of the network topology. These are
    \begin{itemize}
        \item the \textbf{Euler characteristic} defined by $\chi(G) = |V| - |E^{(2)}| + |E^{(3)}|$;
        \item \textbf{Betti numbers} $\beta_k = \dim H_k(G, \mathbb{Z})$, characterizing the k-dimensional holes in the hypergraph structure; and
        \item \textbf{spectral invariants} given by the eigenvalue spectra $\mathrm{spec}(\Delta_k)=\{\lambda_i(\Delta_k)\}$ of the $k$-dimensional (hyper)graph 
        Laplacians $\Delta_k$ (when defined for the chosen hypergraph model).
    \end{itemize}

\newpage

\subsection{Interaction Framework}
\label{subsec:interFramework}

The basic interaction structure is defined as follows
\begin{itemize}
    \item \textbf{Direct connections}: $\{i,j\}\in E^{(2)}(t)\iff \exists\,\tau\in E^{(3)}(t)\; \text{such that}\linebreak
    \{i,j\}\subset\tau$.
    \item \textbf{Triad interactions}: $\tau\in E^{(3)}(t)$ are the elementary higher-order interaction units.
    \item \textbf{Role assignments}: Roles live on incidences, for example $r(t):\mathcal{I}(t)\to\{G_1,G_2,D\}$ with 
    $\mathcal{I}(t)=\{(i,\tau):\tau\in E^{(3)}(t),\,i\in\tau\}$.
\end{itemize}

\noindent and induce a natural update-operator algebra acting on observables, that is
\begin{equation}
\mathfrak{A} \coloneqq \mathrm{span}\{\hat{O}_{\alpha}:\ \hat{O}_{\alpha}:\mathcal{O}\to\mathcal{O}\}
\end{equation}

\noindent where $\mathcal{O}$ is an observable space such as $L^2(\Gamma,\mu)$.

\subsection{Component Integration}
\label{subsec:compIntegration}

The architectural components interact through the mathematical spaces defined in Section \ref{sec:core}. The agent state space $\Omega=\prod_{i\in V}\Omega_i$ 
collects agent properties (knowledge, opinions, temperatures, and node formation field), while the configuration at time $t$, $\Gamma(t)$, augments the agent 
state $X(t)\in\Omega$ by the structural degrees of freedom $\mathfrak{g}(t)=(E^{(3)}(t),r(t))$ and memory variables $M(t)$.

The integration of components is realized through the projection operators defined below.
    \begin{align*}
        \pi_{\Omega}: \Gamma(t) &\to \Omega(t) && \text{(Agent-state projection)}\\
        \pi_{\mathfrak{g}}: \Gamma(t) &\to \mathfrak{g}(t) && \text{(Structural projection)}\\
        \pi_{M}: \Gamma(t) &\to M(t) && \text{(Memory projection)}
    \end{align*}

\noindent This architectural framework provides the structural foundation for the dynamic processes and emergent behaviors described in subsequent sections.


\section{Core Mathematical Framework}
\label{sec:core}

To illustrate these abstract concepts, it can be useful to consider a minimal system, which is simply a triad. The dynamics of this small-scale network provides 
a concrete (and complete) example of how {\cogent{COGENT$^{\textbf{3}}$}} operates. When temperature is high, agents frequently switch roles and explore different 
computational configurations. As temperature decreases, the system begins to settle into stable patterns, analogous to \textbf{how cognitive systems transition 
from exploration to exploitation}.

\subsection{Mathematical Spaces}
\label{subsec:mathSpaces}

Building on the spaces introduced in Section \ref{sec:architecture}, we now develop their full mathematical structure. The state space $\Omega$ and configuration space $\Gamma$ defined 
earlier are complemented by the observable space
    \begin{equation}
        \mathcal{O} = L^2(\Gamma,\mu)
    \end{equation}

\noindent which provides the framework for measuring and analyzing system properties.

The configuration space $\Gamma$ inherits a natural topology from its product structure, enabling both analytical treatment and physical interpretation. The 
observable space $\mathcal{O}$ completes the mathematical framework by providing the setting for operators defined in subsequent sections.


\subsection{Core Operator}
\label{subsec:coreOp}

The dynamics of {\cogent{COGENT$^{\textbf{3}}$}} are defined as a continuous-time Markov process on the configuration space $\Gamma$ (see Section \ref{sec:notation}). 
Its fundamental operator is the (backward) Markov generator $\hat{\mathfrak{L}}\equiv\mathcal{L}$ acting on observables $f$, with associated semigroup 
$P_t=e^{t\mathcal{L}}$ and forward equation for the law $\rho_t$ defined by
    \begin{equation}
        \partial_t \rho_t = \mathcal{L}^{\ast}\rho_t
    \end{equation}

\noindent All subsequent \emph{operator} statements in this paper (e.g., spectral decompositions, invariant measures, steady-state currents) refer to this Markov 
generator and its adjoint. No Poisson-bracket or commutator structure is assumed.


\subsection{Temperature Field Dynamics}
\label{subsec:tempFieldDyn}

The temperature field $(T_i(t))_{i\in V}$ plays a crucial role in mediating between different dynamical regimes. At high temperatures, the system explores its state 
space more freely, analogous to the exploratory phase of cognitive processing. As temperature decreases, the dynamics become more focused, leading to the exploitation 
of established computational pathways. This temperature dependence provides a natural mechanism for balancing exploration and exploitation, which, as noted 
earlier, is a key feature of adaptive cognitive systems.

The spatial variation of temperature across the {\cogent{COGENT$^{\textbf{3}}$}} network \textbf{allows different regions to operate in distinct computational regimes simultaneously}, enhancing the 
system's processing capabilities. The temperature field evolves following
    \begin{equation}
        \frac{dT_i}{dt} = -\kappa_T (LT)_i - \gamma(T_i-T_0) + \eta_i^{(T)}(t)
    \end{equation}

\noindent where $\kappa_T$ is the thermal diffusivity, $\gamma$ controls relaxation to equilibrium temperature $T_0$, and $\eta_i^{(T)}(t)$ 
represents local fluctuations with correlations
    \begin{equation*}
        \langle\eta_i^{(T)}(t)\eta_j^{(T)}(t')\rangle = 2D(T_i(t))\,\delta_{ij}\delta(t-t')
    \end{equation*}

\noindent We take $D(T)\coloneqq \gamma T$ (equilibrium-consistent choice).

This fundamental evolution of the temperature field describes its baseline dynamics in the absence of pattern formation. While $\kappa_T$ governs heat transport, 
the relaxation term $-\gamma(T-T_0)$ maintains thermal stability around equilibrium temperature $T_0$. The noise correlations ensure proper thermodynamic 
behavior at equilibrium.


\subsection{Explicit Lagrangian of the Pattern Formation Network}
\label{subsec:expLagrangian}

For each active triad $\tau\in E^{(3)}(t)$ we attach two generator networks\linebreak $G_\tau^{(m)}(\cdot;\theta_{G,\tau}^{(m)})$ for ($m=1,2$) and one discriminator 
$D_\tau(\cdot;\theta_{D,\tau})$. When norms and inner products are used, we work in parameter space after vectorization, that is
    \begin{equation*}
        \mathbf{G}_\tau^{(m)}\coloneqq \mathrm{vec}(\theta_{G,\tau}^{(m)})\quad \text{and} \quad \mathbf{D}_\tau\coloneqq \mathrm{vec}(\theta_{D,\tau})
    \end{equation*}

\noindent The triad-local objective (GAN-type) with regularization and memory defined as
    \begin{equation}
        \mathcal{J}_\tau(t)=\mathcal{J}_{adv,\tau}(t)+\mathcal{J}_{mem,\tau}(t)
    \end{equation}

\noindent In turn, the schematic temperature-modulated stochastic gradient flow (in parameter space) is given by
    \begin{align*}
        \frac{d\mathbf{G}_\tau^{(m)}}{dt} &= -\nabla_{\mathbf{G}_\tau^{(m)}}\mathcal{J}_\tau + \sigma_G\sqrt{2T_\tau(t)}\,\xi_{\tau,m}(t)\\[0.3em]
        \frac{d\mathbf{D}_\tau}{dt} &= +\nabla_{\mathbf{D}_\tau}\mathcal{J}_\tau + \sigma_D\sqrt{2T_\tau(t)}\,\zeta_{\tau}(t)
    \end{align*}

\noindent with independent standard white noises $\xi, \zeta$ (unit covariance) and $T_\tau(t)=\frac13\sum_{i\in\tau}T_i(t)$.


\subsection{Memory Kernel Properties}
\label{subsec:memKernel}

The incorporation of memory effects in {\cogent{COGENT$^{\textbf{3}}$}} is a \textbf{key departure from traditional approaches}. While most computational 
frameworks operate on instantaneous states, cognitive systems exhibit rich temporal dependencies that span multiple timescales. The use of kernel functions provides 
a mathematically efficient way to capture temporal dependencies.

In our system, the memory kernel $K(t)$ satisfies \textbf{three essential properties} that ensure both physical realizability and mathematical consistency.
    \begin{align*}
        K(t) &\geq 0 && \text{(Positivity)} \\[0.5em]
        \int_0^\infty K(t)dt &= 1 && \text{(Normalization)} \\
        \frac{dK}{dt} &\leq 0 && \text{(Monotonicity)}
    \end{align*}

\noindent Positivity guarantees that memory effects remain physically meaningful, while normalization provides proper scaling of historical information. The 
monotonicity condition reflects the natural decay of memory influence over time, consistent with physical intuition about information preservation in complex systems.

The memory kernel $K(t)$ modifies the effective energy landscape. For instance, if agents $i$ and $j$ have successfully collaborated in past configurations, their 
energy barrier for future interactions is lowered
    \begin{equation*}
        \begin{gathered}
            \Delta E_{eff}(i,j) = \Delta E_0(i,j) - \gamma\int_0^t K(t-s)M_{ij}(s)\,ds\\[0.3em]
            M_{ij}(t)\coloneqq \tfrac12\big(M_i(t)+M_j(t)\big)
        \end{gathered}
    \end{equation*}

\noindent This history dependence is what enables the system to \textbf{learn from past successful configurations}, similar to how cognitive systems develop preferred 
processing pathways.

The kernel takes a general form that combines multiple characteristic timescales, as
    \begin{equation}
        K(t)=\sum_{n=0}^N a_n\frac{1}{\tau_n}e^{-t/\tau_n}\quad \text{with}\quad a_n\ge 0\quad \text{and}\quad \sum_{n=0}^N a_n=1
    \end{equation}

\noindent This exponential-mixture form is positive, normalized, and non-increasing. For computational tractability we represent the filtered triad signal 
$\tilde{u}_{\tau}(t)\coloneqq \int_0^t K(t-s)u_{\tau}(s)\,ds$ via Markovian embedding, that is

\begin{equation}
    \dot{m}_{\tau}^{(n)} = -\frac{1}{\tau_n}m_{\tau}^{(n)} + \frac{a_n}{\tau_n} u_{\tau}(t)\quad \text{with}\quad \tilde{u}_{\tau}(t)=\sum_{n=0}^N m_{\tau}^{(n)}(t)
\end{equation}

In the above equation, the timescales $\tau_n$ span multiple characteristic memory horizons, allowing the system to represent both rapid adaptation and 
slower consolidation effects. The coefficients $a_n$ are nonnegative mixture weights satisfying $\sum_{n=0}^N a_n=1$, so $K(t)$ is positive and normalized (a convex 
mixture of exponential decays), while remaining analytically and numerically tractable.\footnote{While multi-timescale kernels can be computationally burdensome when 
implemented as explicit convolutions, they admit standard approximations via exponential-sum representations and Markovian embeddings. 
See, e.g., \citet{beylkin2005approximation} and \citet{beylkin2010approximation}.}
This multi-timescale structure mirrors the hierarchical organization of biological memory, from fast synaptic-like effects to slower consolidation-like dynamics.
Importantly, the exponential-mixture form enables a Markovian embedding: for a driving signal $u(t)$ we can represent the filtered memory signal 
$\tilde u(t)\coloneqq \int_0^t K(t-s)u(s)\,ds$ using auxiliary states $m^{(n)}(t)$ satisfying

    \begin{equation*}
        \dot m^{(n)}(t) = -\frac{1}{\tau_n}m^{(n)}(t) + \frac{a_n}{\tau_n}u(t)\quad \text{with}\quad \tilde u(t)=\sum_{n=0}^N m^{(n)}(t)
    \end{equation*}

\noindent Consequently, long-history dependence is realized without storing the full past trajectory, while preserving the intended multi-timescale cognitive 
interpretation.


\subsection{System Hamiltonian}
\label{subsec:sysHamiltonian}

The complete system Hamiltonian provides a \textbf{unified description} for the system dynamics, integrating diverse physical mechanisms into a coherent whole.

It is defined as
    \begin{equation}
        \mathcal{H} = \mathcal{H}_{form} + \mathcal{H}_{group} + \mathcal{H}_{mem} + \mathcal{H}_{coupling}
    \end{equation}

\noindent where $\mathcal{H}_{form}$ denotes the triad-local adversarial learning objective in parameter space. The node-field formation theory for 
$\phi_i(t)\in\mathbb{R}$ is developed separately. We expand each of the Hamiltonian components below. 

\subsubsection{Pattern Formation Component}
This component captures (adversarial) learning dynamics within the triad structure, that is
    \begin{equation}
        \mathcal{H}_{form} = \sum_{\tau\in E^{(3)}(t)} \mathcal{J}_\tau(G_\tau^{(1)}, G_\tau^{(2)}, D_\tau)
    \end{equation}

\noindent where each term $\mathcal{J}_\tau$ represents the local form Lagrangian, encoding the interplay between dual formation agents confronting an evaluator node
    \begin{equation*}
        \mathcal{J}_\tau = \mathbb{E}_{x\sim p_{data}}[\log D_\tau(x)] + \sum_{m=1}^2\mathbb{E}_{z\sim p_z}[\log(1-D_\tau(G_\tau^{(m)}(z)))]
    \end{equation*}

\subsubsection{Group Component}
\noindent For each edge $\upsilon=\{i,j\}\in E^{(2)}(t)$ define $J_{\upsilon}\coloneqq J_{ij}$ and $\kappa_{\upsilon}\coloneqq \kappa_{ij}$ (with $J_{ij}=J_{ji}$ 
and $\kappa_{ij}=\kappa_{ji}$). For each triad $\tau\in E^{(3)}(t)$ write the explicit triadic interaction as 
$\mathcal{T}_{\tau}(\mathbf{s}(t))\coloneqq \sum_{\ell=1}^{m}\ \prod_{a\in\tau} s_{a\ell}(t)$. Then, collective opinion dynamics on the induced graph and across all 
active triads is defined by
\begin{equation}
    \begin{aligned}
        \mathcal{H}_{group}(t) = &\;-\sum_{\upsilon=\{i,j\}\in E^{(2)}(t)} J_{\upsilon}\,\mathbf{s}_i(t)\cdot\mathbf{s}_j(t) -\sum_{i\in V}\mathbf{h}_i\cdot\mathbf{s}_i(t)\\
                                 &\;+\frac12\sum_{\upsilon=\{i,j\}\in E^{(2)}(t)} \kappa_{\upsilon}\,\|\mathbf{s}_i(t)-\mathbf{s}_j(t)\|_2^2 -\sum_{\tau\in E^{(3)}(t)} \lambda_{\tau}\,\mathcal{T}_{\tau}(\mathbf{s}(t))
    \end{aligned}
\end{equation}

\newpage

\subsubsection{Memory Component}
The memory contribution combines temporal integration with spatial diffusion, through
    \begin{equation}
        \mathcal{H}_{mem} = \frac{1}{2}\sum_{\{i,j\}\in E^{(2)}(t)} \kappa^{(M)}_{ij}\,(M_i-M_j)^2
    \end{equation}

\noindent The memory field $M:V\to\mathbb{R}$ is a conserved resource redistributed on the network (Section \ref{subsec:exact_conservation}), so $\sum_{i\in V}M_i$ is 
constant in time.

\subsubsection{Coupling Component}
Coupling terms connect formation intensity in a triad to local opinion coherence in that same triad. We treat the node field $(\phi_i)$ as the primary formation field. 
The triad-local GAN parameter dynamics provide a separate adaptive mechanism whose cross-triad alignment is summarized by $\phi_{align,\tau}$ and coupled to group 
coherence through $\mathcal{H}_{coupling}$.

Hence, we define the triad-level formation intensity (parameter-space) as
    \begin{equation*}
        \phi_{align,\tau}(t)\coloneqq
        \frac{1}{2\max\{1,|\mathcal{T}_\tau(t)|\}}
        \sum_{\tau'\in\mathcal{T}_\tau(t)}\sum_{m=1}^2
        \frac{\mathbf{G}_\tau^{(m)}(t)\cdot\mathbf{G}_{\tau'}^{(m)}(t)}
        {\|\mathbf{G}_\tau^{(m)}(t)\|_2\,\|\mathbf{G}_{\tau'}^{(m)}(t)\|_2}
    \end{equation*}

\noindent and the triad-level opinion coherence as
    \begin{equation*}
        m_{\tau}(t)\coloneqq \left\|\frac{1}{3}\sum_{i\in\tau}\mathbf{s}_i(t)\right\|_2
    \end{equation*}

\noindent We then set
    \begin{equation}
        \mathcal{H}_{coupling} = -\sum_{\tau\in E^{(3)}(t)} \gamma_\tau\,\phi_{align,\tau}(t)\,m_{\tau}(t)
    \end{equation}

\noindent which is well-defined and makes explicit how formation intensity and group coherence co-stabilize in triads.


\subsubsection{An Example: Three-Agent Phase Transition}
To show how the mathematical structures discussed so far manifest in a concrete setting, consider a minimal system of three agents $(i,j,k)$ with states 
$(\phi_i, \phi_j, \phi_k)$. For illustration, we use a two-state discretization $\phi_a\in\{-1,1\}$. Their collective Hamiltonian, following the 
structure introduced above, takes the form
    \begin{equation*}
        \mathcal{H}_{ijk} = -J\sum_{(ab)} \phi_a\phi_b - h\sum_a \phi_a + \frac{\gamma}{2}\sum_{\{a,b\}\in E^{(2)}} w_{ab}\,(\phi_a-\phi_b)^2
    \end{equation*}

\noindent where the first term represents pairwise interactions, the second captures external influences, and the gradient term penalizes large state differences 
between connected agents.

The temperature-dependent dynamics follow from the core equations, with noise terms enabling transitions between states, that is

    \begin{equation*}
        \langle\xi_a(t)\xi_b(t')\rangle = 2k_BT\,\delta_{ab}\delta(t-t')
    \end{equation*}

\noindent where the probability distribution of states follows directly from the Hamiltonian structure

    \begin{equation*}
        P(\{\phi_a\}) = Z^{-1}\exp(-\mathcal{H}_{ijk}/(k_BT))
    \end{equation*}

\noindent with partition function that ensures normalization

    \begin{equation*}
        Z = \sum_{\phi_i,\phi_j,\phi_k\in\{-1,1\}} \exp(-\mathcal{H}_{ijk}/(k_BT))
    \end{equation*}

\noindent This \textbf{minimal system} exhibits the essential features of our framework, namely
\begin{itemize}
    \item \textbf{memory effects} through the state-dependent diffusion term;
    \item \textbf{temperature-modulated transitions} between configurations; and
    \item the interplay between \textbf{local interactions} and \textbf{global structure}.
\end{itemize}

\noindent As we will see in later sections, this basic structure gives rise to \textbf{rich collective behaviors and phase transitions}.


\subsection{Exact Conservation Laws}
\label{subsec:exact_conservation}
We impose three exact conserved quantities by choosing \emph{conservative} update rules. Conservation is therefore \emph{structural} (built into the generator 
$\mathcal{L}$).

\noindent \textbf{(C1) Exact conservation of formation norm $Q_1$}\\

\noindent If the triad set $E^{(3)}(t)$ is dynamic (triads can be created/removed) exact conservation of a formation \emph{parameter-norm budget} requires explicit bookkeeping.
We therefore introduce a scalar reservoir variable $R_{form}(t)$ and define
    \begin{equation}
        Q_1(\Gamma)\coloneqq R_{form}(t)\;+\;\sum_{\tau\in E^{(3)}(t)}\left(\sum_{m=1}^2\|\mathbf{G}_\tau^{(m)}\|_2^2+\|\mathbf{D}_\tau\|_2^2\right)
    \end{equation}

For each active triad $\tau\in E^{(3)}(t)$ and $m\in\{1,2\}$ we constrain the formation parameter vectors to fixed norms
    \begin{equation}
        \|\mathbf{G}_\tau^{(m)}\|_2^2 = c_G \quad \text{and} \quad \|\mathbf{D}_\tau\|_2^2=c_D
    \end{equation}

\noindent where all continuous formation updates (deterministic or noisy) are projected onto the tangent spaces of these constraint manifolds.

For structural events, we impose conservation-preserving rules below.
    \begin{itemize}
        \item When a triad $\tau$ is removed, add its parameter-norm budget to $R_{form}$.
        \item When a triad $\tau$ is created, initialize its formation variables using an equal budget removed from $R_{form}$.
    \end{itemize}

\noindent With these constraints and event rules, $Q_1(\Gamma_t)$ is exactly conserved along the full dynamics, equivalently $\mathcal{L}Q_1=0$.\\

\noindent \textbf{(C2) Exact conservation of total opinion $Q_2$}\\

\noindent Let $Q_2(\Gamma)\coloneqq \sum_{i\in V}\mathbf{s}_i\in\mathbb{Z}^m$. Opinion dynamics are chosen to be \emph{Kawasaki-type} (spin-exchange on edges): at an update 
event on an edge $\{i,j\}\in E^{(2)}(t)$ we propose $\mathbf{s}_i\leftrightarrow \mathbf{s}_j$ and accept/reject with a heat-bath/Metropolis rule based on 
$\Delta\mathcal{H}_{group}$. Because updates are exchanges, $Q_2(\Gamma_t)$ is exactly conserved.\\

\noindent \textbf{(C3) Exact conservation of integrated memory $Q_3$}\\

\noindent Let $Q_3(\Gamma)\coloneqq \sum_{i\in V} M_i$. Memory evolves by conservative redistribution on the induced graph
    \begin{equation}
        \dot{M} = -\kappa_M L M + B\,\tilde{u}(t)
    \end{equation}

\noindent where $L$ is the graph Laplacian and $B$ is any node$\times$triad incidence operator satisfying $\mathbf{1}^\top B=0$ (zero-sum forcing). Therefore, 
$\frac{d}{dt}\sum_i M_i = 0$ and $Q_3(\Gamma_t)$ is exactly conserved.\\

\noindent \textbf{Symmetry interpretation}\\

\noindent The above conservation laws are compatible with symmetry statements, but exact conservation in this model is guaranteed by conservative dynamics and 
constraints, that is, by $\mathcal{L}Q_\ell=0$ for $\ell\in\{1,2,3\}$.


\subsection{Thermodynamic Framework}
\label{subsec:thermFrame}

The system is specified by
    \begin{itemize}
        \item an energy functional $\mathcal{H}$ on $\Gamma$; and
        \item a Markov generator $\mathcal{L}$.
    \end{itemize}

\noindent Whenever the jump rates satisfy detailed balance with respect to $\mathcal{H}$ at uniform temperature $T$ (e.g., $T_i\equiv T$), the corresponding 
stationary distribution has Gibbs form $\rho_\infty(\Gamma)\propto \exp(-\mathcal{H}(\Gamma)/(k_BT))$. In non-equilibrium regimes we retain $\mathcal{H}$ as an organizing 
energy landscape while the stationary state may exhibit non-zero probability currents.


\subsection{Update-operator Algebra}
\label{subsec:operatorAlgebra}

The elementary update mechanisms of {\cogent{COGENT$^{\textbf{3}}$}} (role update, triad birth and death, conservative opinion exchange, temperature diffusion, 
memory redistribution, and formation learning steps) define Markov transition kernels on $\Gamma$. Their composition generates a semigroup of evolution operators 
$P_t=e^{t\mathcal{L}}$. This provides the formal language for how local computational operations compose into global evolution.


\subsection{Symmetry Breaking Analysis}
\label{subsec:symBreaking}

{\cogent{COGENT$^{\textbf{3}}$}} exhibits multiple symmetry breaking patterns that \textbf{characterize transitions between different computational phases}. These 
patterns manifest through both continuous and discrete symmetry breaking mechanisms.

\subsubsection{Continuous Symmetries}
A fundamental continuous symmetry characterizes the parameter-coordinate sector of the pattern formation dynamics, as depicted below.
    \begin{align*}
        \textbf{Generator rotation}: \; & \mathbf{G}_\tau^{(m)} \to R_m\,\mathbf{G}_\tau^{(m)} \quad \text{with} \quad R_m\in O(p_m)
    \end{align*}

\noindent In regimes where triad-local objectives and couplings depend on parameter vectors only through norms and inner products (as in cosine-alignment 
statistics), this orthogonal invariance implies that ordering corresponds to the emergence of persistent \emph{directional} structure in parameter space. 
Operationally, this is captured by an alignment-based order parameter ($\phi_{align}(t)$) departing from its disordered baseline to become non-zero/large 
below an alignment threshold (which is model-dependent). This indicates persistent parameter-direction coherence across neighboring triads.

\subsubsection{Discrete Symmetries}
Our system also exhibits two crucial discrete symmetries, stated below in the form relevant to a dynamic triadic hypergraph.
    \begin{itemize}
        \item \textbf{Triad permutation symmetry $S_3$ (unordered hyperedges)} 
        \begin{equation}
            (i,j,k)\mapsto (\sigma(i),\sigma(j),\sigma(k)) \quad \text{where} \quad \sigma\in S_3
        \end{equation}
    \noindent reflecting that active triads $\tau=\{i,j,k\}$ are unordered interaction units. Consequently, triad-level observables and energies written 
    as symmetric functions of the three incidences are invariant under permutations of the triad labels.

    \item \textbf{Agent $\mathbb{Z}_2$ symmetry (when $\mathbf{h}_i\equiv 0$ and $\lambda_\tau\equiv 0$)} 
    \begin{equation}
        \mathbf{s}_i \to -\mathbf{s}_i
    \end{equation}
    \noindent which holds for the group Hamiltonian only when external fields vanish and odd (three-spin) interaction terms are absent. Otherwise, global 
    inversion changes the sign of the triadic interaction contribution.
\end{itemize}

\subsubsection{Coupled Order Parameters}
Symmetry breaking is characterized by a set of coupled order parameters, defined below.
    \begin{enumerate}
        \item \textbf{Alignment (formation-mechanism) order parameter}
        \begin{equation*}
            \phi_{align}(t) \coloneqq \left\langle\frac{1}{|E^{(3)}(t)|}\sum_{\tau\in E^{(3)}(t)} \phi_{align,\tau}(t)\right\rangle
        \end{equation*}

        \item \textbf{Group coherence (edge order)}
        \begin{equation*}
            c(t)\coloneqq \left\langle\frac{1}{|E^{(2)}(t)|}\sum_{\{i,j\}\in E^{(2)}(t)} \mathbf{s}_i(t)\cdot\mathbf{s}_j(t)\right\rangle \quad \text{for} \quad c(t)\in[-1,1]
        \end{equation*}

        \item \textbf{Memory-induced order}
        \begin{equation*}
            \psi_{mem}(t) \coloneqq \left\langle \frac{1}{|V|}\sum_{i\in V}\left(M_i(t)-\frac{1}{|V|}\sum_{j\in V}M_j(t)\right)^2 \right\rangle
        \end{equation*}
    \end{enumerate}

\noindent They interact through the coupled free-energy landscape
    \begin{align}
        F(\phi_{align}, c, \psi_{mem}) = & \; F_0 + \alpha\phi_{align}^2 + \beta c^2 + \gamma\psi_{mem}^2 + \nonumber\\
        & \; \lambda\phi_{align}c\psi_{mem} + O(4)
    \end{align}

\noindent The multiple minima correspond to different broken-symmetry phases, each representing distinct computational regimes in 
{\cogent{COGENT$^{\textbf{3}}$}}.\\

\noindent\underline{Remarks}\\[0.5em]
The three coupled order parameters (alignment order parameter $\phi_{align}$, group coherence $c$, and memory-induced order $\psi_{mem}$) do not evolve 
independently but rather interact nonlinearly within the shared energy landscape defined by $F(\phi_{align}, c, \psi_{mem})$. These interactions manifest 
through the cross-coupling term $\lambda\phi_{align} c \psi_{mem}$, explicitly encoding how changes in one subsystem influence stability and dynamics 
of the others. Theoretically, dominance or coexistence of these order parameters depends on relative magnitudes and signs of the coupling constants 
($\alpha$, $\beta$, $\gamma$, $\lambda$) and proximity to critical temperatures ($T_c^{align}$, $T_c^{group}$, $T_c^{mem}$). For instance, 
when $\lambda$ is sufficiently large, strong coupling induces coexisting order parameters, reflecting stable yet flexible cognitive states that 
simultaneously support robust adaptive structure (captured by $\phi_{align}$), coherent collective decisions ($c$), and memory-based adaptability ($\psi_{mem}$). 
Conversely, smaller coupling strengths or pronounced critical-temperature differences (such as $|T_c^{align} - T_c^{group}|\gg 1$, $|T_c^{align} - T_c^{mem}|\gg 1$, 
or $|T_c^{group} - T_c^{mem}|\gg 1$) typically yield competitive dynamics, where dominance by one order parameter limits others, modeling cognitive states 
characterized by specialization (e.g., highly structured adaptive dynamics without significant memory-driven flexibility, or vice versa). 
The theoretical conditions for dominance or coexistence of these order parameters therefore correspond to distinct cognitive processing regimes, 
providing a nuanced map between model dynamics and cognitive functionalities.

\subsubsection{Critical Behavior}
Near critical points, the order parameters follow well-defined scaling laws that characterize the nature of the transitions
    \begin{align*}
        \phi_{align} &\sim (T_c^{align} - T)^{\beta_{align}} \\[0.2em]
        c &\sim (T_c^{group} - T)^{\beta_{group}} \\[0.2em]
        \psi_{mem} &\sim (T_c^{mem} - T)^{\beta_{mem}}
    \end{align*}

\noindent with crossover behavior governed by the susceptibility matrix
    \begin{equation}
        \chi_{\alpha\beta} = \frac{\partial^2 F}{\partial \phi_\alpha\partial \phi_\beta}
    \end{equation}

\noindent where $(\phi_\alpha)\in\{\phi_{align},c,\psi_{mem}\}$. This critical behavior emerges from the interplay between pattern formation dynamics, agent 
interactions, and memory effects. The distinct critical temperatures $(T_c^{align}, T_c^{group}, T_c^{mem})$ reflect the \textbf{different energy scales 
associated with each subsystem}. Critical exponents $(\beta_{align}, \beta_{group}, \beta_{mem})$ characterize how {\cogent{COGENT$^{\textbf{3}}$}} approaches 
its ordered states. In turn, the coupled response of order parameters near criticality is captured by the susceptibility matrix $\chi_{\alpha\beta}$. This 
component reveals how fluctuations in one subsystem can trigger transitions in others, producing complex crossover phenomena when multiple transition points 
are approached simultaneously.

The above mathematical structures have \textbf{direct cognitive interpretations}. Symmetry breaking corresponds to the emergence of specialized processing 
roles, and the coupling between order parameters reflects how memory, collective coherence, and adaptive formation mechanisms interact in cognitive processing. 
For example, when the alignment field $\phi_{align}$ develops structure it biases the coherence statistic $c$ toward compatible configurations, similar to how 
stabilized internal representations can bias collective decision-making in cognitive systems.


\section{Formation Dynamics}
\label{sec:formation}

\subsection{Formation Mechanisms}
\label{subsec:formMechanism}

Computational structures that emerge in {\cogent{COGENT$^{\textbf{3}}$}} \textbf{follow principles analogous to pattern formation in physical systems}. Rather than 
imposing predetermined architectures, we allow computational structures to emerge (form and dissolve dynamically) through the interaction of fundamental formation 
fields (or equivalently, in response to task demands and environmental conditions). The mathematical treatment we develop below captures both the pattern formation 
process itself and the resulting computational capabilities.

To make these concepts concrete, consider a simple triad configuration where the formation potential parameters have clear interpretations: $a_i$ determines the 
tendency of individual agents to maintain stable states (analogous to neural activation thresholds), $g_{ijk}$ governs three-way interactions (similar to synaptic 
plasticity in neural circuits), and $h_{ij}$ controls higher-order correlations that stabilize emergent patterns. For instance, when $a_i > 0$ and $g_{ijk} < 0$, 
the system tends to form complementary patterns where agents develop specialized, coordinated roles.

These fields, represented by $\phi_i$, evolve according to a stochastic field theory on $\Gamma$ that captures both the deterministic aspects of pattern formation 
and stochastic fluctuations. That is,
    \begin{equation}
        \frac{d \phi_i}{d t} = -D_\phi (L\phi(t))_i - \frac{\partial V(\phi(t);t)}{\partial \phi_i} + \eta_i^{(\phi)}(t)
    \end{equation}

\noindent where $V(\phi(t);t)$ is a formation potential built from the active interaction structure at time $t$
    \begin{equation*}
        V(\phi;t) = \sum_{i\in V} a_i\,\phi_i^2 + \sum_{\tau\in E^{(3)}(t)} g_{\tau}\prod_{a\in\tau}\phi_a + \sum_{\{i,j\}\in E^{(2)}(t)} h_{ij}\,(\phi_i\phi_j)^2
    \end{equation*}

\noindent We assume node-local noise with correlations
    \begin{equation*}
        \langle \eta_i^{(\phi)}(t)\eta_j^{(\phi)}(t')\rangle = 2D_\phi(T_i(t))\,\delta_{ij}\delta(t-t')
    \end{equation*}
    
\noindent For the baseline (temperature-independent) case, $D_\phi(T)\equiv\sigma_\phi^2$.

The formation potential parameters reflect fundamental aspects of cognitive processing: $a_i$ determines individual activation thresholds, similar to neural 
firing thresholds that regulate information flow; $g_{ijk}$ governs pattern formation dynamics, analogous to synaptic plasticity rules that shape learning; while 
$h_{ij}$ stabilizes formed patterns, much like how consolidated memories resist perturbation. The emergence of coherent patterns through these formation mechanisms 
shows the capacity of {\cogent{COGENT$^{\textbf{3}}$}} \textbf{to develop organized computational structures}. This process mirrors how biological systems create 
functional neural assemblies in response to computational demands.

Also note that, in the above expression, the quadratic term represents the local tendency for pattern formation, while the cubic and quartic terms capture the nonlinear 
interactions essential for stable pattern formation. The coefficients $\{a_i, g_{ijk}, h_{ij}\}$ determine not just the strength of these interactions but also the 
types of patterns that can emerge. This formulation allows us to study \textbf{how local interactions give rise to global computational structures}.


\subsection{Topology-Dependent Interaction Terms}
\label{subsec:topologyInterTerms}

The topology of interactions in {\cogent{COGENT$^{\textbf{3}}$}} plays a crucial role in determining both the pattern formation process and the computational capabilities of resulting structures. We 
decompose the topological energy into three fundamental contributions: local interactions between neighboring elements, non-local effects mediated through the memory 
field, and curvature terms that capture the geometric properties of the forming patterns. 

This decomposition provides a natural framework for understanding how different spatial scales contribute to pattern formation. Let us write the topological energy 
$E_{topo}$ as
\vspace*{0.3em}
    \begin{equation}
        E_{topo}(\tau;t) = E_{local}(\tau;t) + E_{nonlocal}(\tau;t) + E_{curvature}(\tau;t)
    \end{equation}

    \noindent for $\tau=\{i,j,k\}\in E^{(3)}(t)$ where the local and non-local contributions on $\Gamma$ are defined by
    \begin{align}
        E_{local}(\tau;t) &\coloneqq -\sum_{\tau'\in\mathcal{T}_\tau(t)}\sum_{m=1}^2 J_{\tau\tau'}^{(m)}\,\mathbf{G}_\tau^{(m)}(t)\cdot\mathbf{G}_{\tau'}^{(m)}(t)\\
        E_{nonlocal}(\tau;t) &\coloneqq \int_0^t K(t-s)\sum_{\tau'\in E^{(3)}(s)} w\!\big(d(\tau,\tau')\big)\,\varphi_\tau(s)\,\varphi_{\tau'}(s)\,ds\\[0.2em]
        E_{curvature}(\tau;t) &\coloneqq \kappa\big((L^{(3)}\varphi(t))_\tau\big)^2 + \lambda\big(((L^{(3)})^2\varphi(t))_\tau\big)^2
    \end{align}

    \noindent and a triad neighborhood structure given by
    \begin{align*}
        \mathcal{T}_\tau(t) &\coloneqq \{\tau'\in E^{(3)}(t):\ \tau'\neq\tau,\ \tau\cap\tau'\neq\emptyset\}\\[0.3em]
        \varphi_\tau(t) &\coloneqq \frac{1}{3}\sum_{a\in\tau}\phi_a(t)\\[0.3em]
        (L^{(3)}f)_\tau &\coloneqq \sum_{\tau'\in\mathcal{T}_\tau(t)} w_{\tau\tau'}(t)\big(f_\tau-f_{\tau'}\big)
    \end{align*}

\noindent with $w_{\tau\tau'}(t)\coloneqq w\!\big(d(\tau,\tau')\big)$ defining the local environment within which direct interactions occur.

The inclusion of non-local terms through the memory kernel $K(t-s)$ allows patterns to be influenced by interactions across larger spatial and temporal scales. 
Meanwhile, the curvature terms ensure that the resulting patterns respect basic geometric constraints, leading to more stable and computationally useful structures.


\subsection{Forming of Triads}
\label{subsec:triads}

Unlike traditional architectures, triads represent the \textbf{key computational motif} in {\cogent{COGENT$^{\textbf{3}}$}}. They emerge dynamically through the interplay of local 
interactions and global constraints. The probability of pattern formation incorporates both instantaneous energetic considerations and historical effects through 
the memory field, providing a natural mechanism for context-dependent computation.

\newpage

\noindent Given a triad $\tau$ the probability of (knowledge) patterns forming follows
\vspace*{0.5em}
    \begin{equation}
        P(\tau|M,t) = \frac{\exp\!\left(-[E_{form}(\tau;t) + \gamma M_{\tau}(t)]/(k_BT_\tau(t))\right)}{\mathcal{Z}(t)}
    \end{equation}

\noindent for $\tau=\{i,j,k\}\in \binom{V}{3}$, and where
\vspace*{0.3em}
    \begin{equation*}
        M_{\tau}(t) = \int_0^t K(t-s)\mathcal{S}_{\tau}(s)\,ds
    \end{equation*}

\noindent Here, $\mathcal{S}_{\tau}(t)\in\mathbb{R}$ is a triad-local success/utility signal (model-dependent) driving memory accumulation. Concerning the formation 
energy $E_{form}$, it includes contributions from multiple sources, namely
\vspace*{0.3em}
    \begin{equation}
        E_{form}(\tau;t) = E_{comp}(\mathcal{K}_i,\mathcal{K}_j,\mathcal{K}_k) + E_{hist}(\tau;t) + E_{topo}(\tau;t)
    \end{equation}

\noindent where
    \begin{itemize}
        \item $E_{comp}$ measures \textbf{knowledge state compatibility} in $\mathcal{K}$;
        \item $E_{hist}$ accounts for \textbf{interaction history} through memory kernel; and
        \item $E_{topo}$ captures \textbf{network topology} effects on $E^{(3)}$.
    \end{itemize}

This equation captures three essential aspects: the compatibility of knowledge states, the influence of past interactions, and the constraints imposed by network 
topology. Such multi-component structure allows triads to form in a way that respects both local computational requirements and global organizational principles. 
The temperature dependence of this process provides a natural mechanism for controlling the balance between exploration and exploitation in pattern formation.


\subsection{Memory-Formation Coupling}
\label{subsec:memCoupling}

Another crucial innovation in {\cogent{COGENT$^{\textbf{3}}$}} is the \textbf{coupling mechanism between memory and pattern formation processes}. Unlike traditional computational 
architectures, where memory and processing are strictly separated, our approach allows past experiences to directly influence the emergence of new computational 
structures. This coupling occurs through multiple channels, from direct local interactions to global field effects, creating rich dynamics.

It should also be noted that this coupling introduces \textbf{characteristic frustration effects} that play a crucial role in computational capabilities. Let 
$\theta_i(t)\in\mathbb{R}/2\pi\mathbb{Z}$ denote an auxiliary phase variable associated with agent $i$ (used for synchronization/frustration diagnostics). The 
frustration score on an active triad $\tau=\{i,j,k\}\in E^{(3)}(t)$ is defined as
    \begin{equation}
        F_{\tau}(t) = \Big[J_{\tau} - \cos(\theta_i(t) + \theta_j(t) + \theta_k(t))\Big] + \gamma\int_0^t K(t-s)\,\overline{M}_{\tau}(s)\,ds
    \end{equation}

\noindent where $\overline{M}_{\tau}(t)\coloneqq \frac{1}{3}\sum_{a\in\tau} M_a(t)$. This frustration leads to modifications in the effective formation energy
    \begin{equation}
        \begin{split}
            E_{form}^{eff}(\tau;t) =&\; E_{form}(\tau;t) + \alpha\,M_{direct}(\tau;t)\\ 
                                    &\; + \beta\,M_{indirect}(\tau;t) + \gamma_M\,M_{global}(t)
        \end{split}
    \end{equation}

\noindent with memory components
\begin{align}
    M_{direct}(\tau;t) &\coloneqq \overline{M}_{\tau}(t) && \text{(Local memory)} \\[1em]
    M_{indirect}(\tau;t) &\coloneqq \frac{1}{|\mathcal{T}_\tau(t)|}\sum_{\tau'\in\mathcal{T}_\tau(t)} \overline{M}_{\tau'}(t) && \text{(Network memory)} \\
    M_{global}(t) &\coloneqq \frac{1}{|V|}\sum_{i\in V} M_i(t) && \text{(Global memory)}
\end{align}

\noindent coupled through the memory kernel $K(t)$ (capturing how past experiences influence current processing, reflecting the multi-timescale nature of cognitive 
memory) as defined in Section \ref{sec:core}.

The distinction between direct, indirect, and global memory effects provides a \textbf{natural hierarchy of memory influences} on pattern formation. Local memory 
effects capture immediate historical constraints, while indirect effects allow for the propagation of information across the network. The global memory field, meanwhile, 
enables system-wide coordination of pattern formation, crucial for maintaining coherent computational states across multiple scales.


\subsection{Formation Stability Analysis}
\label{subsec:stabAnalysis}

The stability of formed patterns in the system is \textbf{essential for reliable computation}. Our analysis encompasses both linear and nonlinear stability, as well 
as the crucial role of memory effects in pattern maintenance. This comprehensive treatment allows us to understand not just whether patterns are stable, but also 
how they respond to perturbations across different temporal and spatial scales.

\subsubsection{Linear Stability}
The linearized dynamics near a formation state $\phi^*$ is defined by
    \begin{equation}
        \frac{\partial}{\partial t}\begin{pmatrix}\delta\phi \\[0.2em] \delta T \\[0.2em] \delta M\end{pmatrix} = 
        \begin{pmatrix}
            A_{\phi\phi} & A_{\phi T} & A_{\phi M} \\[0.2em]
            A_{T\phi} & A_{TT} & A_{TM} \\[0.2em]
            A_{M\phi} & A_{MT} & A_{MM}
        \end{pmatrix}
        \begin{pmatrix}\delta\phi \\[0.2em] \delta T \\[0.2em] \delta M\end{pmatrix}
    \end{equation}

\noindent Stability requires all eigenvalues $\lambda$ of $A$ satisfy
    \begin{equation*}
        \text{Re}(\lambda) < 0
    \end{equation*}

\noindent\underline{Remarks}\\[0.5em]
The linear stability analysis presented above assumes that perturbations $\delta \phi(t)$ around equilibrium solutions remain sufficiently small throughout their 
temporal evolution. Formally, the validity of this approximation requires that perturbation magnitudes satisfy
    \begin{equation*}
        \|\delta\phi(t)\| \ll 1 \quad \forall t > t_0
    \end{equation*}
\noindent ensuring higher-order nonlinear terms remain negligible. Additionally, the spectral gap condition must hold strictly, meaning that the real parts of all 
eigenvalues of the linearized operator $A$ satisfy
    \begin{equation*}
        \text{Re}(\lambda_i) < 0 \quad \forall\, i
    \end{equation*}

This condition breaks down near critical points where $\text{Re}(\lambda_i)\approx 0$, indicating the onset of significant nonlinear effects. Such critical 
scenarios are precisely those of greatest cognitive interest, as they correspond to transitions between distinct cognitive processing regimes.

\newpage

\subsubsection{Nonlinear Stability}
Define
    \begin{itemize}
        \item{Lyapunov function candidate}
            \begin{equation}
                \begin{split}
                    V(\phi,T,M) =&\;\frac{1}{2}\sum_{\{i,j\}\in E^{(2)}(t)} w_{ij}(t)\,(\phi_i-\phi_j)^2 + \sum_{i\in V} F(\phi_i)\\ 
                                    &\;+ \frac{c_v}{2}\sum_{i\in V}T_i^2 + \frac{\gamma}{2}\sum_{i\in V}M_i^2
                \end{split} 
            \end{equation}
        \item{Stability condition}
            \begin{equation}
                \begin{split}
                    \dot{V} =&\;-\eta\sum_{\{i,j\}\in E^{(2)}(t)} w_{ij}(t)\,(\mu_i-\mu_j)^2\\
                             &\;-\sum_{\{i,j\}\in E^{(2)}(t)} \frac{\kappa}{\overline{T}_{ij}(t)}\,w_{ij}(t)\,(T_i-T_j)^2\\ 
                             &\;-\kappa_M\sum_{\{i,j\}\in E^{(2)}(t)} w_{ij}(t)\,(M_i-M_j)^2 \leq 0
                \end{split}
            \end{equation}
            \noindent where $\overline{T}_{ij}(t)\coloneqq \frac{1}{2}(T_i(t)+T_j(t))$. Here, $\mu_i$ denotes the node generalized potential associated 
            with the Lyapunov functional, defined by
            \begin{equation*}
                \mu_i \coloneqq \frac{\partial V}{\partial \phi_i} = \sum_{j:\{i,j\}\in E^{(2)}(t)} w_{ij}(t)\big(\phi_i-\phi_j\big) + F'(\phi_i)
            \end{equation*}
    \end{itemize}

\noindent The construction of appropriate Lyapunov functions in {\cogent{COGENT$^{\textbf{3}}$}} accounts for both the instantaneous state of the pattern formation fields and 
the history-dependent effects captured by the memory kernel. This leads to a \textbf{richer stability analysis} than in traditional pattern formation systems, 
reflecting the additional complexity introduced by memory effects and temperature-dependent dynamics.\\

\noindent\underline{Remarks}\\[0.5em]
Near critical points, the significance of nonlinear terms in the system's dynamics increases considerably. Even small perturbations can drive the system away from 
predictions made by purely linear stability analyses. Formally, near equilibrium states $\phi^*$ at critical temperatures $T\approx T_c$, the nonlinear dynamics can 
be systematically studied using normal form theory. 

In general, the full nonlinear dynamics near criticality are represented by
    \begin{equation*}
        \frac{d}{dt}\delta\phi = \mathcal{L}(\phi^*)\,\delta\phi + N(\delta\phi)
    \end{equation*}
\noindent where the nonlinear remainder $N(\delta\phi)$ typically includes second- and third-order terms in $\delta\phi$. These nonlinear terms qualitatively alter 
the stability landscape, potentially giving rise to characteristic bifurcations, such as pitchfork or Hopf bifurcations, depending on specific system symmetries. 

Importantly, memory kernels modify effective nonlinear stability conditions by introducing history-dependent terms. Nonlinear stability therefore requires a generalized 
Lyapunov condition incorporating memory effects, expressed formally as
    \begin{equation*}
        V(t,\delta\phi) + \gamma\int_0^t K(t-s)W(\delta\phi(s))\,ds \geq 0
    \end{equation*}
\noindent where $W(\delta\phi)$ encapsulates the appropriate nonlinear energy contributions, further constraining the stability landscape.

Nonlinear stability regimes possess profound cognitive significance as they underpin robust yet adaptive cognitive behaviors. Nonlinear terms enable symmetry-breaking 
processes critical to forming specialized computational assemblies, analogous to task-specific neural clusters in biological cognition. These clusters confer 
robustness, enabling coherent computation under noisy or variable environmental conditions, thus reflecting biological cognition’s inherent resilience.

Moreover, nonlinear instabilities near critical points provide the system with cognitive flexibility, allowing transitions between distinct cognitive states. This 
mirrors biological phenomena such as insight generation, creative leaps, and adaptive problem-solving. Specifically, cognitive flexibility emerges when the system 
dynamically exploits nonlinearities to reorganize triadic interaction structures, forming transient yet effective computational pathways responsive to changing 
contexts.

Therefore, characterizing nonlinear stability sheds light not only on when but also precisely how cognitive patterns emerge, stabilize, and dynamically 
reorganize.

\subsubsection{Memory-Dependent Stability}
Memory effects modify stability through
    \begin{equation}
        \mathcal{S}_{mem}(t) = \exp\left(\int_0^t K(t-s)\mathcal{L}(s)\,ds\right)
    \end{equation}

\noindent requiring
    \begin{equation*}
        \|\mathcal{S}_{mem}(t)\| < 1 \quad \forall t > t^*
    \end{equation*}


\subsection{Formation Order Parameter}
\label{subsec:orderParam}

The emergence of coherent computational structures can be characterized through appropriate order parameters that \textbf{capture the transition from disordered to 
ordered states}. 

Drawing inspiration from phase transitions in physical systems, we introduce an order parameter that quantifies the \textbf{degree of pattern 
formation} across different scales. This approach provides a quantitative measure of how local interactions give rise to global computational organization.

Hence, we define the emergence process by the order parameter
    \begin{equation}
        \Psi_{form}(t) = \left\langle \frac{1}{|E^{(3)}(t)|}\sum_{\tau=\{i,j,k\}\in E^{(3)}(t)} \phi_i(t)\phi_j(t)\phi_k(t)\right\rangle_T
    \end{equation}

\noindent exhibiting critical behavior
    \begin{equation*}
        \Psi_{form} \sim (T_c - T)^\beta \text{ for } T < T_c
    \end{equation*}

\noindent This order parameter connects the \textbf{microscopic field dynamics} to \textbf{macroscopic formation emergence}. The critical behavior of the order parameter reveals fundamental aspects of the structure formation process. As the system temperature approaches the critical point, 
the order parameter follows a power-law with exponent $\beta$. This characteristic exponent serves as a signature, revealing that our formation transition belongs to 
a specific class of critical phenomena. 

Its significance runs deep. Systems that appear different at the microscopic level can share the same critical behavior, unified by their common value of $\beta$. 
Such connection to critical phenomena provides deep insights into how computational structures emerge and suggests ways to control the formation process through 
\textbf{temperature modulation}.


\subsection{Example: Pattern Formation Sequence}
\label{subsec:exampleOne}

Let's consider a \textbf{minimal system} of three coupled triads. The formation sequence follows three distinct phases, enumerated below.

\begin{enumerate}
   \item \textbf{High Temperature Phase} ($T > T_c^{form}$)
      \begin{itemize}
         \item Agents explore state space freely.
         \item Roles switch frequently.
         \item No stable patterns emerge.
      \end{itemize}
   \item \textbf{Critical Region} ($T \approx T_c^{form}$)
      \begin{itemize}
         \item Long-range correlations develop.
         \item Power-law scaling of fluctuations.
         \item Intermittent pattern formation.
      \end{itemize}
   \item \textbf{Ordered Phase} ($T < T_c^{form}$)
      \begin{itemize}
         \item Stable patterns emerge.
         \item Roles become well-defined.
         \item Memory effects dominate dynamics.
      \end{itemize}
\end{enumerate}

\noindent The formation field therefore evolves as
    \begin{equation*}
        \phi_i(t) = \phi_0 + \Delta\phi(T)\left(1 - e^{-t/\tau(T)}\right)
    \end{equation*}

\noindent where $\tau(T)$ is the temperature-dependent relaxation time and $\Delta\phi(T)$ is the equilibrium pattern amplitude.


\subsection{Formation-Temperature Coupling}
\label{subsec:formTempCoupling}

Temperature plays a dual role in {\cogent{COGENT$^{\textbf{3}}$}}, acting both as a \textbf{source of fluctuations} and as a \textbf{control parameter} for pattern formation. The 
coupling between formation dynamics and temperature creates a \textbf{feedback loop} that allows the system to \textbf{self-regulate its computational behavior}. 
This mechanism is inspired by biological systems, where metabolic temperature changes can influence cognitive processing.

Building on the fundamental temperature field dynamics introduced earlier, pattern formation processes modify the temperature evolution through additional 
coupling terms
    \begin{equation}
        \begin{split}
            \frac{dT_i}{dt} =&\;-\kappa_{T,i}(t)\,(LT(t))_i - \gamma\big(T_i(t)-T_0\big) + \eta_i^{(T)}(t)\\ 
                             &\;-\eta_f \sum_{j:\{i,j\}\in E^{(2)}(t)} w_{ij}(t)\,(\phi_i(t)-\phi_j(t))^2
        \end{split}
    \end{equation}

\noindent where $(LT)_i=\sum_{j:\{i,j\}\in E^{(2)}(t)} w_{ij}(t)\big(T_i-T_j\big)$ and the memory-modulated diffusivity is
    \begin{equation*}
        \kappa_{T,i}(t) = \kappa_0\left(1+\alpha \sum_{j:\{i,j\}\in E^{(2)}(t)} w_{ij}(t)\,(M_i(t)-M_j(t))^2\right)
    \end{equation*}

\noindent The formation dissipation coefficient follows activated dynamics
    \begin{equation*}
        \eta_f = \eta_0\exp\!\left(-\frac{E_a}{\overline{T}(t)}\right) \quad \text{with} \quad \overline{T}(t)\coloneqq \frac{1}{|V|}\sum_{i\in V}T_i(t)
    \end{equation*}

\noindent and thermal noise is node-local
    \begin{equation*}
        \langle \eta_i^{(T)}(t)\eta_j^{(T)}(t')\rangle = 2D(T_i(t))\,\delta_{ij}\delta(t-t')
    \end{equation*}

This enhanced temperature dynamics captures how pattern formation processes ($\|\nabla\phi\|^2$ term) and memory effects (through $\kappa$) modify the basic thermal 
behavior. The formation dissipation term represents energy costs of pattern reorganization, while the memory-dependent diffusion coefficient allows past configurations 
to influence heat transport.


\subsection{Formation Stability}
\label{subsec:formStability}

The stability of formed patterns is crucial for reliable computation. Our analysis \textbf{extends beyond traditional linear stability} to include the effects of 
temperature fluctuations and memory-induced perturbations. The stability criterion we derive involves the full Hessian matrix of the formation potential, incorporating 
both direct formation-temperature couplings and indirect effects through the memory field.

\noindent Hence, formation stability requires that
    \begin{equation}
        \det\begin{pmatrix}
            \dfrac{\partial^2 V}{\partial \phi_i\partial \phi_j} & \dfrac{\partial^2 V}{\partial \phi_i\partial T} \\
            \dfrac{\partial^2 V}{\partial T\partial \phi_j} & \dfrac{\partial^2 V}{\partial T^2}
        \end{pmatrix} > 0
    \end{equation}

\noindent for all fluctuations around stable configurations.

In this equation, the determinant condition provides a necessary and sufficient criterion for local stability of formed patterns. It must be satisfied for all possible 
fluctuations around stable configurations, ensuring robust computational behavior. The inclusion of temperature and memory effects in the stability analysis represents 
a significant advance over traditional approaches that consider only instantaneous, fixed-temperature dynamics.


\subsection{Cross-Scale Formation Effects}
\label{subsec:formationEffects}

A distinctive feature of {\cogent{COGENT$^{\textbf{3}}$}} is its \textbf{explicit treatment of formation processes across multiple scales}. Pattern formation induces transitions that 
connect (a) microscopic dynamics to (b) macroscopic computational behavior, through (c) mesoscopic organizational structures. This multi-scale approach allows us to 
understand how local computational elements combine to create \textbf{emergent cognitive capabilities}.

The formation process therefore induces transitions across scales, which are as follows.
    \begin{align*}
        \textbf{Micro}: {}& \phi_i \to \text{local formation structure}\\[0.2em]
        \textbf{Meso}: {}& (i,j,k) \to \text{triad dynamics}\\
        \textbf{Macro}: {}& E^{(3)} \to \text{network topology}
    \end{align*}

\noindent They are mediated by both the formation field $\phi$ and temperature $T$, creating a rich interplay between different levels of 
organization. At the microscopic level, individual computational elements undergo local pattern formation. These patterns organize into mesoscopic triad structures 
that, in turn, influence the global network topology. The idea is to mirror the hierarchical organization of cognitive processing, where local neural interactions give rise to 
distributed representations. Scale coupling mechanisms allow us to formalize how information at different levels influences pattern formation, similar to how bottom-up and 
top-down processes interact in cognitive systems.

The cross-scale dynamics manifest in observable ways. At the microscopic level, individual agents adjust their states based on local interactions, analogous to neural 
firing patterns. These local changes aggregate into mesoscopic structures (triads with coordinated behavior) similar to how neural assemblies form. Finally, at the 
macroscopic level, coherent computational patterns emerge, much like how distributed cognitive functions arise from coordinated neural activity.


\subsection{Cross-Scale Coupling Mechanisms}
\label{subsec:crossScaleCoupling}

The coupling between different scales is mediated by multiple mechanisms, each contributing to the emergence of coherent computational behavior. 
We formalize these couplings through a combination of field theories, statistical mechanics, and renormalization group methods, providing a \textbf{comprehensive 
treatment of cross-scale interactions}.

\subsubsection{Scale-Dependent Fields}
Define fields at each scale as
    \begin{align}
        \phi_{micro}(t) &\coloneqq (\phi_i(t))_{i\in V}\\[0.3em]
        \phi_{meso}(t) &\coloneqq (\phi_\tau(t))_{\tau\in E^{(3)}(t)} \quad \text{where} \quad \phi_\tau(t)\coloneqq \frac{1}{3}\sum_{a\in\tau}\phi_a(t)\\
        \phi_{macro}(t) &\coloneqq e^{-\ell L(t)}\,\phi_{micro}(t)
    \end{align}

\noindent Here, $\ell>0$ is a coarse-graining scale and $e^{-\ell L(t)}$ is the graph heat-kernel (a discrete smoothing operator) on $G^{(2)}(t)$.

\subsubsection{Scale Coupling Terms}
The cross-scale Hamiltonian is defined as
    \begin{equation}
        H_{cross}(t) = \lambda_{micro,macro}\sum_{i\in V}\phi_i(t)\,\phi^{macro}_i(t)
    \end{equation}

\noindent where $\alpha,\beta \in \{micro, meso, macro\}$.

\subsubsection{Renormalization Flow}
Define the scale transformation operator as
    \begin{equation}
        \mathcal{R}_\ell:\ \phi \mapsto e^{-\ell L(t)}\phi
    \end{equation}

\noindent with scaling dimensions
    \begin{equation*}
        \Delta_\alpha = d + \frac{\gamma_\alpha}{2} + \eta_\alpha
    \end{equation*}

\noindent In turn, the renormalization group (RG) flow equations are defined as
    \begin{equation}
        \frac{dg_i}{dl} = (\Delta_i - d)g_i + \sum_{jk} C_{ijk}g_jg_k + O(g^3)
    \end{equation}

The equations for the RG flow describe how coupling parameters change across scales, providing insight into the emergence of \textbf{effective computational behaviors 
at different levels of organization}. This analysis reveals how microscopic parameters combine to determine mesoscopic and macroscopic properties, establishing a direct connection 
between local formation rules and global computational capabilities.\footnote{The formulation of cross-scale coupling via renormalization group (RG) flows and 
cross-scale fields presented here introduces considerable computational complexity in practical scenarios. Approximation methods, such as 
\textit{mean-field approximations} (where local interactions are replaced by averaged global fields) or \textit{hierarchical coarse-graining} (where finer-scale 
dynamics are systematically integrated into effective coarser-scale representations) have been successfully applied in multi-scale computational models, preserving 
essential dynamical features while ensuring computational feasibility. See, e.g., \citet{goldenfeld2018lectures} for a comprehensive overview.}

\subsubsection{Emergent Dynamics}
The effective equations of motion incorporate all scales. That is,
    \begin{equation}
        \frac{d\phi_i}{dt} = -\frac{\partial H_{eff}(\phi(t);t)}{\partial \phi_i} + \eta_i^{(\phi)}(t)
    \end{equation}

\noindent with $\langle \eta_i^{(\phi)}(t)\eta_j^{(\phi)}(t')\rangle = 2D_\phi(T_i(t))\,\delta_{ij}\delta(t-t')$, and where
    \begin{equation}
        H_{eff} = H_{micro} + H_{meso} + H_{macro} + H_{cross}
    \end{equation}

\noindent and $\Gamma_{\alpha\beta}$ captures cross-scale noise correlations.


\section{Dynamic Processes}
\label{sec:dynamics}

\subsection{Process Generator (Markov semigroup on $\Gamma$)}
\label{subsec:procGen}

The dynamics are generated by the Markov generator $\hat{\mathfrak{L}}\equiv\mathcal{L}$ on $\Gamma$, with semigroup $P_t=e^{t\mathcal{L}}$. For densities $\rho_t$ 
the evolution is governed by the forward equation
    \begin{equation}
        \partial_t \rho_t = \mathcal{L}^{\ast}\rho_t
    \end{equation}

\noindent The generator decomposes into diffusion and jump parts, corresponding to
\begin{itemize}
    \item temperature diffusion on the induced graph;
    \item formation learning dynamics on active triads;
    \item conservative opinion exchange on $E^{(2)}(t)$;
    \item Triad birth/death on $\binom{V}{3}$;
    \item role reassignment on triad incidences; and
    \item conservative memory redistribution with embedded filtered signals.
\end{itemize}


\subsection{Pattern Formation Dynamics}
\label{subsec:patternForm}

Pattern formation within each active triad $\tau\in E^{(3)}(t)$ is specified by the triad-local objective $\mathcal{J}_\tau(t)$ introduced in Section \ref{sec:core}. The 
formation variables are the generator/discriminator parameters $\theta_{G,\tau}^{(m)}$ ($m=1,2$) and $\theta_{D,\tau}$, with parameter vectors 
$\mathbf{G}_\tau^{(m)}=\mathrm{vec}(\theta_{G,\tau}^{(m)})$ and $\mathbf{D}_\tau=\mathrm{vec}(\theta_{D,\tau})$.

The learning component of the Markov generator $\mathcal{L}$ acts on observables by combining deterministic (gradient) drift in parameter space with 
temperature-modulated diffusion. At the level of parameter coordinates, a schematic triad-local SDE is
    \begin{align*}
        d\mathbf{G}_\tau^{(m)} &= -\nabla_{\mathbf{G}_\tau^{(m)}}\mathcal{J}_\tau\,dt + \sigma_G\sqrt{2T_\tau(t)}\,dW_{\tau,m}(t)\\
        d\mathbf{D}_\tau &= +\nabla_{\mathbf{D}_\tau}\mathcal{J}_\tau\,dt + \sigma_D\sqrt{2T_\tau(t)}\,dB_{\tau}(t)
    \end{align*}

\noindent with independent Wiener processes $W_{\tau,m},B_\tau$.

This is the continuous (diffusion) part of the full process; discrete events (role reassignment, opinion exchange, triad birth/death) are encoded by the jump part 
of $\mathcal{L}$ (Section \ref{sec:dynamics}).


\subsection{State Evolution}
\label{subsec:stateEvolution}

The evolution of the full configuration $\Gamma_t$ is specified as a continuous-time Markov process with generator $\hat{\mathfrak{L}}\equiv\mathcal{L}$ (Section \ref{sec:notation}).
The generator acts on \emph{observables} (test functions) rather than directly as a matrix multiplying a state vector.

Concretely, for any observable $f:\Gamma\to\mathbb{R}$ with suitable regularity, the generator identity is
    \begin{equation}
        \frac{d}{dt}\mathbb{E}\!\left[f(\Gamma_t)\right] = \mathbb{E}\!\left[(\mathcal{L}f)(\Gamma_t)\right]
    \end{equation}

\noindent Equivalently, if $\rho_t$ denotes the law (density) of $\Gamma_t$ with respect to a reference measure, it evolves by the forward equation
    \begin{equation}
        \partial_t\rho_t=\mathcal{L}^{\ast}\rho_t
    \end{equation}

\noindent The diffusion-jump decomposition of $\mathcal{L}$ in Section \ref{subsec:coupling} then specifies how continuous coordinates evolve by drift-diffusion while discrete 
coordinates evolve by jump kernels.


\subsection{Mixed Mode Analysis}
\label{subsec:mixedMode}

Mixed-mode behavior is analyzed by linearizing a \emph{reduced} set of coarse observables that couple group coherence and formation intensity. Because 
$Q_2=\sum_{i\in V}\mathbf{s}_i$ is conserved under Kawasaki exchanges (Section \ref{subsec:exact_conservation}), the natural group observable is not total 
magnetization but a \emph{coherence} statistic on the induced graph
    \begin{equation}
        c(t)\coloneqq \frac{1}{|E^{(2)}(t)|}\sum_{\{i,j\}\in E^{(2)}(t)} \mathbf{s}_i(t)\cdot\mathbf{s}_j(t)
    \end{equation}

\noindent For formation we use the triad-averaged parameter-alignment intensity
    \begin{equation}
        \begin{split}
        \phi_{align}(t)\coloneqq&\; \frac{1}{|E^{(3)}(t)|}\sum_{\tau\in E^{(3)}(t)}\phi_{align,\tau}(t)\\
        \phi_{align,\tau}(t)\coloneqq&\;
        \frac{1}{2\max\{1,|\mathcal{T}_\tau(t)|\}}
        \sum_{\tau'\in\mathcal{T}_\tau(t)}\sum_{m=1}^2
        \frac{\mathbf{G}_\tau^{(m)}(t)\cdot\mathbf{G}_{\tau'}^{(m)}(t)}
        {\|\mathbf{G}_\tau^{(m)}(t)\|_2\,\|\mathbf{G}_{\tau'}^{(m)}(t)\|_2}
        \end{split}
    \end{equation}

Assume the pair of observables $u(t)\coloneqq(\phi_{align}(t),c(t))^\top$ admits a closed (or controlled-closure) effective drift-diffusion description
    \begin{equation}
        du(t)=F(u(t))\,dt+\mathbf{B}(u(t))\,dW_t
    \end{equation}

\noindent for some drift $F:\mathbb{R}^2\to\mathbb{R}^2$ and noise matrix $\mathbf{B}$ induced by the underlying Markov generator $\mathcal{L}$.

\subsubsection{Mode Mixing Dynamics}
We do not represent the Markov generator $\mathcal{L}$ as a finite matrix acting on a state vector of the form $(s,G_1,G_2,D)$. Instead, mode mixing is quantified 
by the linearization of the effective coarse drift introduced above. 

Let $u^\ast$ be a fixed point of the effective drift, $F(u^\ast)=0$. Writing $\delta u(t)=u(t)-u^\ast$, the linearized mixed-mode dynamics are
    \begin{equation}
        d(\delta u)=\mathbf{J}_{mix}\,\delta u\,dt+\mathbf{B}(u^\ast)\,dW_t \quad \text{where} \quad \mathbf{J}_{mix}\coloneqq \left.\nabla_u F(u)\right|_{u=u^\ast}
    \end{equation}

\noindent Mixed-mode stability (in the linearized sense) holds when all eigenvalues of $\mathbf{J}_{mix}$ satisfy $\mathrm{Re}(\lambda)<0$. The off-diagonal entries 
of $\mathbf{J}_{mix}$ quantify mode-coupling. Here, $\partial F_\phi/\partial c$ encodes how group coherence modulates formation intensity, while 
$\partial F_c/\partial \phi$ encodes feedback from formation to group coherence.

\subsubsection{Stability Analysis}
Linear mixed-mode stability holds when all eigenvalues of $\mathbf{J}_{mix}$ satisfy
    \begin{equation*}
        \mathrm{Re}(\lambda)<0
    \end{equation*}

Memory effects and additional fast variables modify the effective drift $F$ and noise $\mathbf{B}$, but stability remains governed by the spectrum of the 
corresponding linearization at the operating point.


\subsection{Noise Correlation Structure}
\label{subsec:noiseCorrel}

The structure of noise correlations plays a key role in shaping our system's dynamics. Rather than treating noise as a simple perturbation, {\cogent{COGENT$^{\textbf{3}}$}} 
incorporates \textbf{carefully structured correlations} that reflect the underlying physical and computational processes. This approach allows noise to play a 
\textbf{constructive role} in pattern formation and computational organization.

\subsubsection{Temporal Correlations}
Two-time correlation functions, or
    \begin{equation}
        C_{\alpha\beta}(t,t') = \langle\xi_\alpha(t)\xi_\beta(t')\rangle = 2D_{\alpha\beta}(T)K_\xi(|t-t'|)
    \end{equation}

\noindent where $K_\xi$ is a noise temporal autocorrelation kernel and is independent of the memory kernel $K$ used in history integrals, and the diffusion 
coefficients depend on temperature
    \begin{equation*}
        D_{\alpha\beta}(T) = D_0\exp(-E_{\alpha\beta}/T)
    \end{equation*}

\subsubsection{Spatial Correlations}
For spatially extended components, we have that
    \begin{equation}
        \langle\xi_{\alpha,i}(t)\xi_{\beta,j}(t')\rangle = 2\Gamma_{\alpha\beta}(T_i(t))\,\delta_{ij}\,\delta(t-t')
    \end{equation}

\noindent with temperature-dependent mobility
    \begin{equation*}
        \begin{split}[
            \Gamma_{\alpha\beta}(T_i(t)) =&\; \Gamma_0\left(1 + \gamma\frac{\|\nabla T(t)\|_i^2}{T_i(t)^2}\right)\\[0.3em]
            \|\nabla T(t)\|_i^2 \coloneqq&\; \sum_{j:\{i,j\}\in E^{(2)}(t)} w_{ij}(t)\,(T_i(t)-T_j(t))^2
        \end{split}
    \end{equation*}

\subsubsection{Cross-Component Correlations}
Between different dynamical variables, then,
    \begin{equation}
        \langle\xi_G(t)\xi_D(t')\rangle = \sigma_{GD}(T)\exp(-|t-t'|/\tau_{GD})
    \end{equation}

\noindent where coupling strength follows
    \begin{equation*}
        \sigma_{GD}(T) = \sigma_0\tanh(T_c/T)
    \end{equation*}

\subsubsection{An Example}
For instance, take spatially correlated noise. It can coordinate transitions across multiple triads, meaning that
    \begin{equation*}
        \langle\xi_{\alpha,i}(t)\xi_{\beta,j}(t')\rangle = 2\Gamma_{\alpha\beta}(T_i(t))\,\delta_{ij}\,\delta(t-t')
    \end{equation*}
\noindent where temperature-dependent mobility $\Gamma_{\alpha\beta}(T)$ controls exploration rate. 

Higher temperatures increase mobility, enabling broader exploration of configuration space, while lower temperatures restrict exploration to local regions. This 
showcases how the structure of noise correlations \textbf{directly influences system exploration}.


\subsection{Non-Equilibrium Steady States}
\label{subsec:NESS}

A non-equilibrium steady state (NESS) is a stationary law $\rho_\infty$ satisfying $\mathcal{L}^\ast\rho_\infty=0$ while supporting non-zero stationary probability 
currents. For the diffusion sector on continuous coordinates $Y$ (Section \ref{subsec:coupling}), the stationary diffusion current is
    \begin{equation}
        \mathbf{J}_{\infty}(Y)\coloneqq \mathbf{b}(Y)\,\rho_\infty(Y)-\mathbf{D}(Y)\nabla_Y\rho_\infty(Y)
    \end{equation}

\noindent noting $\mathbf{J}_{\infty}\neq 0$ in NESS.

\subsubsection{Stationary energy balance}
For any observable $f(\Gamma)$, stationarity implies $\mathbb{E}_{\rho_\infty}[(\mathcal{L}f)(\Gamma)]=0$. In particular, for the energy landscape $\mathcal{H}$, we 
have that
    \begin{equation}
        0=\frac{d}{dt}\mathbb{E}_{\rho_\infty}[\mathcal{H}(\Gamma_t)]
        =\mathbb{E}_{\rho_\infty}\big[(\mathcal{L}\mathcal{H})(\Gamma)\big]
    \end{equation}

\noindent that is, mean energy injection by non-conservative components (cycling drift and jump fluxes) balances mean dissipation.

\subsubsection{Entropy production (diffusion + jumps)}
The diffusion-sector entropy production rate can be written (formally) as
    \begin{equation}
        \dot{S}_{tot}^{diff} =
        \int \frac{\mathbf{J}_{\infty}(Y)^{\top}\mathbf{D}(Y)^{-1}\mathbf{J}_{\infty}(Y)}{\rho_\infty(Y)}\,dY
        \;\ge\;0
    \end{equation}

\noindent Discrete updates (role jumps, opinion exchanges, triad birth/death) contribute an additional nonnegative jump-sector entropy production
    \begin{equation}
        \dot{S}_{tot}^{jump} =
        \sum_{\Gamma\to\Gamma'} \rho_\infty(\Gamma)\,k(\Gamma,\Gamma')\,
        \log\frac{\rho_\infty(\Gamma)\,k(\Gamma,\Gamma')}{\rho_\infty(\Gamma')\,k(\Gamma',\Gamma)}
        \;\ge\;0
    \end{equation}

\noindent where $k(\Gamma,\Gamma')$ are the jump rates of $\mathcal{L}_{jump}$.

\subsubsection{Fluctuation--dissipation violations}
NESS generically violates equilibrium fluctuation--dissipation relations; operationally this is detected by comparing correlation spectra and linear 
response, for example
    \begin{equation}
        \frac{S(\omega)}{\mathrm{Im}\,\chi(\omega)} \neq \frac{2T}{\omega}
    \end{equation}

\noindent with $S(\omega)$ a power spectrum of a chosen observable and $\chi(\omega)$ its response function.


\subsection{Characteristic Timescales}
\label{subsec:charTimescales}

{\cogent{COGENT$^{\textbf{3}}$}} operates across \textbf{multiple characteristic timescales}, from rapid group interactions to slow temperature-dependent processes. 
The separation of these timescales is crucial for the emergence of hierarchical computational structures, and the explicit identification of these timescales allows 
us to understand how different processes interact and contribute to overall system behavior.

Fast local dynamics, characterized by timescale $\omega_{fast}$, correspond to rapid information processing within triad structures. In turn, intermediate 
timescales govern the emergence and dissolution of computational patterns, while slow timescales control memory consolidation and temperature-mediated adaptation. 
This hierarchy \textbf{mirrors the organization of biological cognitive systems} where rapid neural firing coexists with slower synaptic plasticity and even slower 
metabolic processes.

\noindent The characteristic time scales of the system thus follow
    \begin{align*}
        \tau_{group} &\sim O(1/J) && \text{(group coupling)} \\[0.8em]
        \tau_{form} &\sim O(1/\|\mathbf{J}_{mix}\|) && \text{(pattern formation dynamics)} \\
        \tau_{mem} &\sim O\!\left(\max_{0\le n\le N}\tau_n\right) && \text{(memory)} \\
        \tau_{slow} &\sim O\!\big(1/(\kappa_T\,\lambda_2(L(t)))\big) && \text{(temperature)}
    \end{align*}

\noindent establishing a \textbf{natural hierarchy} in the system dynamics
    \begin{equation*}
        \tau_{slow} \gg \tau_{mem} \gg \tau_{form} \gg \tau_{group}
    \end{equation*}

\noindent This hierarchy of timescales reflects the multi-scale nature of cognitive processing, from rapid information integration to slower learning processes. 

The separation between scales guarantees computational stability while maintaining adaptability, and is quantified through the dimensionless ratios stated below.
    \begin{align*}
        \epsilon_1 &= \tau_{group}/\tau_{form} && \text{(Fast-formation ratio)} \\
        \epsilon_2 &= \tau_{form}/\tau_{mem} && \text{(Formation-memory ratio)} \\
        \epsilon_3 &= \tau_{mem}/\tau_{slow} && \text{(Memory-temperature ratio)}
    \end{align*}

\noindent The condition $\epsilon_i \ll 1$ ensures proper scale separation, allowing for systematic adiabatic elimination of fast variables
    \begin{equation*}
        \dot{X}_{slow} = f(X_{slow}, \langle X_{fast} \rangle) + O(\epsilon)
    \end{equation*}

This separation ensures that fast processes can reach quasi-equilibrium before significant changes occur in slower variables, enabling {\cogent{COGENT$^{\textbf{3}}$}} 
to \textbf{maintain computational stability} while adapting to changing conditions. The fast group coupling allows rapid information exchange within triads, while 
slower formation dynamics enable stable pattern emergence. Memory processes operate on longer timescales, facilitating learning and adaptation, while 
temperature-mediated changes provide the slowest, regulatory dynamics.


\subsection{Flow Decomposition}
\label{subsec:flowDecomp}

The evolution combines drift-diffusion on continuous degrees of freedom, and jump flows on discrete degrees of freedom. Accordingly, we decompose the Markov generator 
into diffusion and jump parts
    \begin{equation}
        \mathcal{L}=\mathcal{L}_{diff}+\mathcal{L}_{jump}
    \end{equation}

\noindent where $\mathcal{L}_{diff}$ acts on continuous coordinates (e.g., $T$, $M$, and continuous pattern-formation variables) and $\mathcal{L}_{jump}$ encodes 
discrete updates (e.g., role reassignment, opinion exchanges, triad birth/death). On densities, the forward generator decomposes as 
$\mathcal{L}^{\ast}=\mathcal{L}^{\ast}_{diff}+\mathcal{L}^{\ast}_{jump}$.

\paragraph{Diffusion (continuous variables)}

Let $Y_t$ denote the vector of continuous variables (the continuous coordinates of $\Gamma_t$). We represent the diffusion part by an It\^{o} stochastic differential 
equation
    \begin{equation}
        dY_t = \mathbf{b}(\Gamma_t)\,dt + \mathbf{\Sigma}(\Gamma_t)\,dW_t
    \end{equation}

\noindent where $W_t$ is a standard Wiener process and the diffusion matrix is
    \begin{equation}
        \mathbf{D}(\Gamma)\coloneqq \tfrac12\,\mathbf{\Sigma}(\Gamma)\mathbf{\Sigma}(\Gamma)^{\top}
    \end{equation}

\noindent The associated forward equation for the density $\rho_t$ on $\Gamma$ takes the conservative (current-balance) form
    \begin{equation}
        \partial_t\rho_t = -\nabla_Y\cdot\mathbf{J}_{diff}(\rho_t) \;+\; \mathcal{L}^{\ast}_{jump}\rho_t \quad \text{where} \quad
        \mathbf{J}_{diff}(\rho)\coloneqq \mathbf{b}\,\rho - \mathbf{D}\nabla_Y\rho
    \end{equation}

\noindent with $\nabla_Y$ denoting the gradient with respect to the continuous coordinates.

\paragraph{Interpretability: energy-driven drift plus cycling drift}

To connect with thermodynamic intuition, we split the drift field as
    \begin{equation}
        \mathbf{b} = \mathbf{b}_{diss} + \mathbf{b}_{cyc} \quad \text{with} \quad \mathbf{b}_{diss}\coloneqq -\mathbf{\Gamma}\,\nabla_Y\mathcal{H}
    \end{equation}

\noindent and where $\mathbf{\Gamma}$ is a (possibly state-dependent) mobility operator. The remaining component $\mathbf{b}_{cyc}$ captures non-gradient (circulating) 
transport responsible for sustained probability currents in non-equilibrium regimes. It is characterized operationally by the condition that it does not change 
$\mathcal{H}$ along its own flow,
    \begin{equation}
        \mathbf{b}_{cyc}\cdot\nabla_Y\mathcal{H}=0
    \end{equation}

\noindent and by the fact that it can support $\mathbf{J}_{diff}\neq 0$ at stationarity.

\paragraph{Equilibrium vs. non-equilibrium}

In equilibrium (detailed balance) the stationary diffusion current vanishes, that is $\mathbf{J}_{diff}(\rho_\infty)=0$. In non-equilibrium steady states, 
$\rho_\infty$ may exist while $\mathbf{J}_{diff}(\rho_\infty)\neq 0$, yielding persistent circulation in phase space. Discrete jump fluxes from $\mathcal{L}_{jump}$ 
can contribute additional stationary currents. This decomposition clarifies how energy-driven stabilization, non-gradient cycling, stochastic exploration, and jump 
events jointly determine whether {\cogent{COGENT$^{\textbf{3}}$}} maintains existing patterns or transitions to new configurations.


\subsection{Process Coupling}
\label{subsec:coupling}

The coupling between different dynamical processes is essential for generating coherent computational behavior. Our framework provides a \textbf{systematic way to 
analyze these couplings} through correlation functions that capture both instantaneous and time-delayed interactions between different dynamical variables. It is 
described by
    \begin{equation}
        \mathcal{C}_{\alpha\beta} = \langle\delta\dot{X}_\alpha\delta\dot{X}_\beta\rangle
    \end{equation}

\noindent where $X_\alpha$ represents different dynamical variables, with
    \begin{equation*}
        \delta\dot{X}_\alpha = \dot{X}_\alpha - \langle\dot{X}_\alpha\rangle
    \end{equation*}


\section{Phase Space Analysis}
\label{sec:phase}

\subsection{Phase Space Structure}
\label{subsec:phaseSpace}

The phase space $\Gamma$ captures the full richness of possible computational states and their dynamics. However, the relevant spectral decomposition is not a 
decomposition of $\Gamma$ itself (which contains discrete and continuous coordinates) but a decomposition of \emph{function spaces} on $\Gamma$. We therefore 
formulate the modal structure in terms of the Markov generator $\hat{\mathfrak{L}}\equiv\mathcal{L}$ acting on observables.

Let $\mathcal{O}$ denote a Hilbert space of observables (e.g., $\mathcal{O}=L^2(\Gamma,\mu)$ for a reference measure $\mu$ when it exists). Formally, we consider 
a spectral (or generalized spectral) decomposition into invariant subspaces
    \begin{equation}
        \mathcal{O} = \overline{\bigoplus_{\alpha}\mathcal{O}_{\alpha}(T,M)}
    \end{equation}

\noindent where each mode space is characterized by the action of the Markov generator,
    \begin{equation*}
        \mathcal{O}_{\alpha} = \{f\in\mathcal{O}:\ \mathcal{L}f = \lambda_{\alpha} f\}
    \end{equation*}

\noindent Eigenvalues with $\mathrm{Re}(\lambda_{\alpha})<0$ define characteristic relaxation timescales
    \begin{equation*}
        \tau_{\alpha} = -\frac{1}{\mathrm{Re}(\lambda_{\alpha})}
    \end{equation*}

\noindent This spectral viewpoint provides insight into the stability of computational modes and the dominant pathways through which the system transitions 
between configurations.


\subsection{Manifold Structure}
\label{subsec:manifold}

Because $\Gamma$ contains both discrete coordinates (opinions, triad sets, role incidences) and continuous coordinates (temperatures, continuous memory variables, 
and formation parameter coordinates under the chosen embedding), $\Gamma$ is not treated as a global smooth manifold. Differential-geometric objects are therefore 
introduced only on the continuous coordinate subspace.

Let $Y$ denote the vector of continuous coordinates of $\Gamma$ (as in Section \ref{subsec:coupling}). On this continuous sector one may speak of tangent and cotangent spaces $T_Y$ 
and $T_Y^\ast$, and introduce an inner product (metric) on continuous increments $\delta Y$. For example, a quadratic local metric may be written schematically as
    \begin{equation*}
        ds^2 = \langle \delta Y, \mathbf{G}(Y)\,\delta Y\rangle
    \end{equation*}

\noindent where $\mathbf{G}(Y)$ is a positive definite operator encoding relative weighting of the continuous degrees of freedom.


\subsection{Energy Landscape}
\label{subsec:energyLandscape}

The energy landscape in {\cogent{COGENT$^{\textbf{3}}$}} plays a crucial role in determining both its \textbf{computational capabilities and limitations}. Free energy 
barriers between different computational states incorporate both instantaneous energy and memory effects through the kernel $K(t)$. This memory-dependent landscape 
provides a richer structure than traditional energy landscapes, reflecting the history-dependent nature of cognitive computation.

Free energy barriers between states incorporate memory effects
    \begin{equation}
        \Delta F_{\alpha\beta} = F_{\alpha\beta}(T,M) - \min(F_\alpha, F_\beta)
    \end{equation}

\noindent with components
    \begin{equation*}
        F_\alpha = E_\alpha - TS_\alpha - \gamma\int_0^t K(t-s)M_\alpha(s)ds
    \end{equation*}

\noindent The memory-dependent free energy landscape determines system evolution. For a simple two-state subsystem, the effective barrier becomes
    \begin{equation*}
        \Delta F_{\alpha\beta}^{eff} = \Delta E_0 - T\Delta S - \gamma\int_0^t K(t-s)[M_\alpha(s) - M_\beta(s)]ds
    \end{equation*}

\noindent where $\Delta E_0$ is the bare energy difference, $\Delta S$ the entropy difference, and the memory term can either enhance or suppress transitions 
based on past history.


\subsection{Phase Space Flows}
\label{subsec:phaseSpaceFlows}

The dynamics in phase space are characterized by flow objects that respect both conservation laws and non-equilibrium constraints. The decomposition of these flows 
into conservative, dissipative, and memory-driven components provides a \textbf{complete description of the system's evolution}. This is crucial for understanding how 
computational structures emerge, persist, and transform.\footnote{Please note that while Section \ref{sec:dynamics} addressed the general decomposition of flows into their fundamental 
components, here we examine how these flows manifest specifically in phase space, subject to physical constraints and conservation laws. This treatment complements the 
earlier decomposition by showing how the system's phase space structure shapes and constrains the possible flow patterns.}

The density evolution on the continuous coordinate sector $Y$ is given by the current-balance form in Section \ref{subsec:coupling}, and phase-space transport is characterized by the 
associated probability currents (together with jump fluxes for discrete updates). In the Markov setting, the relevant \emph{flow} objects are therefore probability 
currents generated by $\mathcal{L}$ rather than deterministic trajectories on $\Gamma$.

Accordingly, we decompose phase-space transport into diffusion-induced currents, dissipative drift, and jump fluxes. On the continuous sector, the drift field is 
written as $\mathbf{b}=\mathbf{b}_{diss}+\mathbf{b}_{cyc}$ with $\mathbf{b}_{diss}\coloneqq -\mathbf{\Gamma}\nabla_Y\mathcal{H}$, where $\mathbf{b}_{diss}$ drives 
relaxation toward low-energy configurations while $\mathbf{b}_{cyc}$ generates conservative circulations that preserve the energy landscape. Memory effects 
influence transport through history-dependent forcing and gradients on the induced graph, providing an additional mechanism for adaptive reorganization.

The relative strength of these components is \textbf{modulated by temperature and memory effects}, that is
    \begin{align*}
        \|\mathbf{b}_{cyc}\|^2 &\sim \text{const},\qquad \mathbf{b}_{cyc}\cdot\nabla_Y\mathcal{H}=0 && \text{(Structure preservation)} \\[0.7em]
        \|\mathbf{b}_{diss}\|^2 &\sim \big\|\mathbf{\Gamma}\nabla_Y\mathcal{H}\big\|_2^2 && \text{(Pattern stabilization)} \\
        \mathcal{A}_{mem}(t) &\coloneqq \int_0^t K^2(t-s)\,\|\nabla_G M(s)\|^2\,ds && \text{(Memory-driven adaptation)}
    \end{align*}

\noindent where $\|\nabla_G M(s)\|^2\coloneqq \sum_{\{i,j\}\in E^{(2)}(s)} w_{ij}(s)\big(M_i(s)-M_j(s)\big)^2$.


\subsection{Phase Space Measures}
\label{subsec:phaseSpaceMeasures}

The construction of appropriate measures on our phase space must account for both the conservative and dissipative aspects of the dynamics. Its invariant measure, 
satisfying the stationary forward equation $\mathcal{L}^{\ast}\rho=0$, provides the \textbf{probability distribution for computational states}. 

This measure incorporates temperature effects and memory contributions, reflecting the full complexity of our system. It's invariant measure satisfies
    \begin{equation*}
        \mathcal{L}^{\ast}\rho = 0
    \end{equation*}

\noindent in the detailed-balance case, with explicit form

    \begin{equation}
        \rho_\infty(\Gamma)\propto \exp\!\left(-\frac{\mathcal{H}(\Gamma)}{k_BT}\right)
    \end{equation}

\noindent characterizing \textbf{long-time behavior} in {\cogent{COGENT$^{\textbf{3}}$}}. Note $\rho_\infty(\Gamma)$ is in equilibrium when $T_i\equiv T$ and 
detailed balance holds. For heterogeneous $(T_i)$ or in the presence of cycling drift/jumps that violate detailed balance, the stationary law (when it exists) 
is generally a NESS.


\subsection{Ergodic Decomposition}
\label{subsec:ergodicDecomp}

Concerning ergodic properties, these are crucial for understanding the long-term computational behavior of the system. The decomposition of 
phase space into ergodic components reveals the \textbf{natural computational modes} and their \textbf{mixing properties}. This structure is essential for 
understanding both the stability of computational states and the system's capacity for exploration.

The phase space in {\cogent{COGENT$^{\textbf{3}}$}} admits the decomposition
    \begin{equation}
        \Gamma = \bigcup_\alpha \Gamma_\alpha
    \end{equation}
\noindent with mixing properties
    \begin{equation*}
        \lim_{t\to\infty}\langle A(t)B(0)\rangle = \langle A\rangle\langle B\rangle
    \end{equation*}


\subsection{Transition State Theory}
\label{subsec:transitionState}

The transitions between basins in phase space can be analyzed through an extended version of transition state theory that incorporates memory effects. This 
framework allows us to calculate transition rates between computational states while accounting for the history-dependent nature of these transitions. The resulting 
theory provides \textbf{quantitative predictions for computational state changes} and their dependence on system parameters.

\subsubsection{Free Energy Barriers}
The transition rate between states $\alpha$ and $\beta$ is
    \begin{equation}
        k_{\alpha\beta} = \nu_0\exp\left(-\frac{\Delta F_{\alpha\beta}^{\ddagger}}{k_BT}\right)\kappa(T,M)
    \end{equation}

\noindent where
    \begin{itemize}
        \item $\nu_0$ is the characteristic frequency of state transitions;
        \item $\Delta F_{\alpha\beta}^{\ddagger}$ is the free energy barrier; and
        \item $\kappa(T,M)$ is the transmission coefficient incorporating memory effects.
    \end{itemize}

\noindent In turn, the memory-modified barrier height is
    \begin{equation}
        \Delta F_{\alpha\beta}^{\ddagger} = \Delta F_0 - \gamma\int_0^t K(t-s)M_{\alpha\beta}(s)ds
    \end{equation}

\subsubsection{Reaction Coordinates}
The optimal transition path $\mathbf{q}^*(s)$ minimizes the action
    \begin{equation}
        S[\mathbf{q}] = \int_0^1 ds\left[\|\dot{\mathbf{q}}\|^2 + \|\nabla V(\mathbf{q})\|^2\right]
    \end{equation}

\noindent subject to boundary conditions
    \begin{equation*}
        \mathbf{q}(0) \in \mathcal{B}_\alpha \quad \text{and} \quad \mathbf{q}(1) \in \mathcal{B}_\beta
    \end{equation*}

\subsubsection{Transmission Coefficient}
The memory-dependent transmission coefficient is
    \begin{equation}
        \kappa(T,M) = \left\langle\dot{q}(0)\int_0^\infty dt\;\delta(q(t))H(p(t))\right\rangle_{q=q^\ddagger}
    \end{equation}
\noindent where $H$ is the Heaviside function, and averaging is over the dividing surface.

The memory-modified transition rates between computational states follows
    \begin{equation*}
        k_{\alpha\beta} = k_0\exp\left(-\frac{\Delta F_{\alpha\beta}^{\ddagger}}{k_BT}\right)\kappa(T,M)
    \end{equation*}

\noindent where $\Delta F_{\alpha\beta}^{\ddagger}$ is the effective barrier height and $\kappa(T,M)$ is the transmission coefficient. This structure explains 
\textbf{how memory can either facilitate or inhibit transitions} by modifying effective barriers, while temperature controls the overall transition frequency.


\subsection{Basin Structure Analysis}
\label{subec:basinAnalysis}

Splitting the phase space into basins of attraction provides crucial insight into the stability and accessibility of different computational states. The 
analysis of basin structure reveals both the \textbf{local stability properties} of individual states and the \textbf{global organization} of the computational 
landscape, where its hierarchical nature reflects the multi-scale organization of the system's computational capabilities.

\subsubsection{Basin Decomposition}
For each attractor $\mathcal{A}_\alpha \subset \Gamma$, its basin of attraction becomes
    \begin{equation}
        \mathcal{B}_\alpha = \{\mathbf{x} \in \Gamma: \lim_{t\to\infty}\mathcal{U}(t)\mathbf{x} \in \mathcal{A}_\alpha\}
    \end{equation}

\noindent with basin boundary set defined by
    \begin{equation*}
        \partial\mathcal{B} = \bigcup_{\alpha\neq\beta}(\overline{\mathcal{B}_\alpha} \cap \overline{\mathcal{B}_\beta})
    \end{equation*}

The organization of phase space into these basins and their boundaries provides the foundation for understanding how different computational states interact. 
The connectivity between basins, analyzed in detail below, reveals the pathways through which the system can transition between different computational configurations.

\subsubsection{Basin Connectivity Graph}
The basin connectivity matrix is defined by
    \begin{equation}
        C_{\alpha\beta} = \begin{cases}
            1 & \text{if } \overline{\mathcal{B}_\alpha} \cap \overline{\mathcal{B}_\beta} \neq \emptyset \\
            0 & \text{otherwise}
        \end{cases}
    \end{equation}
    
\noindent with an overlap measure
    \begin{equation*}
        \omega_{\alpha\beta} = \frac{\text{vol}(\overline{\mathcal{B}_\alpha} \cap \overline{\mathcal{B}_\beta})}
        {\min(\text{vol}(\mathcal{B}_\alpha), \text{vol}(\mathcal{B}_\beta))}
    \end{equation*}

To illustrate this structure, consider three basins $\mathcal{B}_1$, $\mathcal{B}_2$, and $\mathcal{B}_3$ corresponding to different computational states. Their 
connectivity matrix
    \begin{equation*}
        C_{\alpha\beta} = \begin{pmatrix}
            0 & 1 & 0 \\
            1 & 0 & 1 \\
            0 & 1 & 0
        \end{pmatrix}
    \end{equation*}

\noindent shows that transitions must occur through $\mathcal{B}_2$, creating a hierarchical structure in phase space.

\subsubsection{Hierarchical Basin Structure}
The basin hierarchy function is defined as
    \begin{equation}
        d_{\mathcal{B}}(\alpha,\gamma)\coloneqq \min\{h\ge 0:\ (C^{h})_{\alpha\gamma}>0\}
    \end{equation}
\noindent which characterizes the minimal path length between basins in the connectivity graph.


\subsection{Fixed Point Structure}
\label{subsec:FPS}

Fixed points represent stable or metastable computational states. Their analysis through the Jacobian operator reveals not only their stability properties but also 
the nature of computational processes they can support. 

The spectrum of the Jacobian provides detailed information about the response of these states to perturbations and their role in information processing, which follows
    \begin{equation*}
        \det(\hat{\mathcal{J}}(\mathbf{x}^*) - \lambda\mathbf{I}) = 0
    \end{equation*}

\noindent For example, a pattern formation fixed point might have eigenvalues indicating stability to local perturbations but sensitivity to memory-induced 
modifications.


\subsection{Periodic Orbits}
\label{subsec:orbits}

Beyond fixed points, {\cogent{COGENT$^{\textbf{3}}$}} supports periodic orbital solutions that represent \textbf{cyclic computational processes}. The stability of these orbits, analyzed 
through Floquet theory\footnote{See, e.g., \cite{arnold1988geometric} or \cite{yakubovich1975linear} for a treatment of Floquet theory.} is crucial for understanding 
\textbf{rhythmic computational behaviors}. 

The Floquet multipliers provide \textbf{quantitative measures of orbital stability} and the system's response to \textbf{perturbations around these cycles}. Periodic 
solutions satisfy
    \begin{equation}
        \mathbf{x}(t + T) = \mathbf{x}(t)
    \end{equation}

\noindent with stability determined by the Floquet multipliers
    \begin{equation*}
        \det(\mathbf{M}(T) - \mu\mathbf{I}) = 0
    \end{equation*}


\subsection{Phase Space Transport}
\label{subsec:phaseStateTransport}

The evolution of probability density in phase space describes how computational states transition and transform over time. The rates of transition between different 
regions provide insight into preferred processing pathways and computational mechanisms, revealing both the natural progression of computation and the 
barriers that separate different processing modes.

Transport between regions follows
    \begin{equation}
        J_{\alpha\beta}(t) = \int_{\Sigma_{\alpha\beta}} \mathbf{J}_{diff}(\rho_t)\cdot d\mathbf{S}
    \end{equation}

\noindent with transition rates
    \begin{equation*}
        k_{\alpha\beta} = \frac{1}{\tau_c}\exp(-\Delta F_{\alpha\beta}/k_BT)
    \end{equation*}

\noindent Transition rates between different regions of phase space serve to \textbf{quantify the accessibility of different computational states}. These rates, determined by 
the interplay of energy barriers and noise, provide a quantitative framework for understanding both the stability of computational states and the flexibility of 
processing pathways.

The global organization of phase space, from local fixed points to transport processes, reflects both symmetry properties and conservation laws. Symmetries 
characterize transformations preserving computational structure, while the conservation laws (enforced by $\mathcal{L}Q=0$) constrain possible computational pathways 
and ensure preservation of key computational capabilities.


\subsection{Symmetry Properties}
\label{subsec:symmetry}

The system's symmetries characterize transformations that preserve its essential computational features. These go beyond simple invariances in space and time to 
include transformations that maintain computational capabilities. By analyzing the transformations that preserve both the structure of the state space and the 
system's energy (Hamiltonian), we can uncover fundamental constraints on possible computational processes.

The phase space exhibits symmetries under the group action
    \begin{equation}
        \mathcal{G} = \{g \in \text{Aut}(\Gamma): g\Gamma = \Gamma, g^*\mathcal{H} = \mathcal{H}\}
    \end{equation}
\noindent and the exact conserved quantities are enforced by conservative dynamics, e.g., by $\mathcal{L}Q=0$ for the chosen invariants.

The conserved quantities, enforced by conservative dynamics rather than derived from symmetries, provide fundamental constraints on the system's evolution, ensuring 
the preservation of essential computational capabilities while allowing for dynamic adaptation. This interplay between symmetries and conservation laws reveals the 
deep mathematical structure underlying the system's computational properties.


\section{Critical Phenomena}
\label{sec:critical}

{\cogent{COGENT$^{\textbf{3}}$}} exhibits rich critical behavior arising from the interplay between pattern formation, group dynamics, and memory effects. The analysis 
of these phenomena provides deep insight into \textbf{how computational capabilities emerge and transform} across different phases. Unlike traditional computational 
architectures with fixed processing structures, {\cogent{COGENT$^{\textbf{3}}$}} undergoes several distinct phase transitions that enable \textbf{dynamic 
reorganization of its computational resources}.

Each critical transition corresponds to a fundamental \textbf{computational or cognitive process}, as described below.
\begin{itemize}
    \item \textbf{Pattern Formation (\(T_c^{form}\)):} Analogous to the stabilization of neural activation patterns during problem-solving or perception.
    \item \textbf{Memory Consolidation (\(T_c^{mem}\)):} Represents the shift from transient, volatile states to stable, long-term memory structures, critical 
    for learning and adaptation.
    \item \textbf{Synchronization (\(T_c^{sync}\)):} Models phase alignment in distributed processes, akin to neural synchronization supporting attention 
    or group coordination.
\end{itemize}


\subsection{Computational Phase Transitions}
\label{subsec:compPhaseTran}

Building on the phase space structure defined in Section \ref{sec:phase}, {\cogent{COGENT$^{\textbf{3}}$}} exhibits multiple critical points.
    \begin{align*}
        T_c^{form} &: \text{Transition from local processing to coherent pattern formation}.\\
        T_c^{role} &: \text{Emergence of stable functional specialization}.\\[0.2em]
        T_c^{mem} &: \text{Transition from volatile to persistent information storage}.\\[0.2em]
        T_c^{sync} &: \text{Development of coordinated processing}.
    \end{align*}

The emergence of these critical points reflects fundamental changes in the system's computational architecture. While traditional computational systems rely on 
fixed processing structures, {\cogent{COGENT$^{\textbf{3}}$}} exhibits \textbf{dynamic reorganization at multiple scales}. Each critical temperature marks a qualitative shift in how 
information is processed.
    \begin{itemize}
        \item At $T_c^{form}$ local computational elements begin to exhibit coherent behavior, transitioning from isolated processing to coordinated pattern formation. 
        This enables the emergence of distributed computational structures capable of processing more complex information patterns.    
        \item The transition at $T_c^{role}$ represents the spontaneous emergence of functional differentiation, similar to how biological neural networks develop 
        specialized processing modules. This specialization enhances the system's ability to handle diverse computational tasks efficiently.
    \end{itemize}


\subsection{Computational Order Parameters}
\label{subsec:compOrderParam}

The phase transitions are quantified by distinct \textbf{order parameters} that capture different aspects of computational organization, as shown below.
    \begin{itemize}
        \item \textbf{Node-field formation}
            \begin{equation}
                \Psi_{form}(t) \coloneqq \left\langle \frac{1}{|E^{(3)}(t)|}\sum_{\tau=\{i,j,k\}\in E^{(3)}(t)} \phi_i(t)\phi_j(t)\phi_k(t)\right\rangle_T
            \end{equation}
        \item \textbf{Role stability}
            \begin{equation}
                \phi_{role}(t) = \left\langle \frac{1}{|\mathcal{I}(t)|}\sum_{(i,\tau)\in\mathcal{I}(t)} \mathbbm{1}\!\big[r(i,\tau)=r^0(i,\tau)\big]\right\rangle_T
            \end{equation}
        \item \textbf{Memory}
            \begin{equation}
                \phi_{mem} = \left\langle \frac{1}{|V|}\sum_{i\in V}\left(M_i(t)-\frac{1}{|V|}\sum_{j\in V}M_j(t)\right)^2 \right\rangle_T
            \end{equation}
        \item \textbf{Synchronization}
            \begin{equation}
                \phi_{sync}(t) \coloneqq \left\langle \left|\frac{1}{|V|}\sum_{i\in V} e^{i\theta_i(t)}\right| \right\rangle_T
            \end{equation}
    \end{itemize}

\noindent Averages are taken with respect to the measure defined in Section \ref{sec:phase}. These order parameters provide \textbf{quantitative measures of how 
different computational aspects emerge and stabilize}. The formation order parameter $\Psi_{form}$ measures the coherence of pattern formation processes, while 
$\phi_{role}$ quantifies the stability of computational roles (functional specialization) within triads. The memory order parameter $\phi_{mem}$ captures information 
storage capacity, and $\phi_{sync}$ measures the degree of coordinated processing across the network.


\subsection{Critical Exponents}
\label{subsec:criticalExp}

{\cogent{COGENT$^{\textbf{3}}$}} exhibits distinct scaling behaviors near each critical point. For each transition $\alpha \in \{form, role, mem, sync\}$, we define the reduced 
temperature $t_\alpha = (T - T_c^\alpha)/T_c^\alpha$, allowing us to characterize the scaling behavior systematically.

\subsubsection{Static Critical Exponents}
The order parameters scale as
\begin{align*}
    \phi_\alpha(t_\alpha) &\sim |t_\alpha|^{\beta_\alpha} && t_\alpha < 0 \\
    \chi_\alpha(t_\alpha) &\sim |t_\alpha|^{-\gamma_\alpha} && t_\alpha \to 0
\end{align*}
\noindent where $\beta_\alpha$ characterizes the emergence of order below the critical temperature, while $\gamma_\alpha$ describes the divergence of response 
functions near criticality.

\subsubsection{Dynamic Critical Exponents}
The temporal evolution near criticality follows
\begin{align*}
    \langle\phi_\alpha(t)\phi_\alpha(0)\rangle &\sim t^{-\theta_\alpha} && t \to \infty \\
    \tau_\alpha &\sim |t_\alpha|^{-z_\alpha} && t_\alpha \to 0
\end{align*}

\noindent with memory effects modifying these dynamic exponents through
\begin{equation*}
    z_\alpha = z_\alpha^0 + \gamma\int_0^\infty K(t)t^{\theta_\alpha-1}dt
\end{equation*}

This structure captures how memory effects modify dynamical scaling near criticality. The correlation decay exponent $\theta_\alpha$ characterizes the temporal 
persistence of patterns, while the dynamic exponent $z_\alpha$ determines how relaxation times diverge near the critical point. Memory effects couple these behaviors, 
allowing past states to influence critical dynamics.

In cognitive terms, these exponents \textbf{characterize different aspects of information processing}. The decay exponent $\theta_\alpha$ determines how quickly 
correlations between computational states diminish, while $z_\alpha$ governs the characteristic timescales of system reorganization. Higher values of $z_\alpha$ 
indicate increasingly sluggish dynamics near criticality, corresponding to more careful processing but slower adaptation to change. Lower values enable faster 
adaptation but may reduce processing depth.


\subsection{Multicritical Behavior}
\label{subsec:multicritical}

Rich multicritical phenomena arise in {\cogent{COGENT$^{\textbf{3}}$}} from the interplay between different ordering fields. At multicritical points, multiple instabilities 
coincide, leading to more complex critical behavior than at ordinary critical points. The free energy expansion captures these competing or cooperating ordering 
tendencies through coupled terms.

\subsubsection{Multicritical Points}
Multicritical points arise when multiple instabilities coincide, creating complex free energy landscapes and allowing for simultaneous transitions in different 
order parameters. The free energy near a multicritical point is defined as
    \begin{equation}
        F(\{\phi_i\}, \{r_j\}) = F_0 + \sum_{i,j} r_j\phi_i^2 + \sum_{ijkl} u_{ijkl}\phi_i\phi_j\phi_k\phi_l + O(\phi^6)
    \end{equation}
\noindent where the coefficients $r_j$ act as control parameters for different types of ordering, while the coupling terms $u_{ijkl}$ determine how different order 
parameters interact. 

This structure allows for the emergence of phases with multiple types of order, as well as complex phase boundaries and crossover phenomena, thus having significant 
computational implications.
\begin{itemize}
    \item \textbf{Enhanced Sensitivity:} At multicritical points, small perturbations in input or system parameters can trigger large-scale reconfigurations, 
    enabling flexible adaptation to changing environments.
    \item \textbf{Complex Phases:} Coupled order parameters (\(\phi_{form}, \phi_{mem}, \phi_{sync}\)) allow for phases with coexisting structures, such as 
    stable patterns that adapt dynamically to memory constraints.
    \item \textbf{Biological Analogy:} Similar behavior is observed in critical brain dynamics, where multicriticality supports flexible transitions between cognitive 
    states like task-switching and memory retrieval [see \cite{beggs2008criticality}].
\end{itemize}

Hence, multicritical points reflect the ability for handling simultaneous transitions across cognitive dimensions, such as \textbf{shifting between memory recall and 
real-time computation}. For example, in multitasking scenarios, these points enable dynamic balancing of resource allocation, ensuring optimal performance despite 
competing demands.

\subsubsection{Coupled Order Parameters}
The coupled equations near multicritical points are
    \begin{align}
        \frac{\partial\phi_1}{\partial t} &= r_1\phi_1 - u_1\phi_1^3 - v_{12}\phi_1\phi_2^2 + \xi_1 \\
        \frac{\partial\phi_2}{\partial t} &= r_2\phi_2 - u_2\phi_2^3 - v_{21}\phi_2\phi_1^2 + \xi_2
    \end{align}

\noindent The interaction between coupled order parameters (\(\phi_{form}, \phi_{mem}, \phi_{sync}\)) enables computational flexibility.
\begin{itemize}
    \item \textbf{Formation (\(\phi_{form}\)):} Provides \textbf{structural coherence} for local computations.
    \item \textbf{Memory (\(\phi_{mem}\)):} Introduces \textbf{temporal depth}, enabling {\cogent{COGENT$^{\textbf{3}}$}} to incorporate past states.
    \item \textbf{Synchronization (\(\phi_{sync}\)):} \textbf{Aligns distributed agents} for coordinated computation.
\end{itemize}

These parameters dynamically interact, allowing the system to balance stability and adaptability across multiple scales. The interaction of order parameters is 
crucial in tasks requiring simultaneous stabilization (memory), coordination (synchronization), and adaptability (formation). For example, in collaborative 
problem-solving, synchronized agents (\(\phi_{sync}\)) rely on stable memory (\(\phi_{mem}\)) to dynamically reorganize patterns (\(\phi_{form}\)) based on 
task demands.

\subsubsection{Phase Diagram Structure}
Multicritical point occurs at
    \begin{equation}
        \mathbf{r}_c = \{r_j: \det(\partial^2F/\partial\phi_i\partial\phi_j) = 0\}
    \end{equation}

\noindent with scaling form defined by
    \begin{equation*}
        F_{sing} \sim |r-r_c|^{2-\alpha}\mathcal{F}_\pm(\{g_i|r-r_c|^{-\phi_i}\})
    \end{equation*}


\subsection{Finite-Size Scaling}
\label{subsec:finiteScale}

The system size $N\coloneqq |V|$ introduces an additional length scale that modifies critical behavior. Finite-size scaling theory provides a mechanism for understanding how 
critical phenomena manifest in finite systems and for extracting critical exponents from finite-size data. In {\cogent{COGENT$^{\textbf{3}}$}} it serves to 
quantify \textbf{how the system's behavior is modified by size constraints}, providing insights into real-world computational limits. 

\noindent For example,
\begin{itemize}
    \item \textbf{Memory Scaling (\(M_L\)):} Reflects how the system adapts its information processing tasks given limited resources.
    \item \textbf{Role Specialization (\(\phi_{role}\)):} Demonstrates how stable roles emerge even in finite networks, ensuring functional modularity without 
    requiring infinite agent counts.
\end{itemize}

This is not unusual. Finite-size effects often emerge, say, in neural network simulations, where the number of nodes limits the expressiveness of learned 
representations. Similarly, in multi-agent systems, finite resources constrain role specialization and synchronization, leading to observable scaling deviations 
from ideal behavior.

\subsubsection{Scaling Functions}
The general scaling form is
    \begin{equation}
        O(t,h,N) = N^{x_O/\nu}\Phi_O(tN^{1/\nu}, hN^{y_h/\nu})
    \end{equation}

\newpage

\noindent where
    \begin{itemize}
        \item $x_O$ is the scaling dimension of observable $O$;
        \item $y_h$ is the scaling dimension of the field $h$; and
        \item $\Phi_O$ is the scaling function.
    \end{itemize}
\noindent The scaling function $\Phi_O$ encapsulates how observables depend on both the distance from criticality and the system size. The scaling dimensions $x_O$ 
and $y_h$ characterize how different quantities scale with system size, while maintaining the connection to bulk critical behavior.

\subsubsection{Specific Observables}
Key quantities scale as
    \begin{align}
        \chi_N(t) &= N^{\gamma/\nu}\tilde{\chi}(tN^{1/\nu}) \\
        c_N(t) &= N^{-\beta/\nu}\tilde{c}(tN^{1/\nu}) \\
        C_N(t) &= N^{\alpha/\nu}\tilde{C}(tN^{1/\nu})
    \end{align}

\noindent These scaling forms for susceptibility, coherence, and specific heat provide practical tools for analyzing critical behavior in finite systems. The 
universal scaling functions $\tilde{\chi}$, $\tilde{c}$, and $\tilde{C}$ contain information about both the approach to criticality and finite-size effects. 

\subsubsection{Corrections to Scaling}
Leading corrections take the form
    \begin{equation}
        O(t,L) = L^{x_O/\nu}[\Phi_O(tL^{1/\nu}) + L^{-\omega}\Psi_O(tL^{1/\nu})]
    \end{equation}

\noindent where $\omega$ is the correction-to-scaling exponent, and modified correlation length given by
    \begin{equation*}
        \xi_L = L\left[1 + aL^{-\omega} + O(L^{-2\omega})\right]
    \end{equation*}

The corrections to scaling quantify how finite-size effects systematically modify the critical behavior observed in infinite systems, providing a way for 
\textbf{understanding scaling in real-life implementations}.


\subsection{Crossover Behavior}
\label{subsec:crossover}

The presence of multiple critical points leads to \textbf{rich crossover phenomena} where the system interpolates between different universality classes. 
This crossover behavior in {\cogent{COGENT$^{\textbf{3}}$}} is particularly relevant when multiple length scales compete in determining critical system properties.

Near multicritical points, crossover scaling functions become
    \begin{equation}
        \Phi_{cross}(t_1L^{y_1}, t_2L^{y_2}, gL^{y_g}) = L^{-\beta/\nu}\tilde{\Phi}(t_1L^{y_1}, t_2L^{y_2}, gL^{y_g})
    \end{equation}

\noindent with generalized hyperscaling
    \begin{equation*}
        2-\alpha = \nu\sum_i y_i
    \end{equation*}

\noindent where $y_i$ are the scaling dimensions of the relevant fields. This generalized hyperscaling relation shows how different scaling fields combine to 
determine the critical behavior, with the scaling dimensions $y_i$ characterizing the relative importance of different fields in driving the crossover.


\subsection{Correlation Functions}
\label{subsec:correl}

The spatial organization of correlations in {\cogent{COGENT$^{\textbf{3}}$}} provides key insights into \textbf{how information processing capabilities emerge across 
different scales}. While traditional architectures often assume fixed connection patterns, we allow for \textbf{dynamic correlation structures} that adapt 
to computational demands. The two-point correlation functions quantify how computational coherence develops spatially.

Two-point correlation functions on the configuration space $\Gamma$ quantify how computational coherence develops spatially, and are defined as
    \begin{equation}
        G_\alpha(i,j;t) \coloneqq \langle \phi_{\alpha,i}(t)\,\phi_{\alpha,j}(t)\rangle \quad \text{where} \quad d_{ij}(t)\coloneqq \mathrm{dist}_{G^{(2)}(t)}(i,j)
    \end{equation}

\noindent with scaling form
    \begin{equation*}
        G_\alpha(i,j;t) \sim \frac{1}{d_{ij}(t)^{d_{eff}-2+\eta_\alpha}}\,\exp\!\left(-\frac{d_{ij}(t)}{\xi_\alpha}\right)
    \end{equation*}

\noindent where correlation lengths diverge as
    \begin{equation*}
        \xi_\alpha \sim |T - T_c^\alpha|^{-\nu_\alpha}
    \end{equation*}

These correlation lengths $\xi_\alpha$ have direct computational significance, as they determine the spatial scale over which information processing can occur. 
When correlation lengths diverge near critical exponents $\nu_\alpha$ computation are coordinated in {\cogent{COGENT$^{\textbf{3}}$}} across all scales, enabling 
both fine-grained local processing and system-wide information integration. This multi-scale processing capability is \textbf{crucial for cognitive tasks that 
require both detailed feature analysis and global pattern recognition}.


\subsection{Scaling Relations}
\label{subsec:scalingRel}

Standard scaling laws connect different aspects of system behavior near critical points, reflecting deep symmetries in how {\cogent{COGENT$^{\textbf{3}}$}} reorganizes.\footnote{See 
\cite{goldenfeld2018lectures} or \cite{stanley1987introduction} for an exposition of scaling relations and critical phenomena.} 
    \begin{align*}
        \alpha + 2\beta + \gamma &= 2 && \text{(Rushbrooke)} \\
        \gamma &= \nu(2-\eta) && \text{(Fisher)} \\
        \nu d &= 2-\alpha && \text{(Josephson)} \\
        \gamma &= \beta(\delta-1) && \text{(Widom)}
    \end{align*}

\noindent They connect the critical exponents of different physical quantities, providing consistency checks for the critical behavior and reflecting 
deeper symmetries in the system's organization, and also reflect underlying cognitive processes. Hence,
    \begin{itemize}
        \item \textbf{\(\beta\)} describes how ordered states emerge, like memory consolidation or pattern stabilization;
        \item \textbf{\(\gamma\)} captures how strongly the system responds to small perturbations;
        \item \textbf{\(\alpha\)} measures singularities in system's energy usage;
        \item \textbf{\(\nu\)} characterizes how correlation length grows;
        \item \textbf{\(\eta\)} describes spatial decay of correlations; and
        \item \textbf{\(\delta\)} measures response at the critical point itself.
    \end{itemize}

These scaling relations are universal, appearing in systems ranging from magnets to neural networks. In our design, they describe how different aspects 
of information processing (like pattern formation, memory consolidation, and collective decision-making) relate to each other \textbf{near critical transitions}. For 
instance, the Rushbrooke relation connects pattern emergence (\(\beta\)), response sensitivity (\(\gamma\)), and energy efficiency (\(\alpha\)), showing how these 
cognitive capabilities must balance each other.


\subsection{Universality Classes}
\label{subsec:univClass}

The critical behavior of {\cogent{COGENT$^{\textbf{3}}$}} falls into distinct universality classes, determined by the symmetries of the order parameters and dimensionality. Each transition 
exhibits critical exponents characteristic of its universality class. Critical behavior classification is based on
\begin{itemize}
    \item \textbf{Formation transition:} $O(n)$ universality with $n = \dim(\mathcal{K})$;
    \item \textbf{Role freezing:} $Z_q$ symmetry where $q = |\mathcal{R}|$;
    \item \textbf{Memory transition:} Long-range interactions on $\mathcal{K}$; and
    \item \textbf{Synchronization:} Kuramoto class on $\Gamma$.
\end{itemize}

The formation transition belongs to the O(n) universality class reflecting the continuous symmetry of the order parameter space, while the role freezing transition 
exhibits discrete symmetry breaking. The memory and synchronization transitions introduce additional complexity through their non-local nature and dynamical aspects.

In {\cogent{COGENT$^{\textbf{3}}$}} critical behavior falls into distinct universality classes. These classes have direct cognitive and computational interpretations, 
as outlined below.
\begin{itemize}
    \item \textbf{Formation transition ($T_c^{form}$):} Belongs to the $O(n)$ universality class, reflecting the continuous symmetry in the knowledge state space 
    $\mathcal{K}$. In cognitive terms, this class represents flexibility in adapting to different learning contexts, with symmetry breaking corresponding to the 
    emergence of stable patterns (e.g., converged neural activation states).
    \item \textbf{Role freezing ($T_c^{role}$):} Mapped to the $Z_q$ universality class with $q = |\mathcal{R}|$, indicating discrete role specialization. 
    Computationally, this describes transitions where agents adopt distinct, stable functional roles, enhancing modular organization.
    \item \textbf{Synchronization ($T_c^{sync}$):} Falls under the Kuramoto universality class, associated with phase coherence in oscillator systems. This 
    transition models distributed agents achieving coordination, analogous to neural synchronization supporting attention or collective decision-making.
    \item \textbf{Memory consolidation ($T_c^{mem}$):} Reflects transitions driven by long-range interactions in the knowledge state space $\mathcal{K}$. 
    This class captures the system’s ability to consolidate volatile short-term memory into persistent long-term memory states.
\end{itemize}

These universality classes have analogs in diverse domains. For instance, the $O(n)$ universality class is often observed in spin-glass systems and neural field 
models, where symmetry breaking leads to the emergence of stable states. The Kuramoto class, widely studied in oscillator synchronization, finds parallels in 
biological rhythms like circadian cycles (or, equivalently, time-based modulation of computational resources based, say, on data inflow patterns) and motor 
coordination (mirrored by the synchronization and timing in the distributed actions of agents). 

Overall, it builds the foundation of our design, that uses biological motifs (and their mechanics) in the search for a (basic) computational equivalence 
to human cognition. 


\subsection{Renormalization Group Analysis}
\label{subsec:renormGA}

The Renormalization Group (RG) provides a systematic framework for understanding the emergence of universal behavior near critical points. By analyzing how couplings 
transform under changes of scale, we gain insight into the stability of different critical behaviors and their interdependence.

\noindent The RG flow equations on coupling space are defined as
    \begin{equation}
        \frac{dg_i}{dl} = \beta_i(\{g_j\})
    \end{equation}

\noindent with fixed points satisfying
    \begin{equation*}
        \beta_i(\{g_j^*\}) = 0
    \end{equation*}

The RG transformations preserve the structure of $\mathcal{H}$ defined in Section \ref{sec:core}. These fixed points of the RG flow determine the universality classes 
of the critical behavior. The stability of these fixed points under RG transformations dictates which one controls the observed critical phenomena.


\subsection{Crossover Phenomena}
\label{subsec:crossPhen}

The crossover scaling function incorporates multiple fields
    \begin{equation}
        \phi = |t|^\beta F_\pm(h|t|^{-\beta\delta}, L|t|^{-\nu})
    \end{equation}
\noindent where $t = (T-T_c)/T_c$ and $F_\pm$ are scaling functions on $\Omega$.


\subsection{Critical Dynamics}
\label{subsec:critDyn}

Dynamic scaling extends critical behavior to temporal evolution, relating characteristic time scales to spatial correlation lengths through the dynamic 
critical exponent $z$. 

The dynamical critical exponent relating space and time scales are
    \begin{equation}
        \tau \sim \xi^z \sim |T-T_c|^{-\nu z}
    \end{equation}

\noindent In turn, the mode-coupling equations on $\Gamma$ become
    \begin{equation}
        \frac{\partial\phi_\alpha}{\partial t} = -\Gamma_\alpha\frac{\delta\mathcal{F}}{\delta\phi_\alpha} + \eta_\alpha
    \end{equation}
\noindent describing how different fields interact dynamically near criticality, with temperature-dependent coefficients $\Gamma_\alpha$ controlling the relaxation 
rates. These equations capture how fluctuations of different fields influence each other's dynamics.

\subsubsection{Transition Rates}

Transition rates $k_{\alpha\beta}$ quantify the \textbf{likelihood of transitions between computational states} $\alpha$ and $\beta$ in phase space. These rates are 
determined by the free energy barrier $\Delta F_{\alpha\beta}$ and the system's temperature $T$, following an Arrhenius-like form, hence
\begin{equation}
    k_{\alpha\beta} = \nu_0 \exp\left(-\frac{\Delta F_{\alpha\beta}}{k_BT}\right)
\end{equation}

\noindent where \(\nu_0\) is the attempt frequency. Faster transition rates correspond to higher flexibility in adapting to new computational states or stimuli, 
while slower rates indicate greater stability, reducing susceptibility to transient perturbations.

The interplay between dynamic exponents $z$, correlation lengths $\xi$, and memory effects modifies these rates, creating a \textbf{balance between flexibility and 
stability} that is \textbf{crucial for emergent cognitive processes}. For instance, near $T_c$, where critical fluctuations dominate, transition rates can exhibit 
significant scaling behavior, facilitating dynamic reorganization of computational resources.


\subsection{Multi-Critical Points}
\label{subsec:multiCrit}

At multi-critical points, multiple instabilities compete or cooperate, leading to a more complex free energy landscape than at ordinary critical points. The free 
energy functional includes couplings between different order parameters, allowing for rich phase diagrams.

The free energy functional takes the form
    \begin{equation}
        \mathcal{F} = \sum_{\alpha,\beta} r_{\alpha\beta}\phi_\alpha\phi_\beta + \sum_{\alpha,\beta,\gamma,\delta} u_{\alpha\beta\gamma\delta}\phi_\alpha\phi_\beta\phi_\gamma\phi_\delta
    \end{equation}
\noindent defined on the full state space $\Omega$. The coefficients $r_{\alpha\beta}$ control the distance to instability in different channels, while the 
fourth-order couplings $u_{\alpha\beta\gamma\delta}$ determine how different ordering tendencies interact. This structure allows for phases with multiple types 
of order as well as complex phase boundaries.


\section{Collective States}
\label{sec:collective}

Collective states represent the emergent behavior of the system as it transitions from localized, individual computational processes to large-scale coordinated 
dynamics. This section explores the interplay between individual agents and their aggregation into structured patterns, providing insight into {\cogent{COGENT$^{\textbf{3}}$}} 
computational capacity and its cognitive analogs. Each subsection focuses on a specific aspect of this collective behavior, bridging the mathematical 
formalization with its computational and cognitive implications.

\subsection{State Space Decomposition}
\label{subsec:ssDecomp}

Collective behavior is most naturally described through a modal decomposition of \emph{observable functions} on the configuration space, not by decomposing $\Gamma$ 
itself. Hence, let $\mathcal{O}$ be an observable Hilbert space such as $L^2(\Gamma,\mu)$. A spectral (or generalized spectral) decomposition can be written as
    \begin{equation}
        \mathcal{O} = \overline{\bigoplus_{\alpha}\mathcal{O}_{\alpha}}
        \qquad
        \mathcal{O}_{\alpha}=\{f\in\mathcal{O}:\ \mathcal{L}f=\lambda_\alpha f\}
    \end{equation}

\noindent where $\mathcal{L}$ is the Markov generator. Each mode corresponds to a collective computational pattern in the sense that it defines a coherent relaxation 
channel and characteristic timescale
    \begin{equation*}
        \tau_\alpha = -\frac{1}{\mathrm{Re}(\lambda_\alpha)}\quad (\mathrm{Re}(\lambda_\alpha)<0)
    \end{equation*}


\subsection{Computational States}
\label{subsec:compStates}

Building on the individual agent states defined in Section \ref{subsec:agentProperties}, the collective state takes the form
    \begin{equation}
        \Psi_{comp}(t) \coloneqq \sum_{\alpha} c_\alpha\!\big(\overline{T}(t),\overline{M}(t)\big)\,\psi_\alpha(\Gamma_t)
    \end{equation}

\noindent where
    \begin{equation*}
        \overline{T}(t)\coloneqq \frac{1}{|V|}\sum_{i\in V}T_i(t) \quad \text{and} \quad \overline{M}(t)\coloneqq \frac{1}{|V|}\sum_{i\in V}M_i(t)
    \end{equation*}

\noindent with the temperature-dependent weights $c_\alpha(\overline{T}(t),\overline{M}(t))$ modulating the contribution of each basis state, 
and $\psi_\alpha(\Gamma_t)$ incorporating historical influences through the memory variables in $\Gamma_t$. This structure captures how collective 
computation emerges from the interplay between instantaneous dynamics and memory effects.

Basis states follow a statistical mechanical form on the configuration space $\Gamma$

    \begin{equation}
        \psi_\alpha(\Gamma) = \mathcal{N}_\alpha\exp\!\left(-\frac{E_\alpha\!\big[\{\mathbf{G}_\tau^{(m)}\}, \{\mathbf{s}_i\}, M\big]}{\overline{T}(\Gamma)}\right)
        \; \text{with} \;\;
        \overline{T}(\Gamma)\coloneqq \frac{1}{|V|}\sum_{i\in V}T_i
    \end{equation}

\noindent where $E_\alpha$ represents the effective energy of collective configurations, incorporating pattern formation fields $\mathbf{G}_\tau$, agent opinions 
$s_i$, and the memory field $M$.

\newpage

The normalization factor $\mathcal{N}_\alpha$ ensures proper probability interpretation. It reveals how collective computation is possible through
\begin{itemize}
    \item temperature-dependent exploration of state space through $c_\alpha(T(t))$,
    \item memory-mediated adaptation via $\psi_\alpha(\Gamma_t)$, and
    \item interaction between pattern formation and opinion dynamics in $E_\alpha$,
\end{itemize}

\noindent providing a framework for understanding how the system balances exploration and exploitation at the collective level.


\subsection{Information Processing}
\label{subsec:infoProc}

The collective processing capabilities of {\cogent{COGENT$^{\textbf{3}}$}} emerge from information flow between computational modes. We quantify this through the 
mutual information between modes $\alpha$ and $\beta$:
    \begin{equation}
        \mathcal{I}_{\alpha\beta} = \int dx dy P(x,y)\log\frac{P(x,y)}{P(x)P(y)}
    \end{equation}

\noindent measuring the interdependence of different collective computational processes. High mutual information indicates strong coordination between modes, 
essential for complex collective computation.

The system's \textbf{processing capacity} is characterized through the spectral measure
    \begin{equation}
        C(\omega) = \int_0^\infty d\tau e^{i\omega\tau}\langle\Psi_{comp}(t+\tau)\Psi_{comp}(t)\rangle_T
    \end{equation}

\noindent where peaks in $C(\omega)$ reveal frequencies at which collective computation is most efficient. This spectral decomposition provides insight into the 
temporal organization of collective processing, with distinct peaks corresponding to different cognitive timescales.

The relationship between mutual information and processing capacity reveals \textbf{fundamental aspects of collective computation}.
\begin{itemize}
    \item Mode coupling strength determines information integration capacity.
    \item Spectral peaks indicate preferred collective processing frequencies.
    \item Memory effects modulate both mode coupling and processing efficiency.
\end{itemize}

\noindent Also note the collective processing efficiency depends on both instantaneous coordination and historical context

    \begin{equation*}
        \eta_{proc}(\omega) = \frac{C(\omega)}{\mathcal{I}_{total}}\left(1 + \gamma\int_0^t K(t-s)M(s)ds\right)
    \end{equation*}

\noindent where $\mathcal{I}_{total} = \sum_{\alpha,\beta}\mathcal{I}_{\alpha\beta}$ measures total inter-mode information flow. This efficiency metric captures 
how memory effects ($K(t)$) and mode coupling together enable robust collective computation.


\subsection{Memory Embedding}
\label{subsec:memEmbed}

Building on the memory kernel properties established in Section \ref{subsec:expLagrangian}, and unlike individual agent memory discussed in Section \ref{sec:architecture}, collective knowledge in 
{\cogent{COGENT$^{\textbf{3}}$}} manifests through patterns embedded within the knowledge space $\mathcal{K}$. These patterns take the form
    \begin{equation}
        \xi_\mu \coloneqq \sum_{i\in V} \mathbf{w}_i^\mu \cdot \mathbf{s}_i \quad \mathbf{s}_i\in\mathcal{S}=\{-1,1\}^m
    \end{equation}

\noindent where $w_i^\mu$ represents the contribution of agent $i$ to collective pattern $\mu$. The distributed nature of this storage enables \textbf{robust pattern 
preservation despite individual agent fluctuations}.

\noindent The system's collective storage capacity follows a fundamental bound
    \begin{equation}
        \alpha_c = \frac{P_{stored}}{|V|} \sim O(1)
    \end{equation}

\noindent where $P_{stored}$ is the number of distinct patterns that can be reliably maintained. This capacity scaling reflects the balance between network size and 
pattern complexity.

The \textbf{quality of collective memory} depends on both pattern distribution and retrieval dynamics, that is

    \begin{equation}
        Q_\mu(t,T) = \frac{\langle \xi_\mu,\,\Psi_{comp}(t)\rangle_T}{\|\xi_\mu\|\,\|\Psi_{comp}(t)\|}\left(1 + \gamma(T)\int_0^t K(t-s)M(s,T)ds\right)
    \end{equation}

\noindent where the overlap measure $Q_\mu(t)$ quantifies pattern retrieval fidelity. The memory kernel $K(t)$ enables experience-dependent modulation of retrieval 
quality, while the memory field $M(s)$ captures historical context.

\noindent This collective memory architecture supports key cognitive capabilities, outlined below.
\begin{itemize}
    \item Distributed storage enhances robustness to local perturbations.
    \item In turn, capacity scales with network size (while maintaining retrieval quality).
    \item Memory kernel enables context-dependent pattern accessibility.
    \item Collective patterns persist beyond individual agent timescales.
\end{itemize}

The interplay between storage capacity and retrieval quality is a key trade-off in collective memory
\begin{equation*}
    P_{eff} = P_{stored}\cdot\langle Q_\mu\rangle_\mu \leq P_{max}\exp(-\beta/T)
\end{equation*}

\noindent where temperature $T$ modulates the balance between capacity and fidelity. Higher temperatures increase capacity but reduce retrieval quality, mirroring 
the exploration-exploitation trade-off in cognitive systems.


\subsection{Attractor Dynamics}
\label{subsec:attractorDyn}

Building on the basin structure analysis developed in Section \ref{subec:basinAnalysis}, we examine how attractors emerge in collective computational states. These collective attractors 
represent \textbf{stable computational configurations} that support sustained information processing. State transitions between attractors $\mathcal{A}_\alpha$ and 
$\mathcal{A}_\beta$ follow pathways $\mathcal{P}_{\alpha\beta}$ in configuration space
    \begin{equation}
        \mathcal{P}_{\alpha\beta} = \{\gamma(t): \gamma(0)\in\mathcal{A}_\alpha, \gamma(1)\in\mathcal{A}_\beta\}
    \end{equation}

\noindent where $\gamma(t)$ represents collective transition trajectories in the continuous coordinate sector $Y$. These pathways enable the system to shift between 
different collective computational modes while maintaining coherent processing.

The minimal action principle identifies efficient transition paths through collective state space, that is
    \begin{equation*}
        S[y] = \int_0^1 \big(\dot{y}(s)-\mathbf{b}(y(s))\big)^{\top}\mathbf{D}(y(s))^{-1}\big(\dot{y}(s)-\mathbf{b}(y(s))\big)\,ds
    \end{equation*}

\noindent where $\mathbf{b}$ and $\mathbf{D}$ are the effective drift and diffusion on the continuous coordinate sector $Y$. This principle reveals how collective 
computation balances the need to maintain stable processing states with the ability to transition between them.

The probability of collective transitions depends on both energetic and memory effects
    \begin{equation*}
        P_{\alpha\beta}(T) = \frac{\exp(-\Delta F_{\alpha\beta}(T)/(k_BT))}{\mathcal{Z}(T)}\left(1 + \gamma(T)\int_0^t K(t-s)M_{\alpha\beta}(s,T)ds\right)
    \end{equation*}

\noindent where $\Delta F_{\alpha\beta}$ is the free energy barrier between collective states and $M_{\alpha\beta}(s)$ captures the history of transitions through 
the memory kernel $K(t)$.

This collective attractor structure supports crucial computational capabilities, as indicated below.
\begin{itemize}
    \item Stable attractors maintain coherent processing states.
    \item Transition pathways enable flexible computation.
    \item Memory effects modulate transition probabilities.
    \item Temperature controls exploration-exploitation balance.
\end{itemize}

\noindent The relationship between attractor stability and transition dynamics defines the system's \textbf{computational regime}
\begin{equation*}
    \tau_{comp} = \tau_0\exp(\Delta F/(k_BT))\left(1 - \gamma\int_0^t K(t-s)M(s)ds\right)
\end{equation*}

\noindent where $\tau_{comp}$ is the characteristic computation time. This reveals how collective attractor dynamics naturally implement a form of metastable 
computation, balancing stability with adaptability.


\subsection{Coherent Structures}
\label{subsec:coherentStruct}

Coherent spatiotemporal patterns emerge as fundamental computational structures in {\cogent{COGENT$^{\textbf{3}}$}}. These patterns represent \textbf{coordinated activity across 
multiple agents}, enabling distributed information processing. Collective patterns take the form
    \begin{equation}
        \Phi_i(t) = \sum_k A_k(t)\,u_k(i;t) 
    \end{equation}

\noindent with $(u_k(\cdot;t))_{k=1}^{|V|}$ the eigenvectors of $L(t)$ on $V$, and where $A_k(t)$ represents the time-dependent amplitude of mode $k$. Building on the 
Lyapunov stability analysis developed in Section \ref{subsec:crossover}, the stability of these collective patterns manifests in the time evolution of their mode amplitudes.

The emergence of coherent structures depends on the interplay between local dynamics and collective organization, or
    \begin{equation*}
        \frac{d\Phi_i}{dt} = -(L(t)\Phi(t))_i + \int_0^t K(t-s)\Phi_i(s)\,ds + \eta_i(t)
    \end{equation*}

\noindent where $K(t)$ is the memory kernel and $\eta_i(t)$ represents node-local fluctuations. These coherent structures enable distributed computation through 
a number of mechanisms, being
\begin{itemize}
    \item synchronized activity across spatial regions;
    \item stable information propagation pathways;
    \item memory-dependent pattern modulation; and
    \item noise-resistant collective processing.
\end{itemize}

\noindent The computational capacity of coherent structures depends on their stability and extent, defined by
    \begin{equation}
        C_{coh}(T) = \sum_k |A_k(T)|^2\left(\sum_{i\in V}|u_k(i;t)|^2\right)\left(1 + \gamma(T)\int_0^t K(t-s)\,ds\right)
    \end{equation}

\noindent where larger $C_{coh}$ indicates greater potential for distributed computation. This capacity measure reveals how coherent structures naturally implement 
parallel processing through spatially extended collective patterns.

Concerning the relationship between pattern coherence and computational capability, it's modulated by temperature
    \begin{equation*}
        \xi_{coh}(T) = \sqrt{\frac{\langle\|\Phi(t)\|_2^2\rangle_T}{\langle\|\eta(t)\|_2^2\rangle_T}} \sim \exp(-E_a/\overline{T}(t))
    \end{equation*}

\noindent where $\xi_{coh}$ quantifies the ratio of coherent signal to noise and $E_a$ is an activation energy for pattern formation. This temperature dependence 
enables adaptive control of collective pattern formation and computational organization.


\subsection{Computational Complexity}
\label{subsec:compComplex}

{\cogent{COGENT$^{\textbf{3}}$}} computational complexity scales with the size of its constituent network, captured by the \textbf{resource function}
    \begin{equation}
        R(N) = \alpha N\log N + \beta N\sqrt{T} + \gamma N\exp(-\Delta/T)
    \end{equation}

\noindent where the interplay of logarithmic, square root, and exponential terms reflects the balance between resource allocation, thermal fluctuations, and 
stability in collective computation.

Resource scaling follows distinct regimes based on collective organization, that is
\begin{align*}
    \textbf{Time complexity} \rightarrow \; & T(|V|) \sim O(|V|\log |V|); \\[0.2em]
    \textbf{Space complexity} \rightarrow \; & S(|V|) \sim O(|V|); \, \text{and} \\
    \textbf{Energy complexity} \rightarrow \; & E(|V|) \sim O(|V|\sqrt{T}).
\end{align*}

\noindent The full time complexity incorporating memory effects becomes
    \begin{equation*}
        \tau(N) = \tau_0N^\alpha\log^\beta N\left(1 + \lambda M(N)\right)
    \end{equation*}

\noindent where $M(N)$ captures how collective memory requirements scale with system size. Therefore, space complexity with memory effects follows
    \begin{equation*}
        S(N) = cN + \gamma N\log N + \delta\sqrt{N}\left(1 + \mu\int_0^t K(t-s)ds\right)
    \end{equation*}

\noindent demonstrating how memory kernel effects modify basic scaling laws.

These complexity measures reveal the barrier to collective computation, given by
\begin{equation}
    P_{max}(N,T,E) = \min\left\{\frac{N}{\log N}, \frac{E}{T\log(2)}, \frac{T_c}{T}-1\right\}
\end{equation}

\noindent where $P_{max}$ represents maximum computational performance bounded by network size, energy constraints, and critical temperature effects.


\subsection{Measures of Emergence}
\label{subsec:emergence}

The complexity measure defined on $\Omega$ quantifies the emergence of collective behavior through
    \begin{equation}
        \mathcal{C}(T) = \sum_\alpha p_\alpha(T)\log\frac{p_\alpha(T)}{p_\alpha^{ind}(T)}
    \end{equation}

\noindent where $p_\alpha$ represents probabilities in the collective state space and $p_\alpha^{ind}$ represents independent behavior. High $\mathcal{C}$ values 
indicate greater integration and interdependence among agents, reflecting coordinated computation.

The integration measure across subspaces provides a complementary perspective

    \begin{equation*}
        \Phi(T) = H(X,T) - \sum_i H(X_i|X\setminus X_i,T)
    \end{equation*}

\noindent where $H$ is the entropy functional defined on $\Gamma$. This measure quantifies the entropy reduction achieved through cooperative interactions, 
\textbf{providing a direct link between distributed processing and emergent intelligence}.

These measures enable quantitative analysis of emergence through several mechanisms, outlined below.
\begin{itemize}
    \item Distinction between collective and independent behaviors.
    \item Quantification of information integration across scales.
    \item Measurement of computational coherence.
    \item Assessment of emergent processing capabilities.
\end{itemize}


\subsection{Pattern Formation Analysis}
\label{subsec:patternForm}

Building on the pattern formation framework established in Section \label{subsec:coreOp}, we analyze how \textbf{collective patterns emerge from coordinated multi-agent dynamics}. 
While individual pattern formation follows local stability criteria, collective patterns exhibit distinct emergent properties arising from agent synchronization and 
large-scale coordination.

The collective pattern field combines contributions from synchronized agent groups, defined by
    \begin{equation}
        \Phi_i(t) = \sum_{\alpha} w_\alpha(t)\,\phi_{\alpha,i}(t)\left(1 + \gamma\int_0^t K(t-s)M_\alpha(s)\,ds\right)
    \end{equation}

\noindent where $w_\alpha(t)$ represents the time-dependent weight of pattern $\alpha$, and the memory term captures how historical collective states influence 
current pattern formation.

\noindent The synchronization of pattern formation across agents enables a number of \textbf{emergent computational capabilities}.
\begin{itemize}
    \item Distributed pattern recognition through collective state alignment.
    \item Robust information encoding in synchronized agent groups.
    \item Memory-dependent pattern modulation at collective scale.
    \item Adaptive reconfiguration of collective computational structures.
\end{itemize}

The collective pattern formation efficiency depends on inter-agent coordination
    \begin{equation}
        \eta_{coll}(T) = \frac{\|\Phi(T)\|_2^2}{\sum_i \|\phi_i(T)\|^2} \leq 1
    \end{equation}

\noindent where equality is achieved under perfect agent synchronization. This efficiency measure quantifies how effectively individual pattern formation processes 
combine into coherent collective computation.


\subsection{Knowledge Validation Dynamics}
\label{subsec:knowValidation}

The validation of knowledge states emerges through the \textbf{interplay of formation dynamics and collective states}. When the system processes external queries, 
knowledge propagates with an effective energy reflecting its collective validation, as
\begin{equation}
    E_{eff}^{val}(i,j) = \Delta E_{eff}(i,j) + \beta \sum_{k \in \mathcal{N}_{ij}} W(\mathcal{K}_i, \mathcal{K}_j, \mathcal{K}_k)
\end{equation}

\noindent where $\Delta E_{eff}(i,j)$ is the effective energy defined in Section \ref{subsec:expLagrangian}, $W(\mathcal{K}_i, \mathcal{K}_j, \mathcal{K}_k)$ measures knowledge consistency 
in triads, $\beta$ controls validation sensitivity, and $\mathcal{N}_{ij}$ represents the shared neighborhood of nodes $i$ and $j$.

The validation term $W$ emerges from triad interactions, that is
\begin{equation*}
    W(\mathcal{K}_i, \mathcal{K}_j, \mathcal{K}_k;t) \coloneqq \frac{1}{3}\big(\Phi_i(t)+\Phi_j(t)+\Phi_k(t)\big)
\end{equation*}

\noindent where $\Phi_{coll}$ is the collective pattern field.

This structure enables {\cogent{COGENT$^{\textbf{3}}$}} to \textbf{answer external queries through collective validation}. The effective validation energy modifies 
state transitions as
\begin{equation}
    P(\mathcal{K}_i \to \mathcal{K}_j|q) = \frac{\exp(-E_{eff}^{val}(i,j)/(k_BT))}{\mathcal{Z}(T)}
\end{equation}

\noindent where $q$ represents an external query and $T$ is the local temperature. The dynamics now incorporate \textbf{three coupled processes}, which provide
\begin{itemize}
    \item pattern formation within triads;
    \item memory-mediated stability through $K(t)$; and
    \item knowledge validation through $E_{eff}^{val}$.
\end{itemize}

\noindent These processes interact through the node temperatures $(T_i(t))_{i\in V}$ and memory kernel $K(t)$ to
\begin{itemize}
    \item maintain multiple valid knowledge states;
    \item propagate knowledge based on collective validation;
    \item adapt responses to external queries; and
    \item preserve the stability of well-validated patterns.
\end{itemize}

The validation dynamics integrate seamlessly with our framework through several avenues, outlined below.
\begin{enumerate}
    \item \textbf{Pattern Fields:} The validation term $W$ uses the collective pattern field structure established in Section \ref{subsec:attractorDyn}.
    \item \textbf{Memory Effects:} The effective energy $E_{eff}^{val}$ builds on the memory-modified energy landscape.
    \item \textbf{Temperature Dependence:} Knowledge validation remains subject to temperature modulation.
    \item \textbf{Critical Behavior:} The validation energy contributes to phase transitions consistently with Section \ref{sec:phase}.
\end{enumerate}

\noindent Consequently, knowledge validation emerges as a \textbf{natural consequence of the system's fundamental dynamics}.


\subsection{Memory Capacity Analysis}
\label{subsec:memCapacity}

Building on the memory embedding framework established in Section \ref{subsec:memEmbed}, we analyze the theoretical bounds and scaling properties of collective memory capacity. 
For $P$ stored patterns, the maximum capacity follows
    \begin{equation*}
        P_{max} = \frac{\alpha_c N}{\log N}\left(1 + \gamma\int_0^t K(t-s)M(s)ds\right)
    \end{equation*}

\noindent where $\alpha_c$ is the critical storage ratio and the memory kernel term captures how historical context modifies basic capacity scaling.

The signal-to-noise ratio in pattern retrieval becomes
    \begin{equation*}
        \text{SNR}(T) = \frac{\langle m_\mu^2 \rangle_T}{\sum_{\nu\neq\mu}\langle m_\nu^2 \rangle_T}
    \end{equation*}

\noindent leading to a fundamental capacity-reliability trade-off
    \begin{equation}
        P_{eff}(T) = P(T)\cdot q(P,T) \leq P_{max}\exp(-\beta/T)
    \end{equation}

\noindent This trade-off between memory capacity and reliability highlights {\cogent{COGENT$^{\textbf{3}}$}} intrinsic limits in managing noisy environments while 
maintaining accurate retrieval. The temperature dependence enables dynamic control of this trade-off, providing a mechanism for adaptive memory management in 
collective computation.


\subsection{Computational Trade-offs}
\label{subsec:tradeoffs}

{\cogent{COGENT$^{\textbf{3}}$}} performance hinges on fundamental trade-offs between \textbf{processing speed, accuracy, energy consumption, and parallelization}. 
These trade-offs manifest distinctly in \textbf{collective computation}, where multiple agents must coordinate their processing.

\subsubsection{Speed-Accuracy Trade-off}
Processing time versus error rate in collective computation follows
    \begin{equation}
        \tau_{proc} \cdot \epsilon = c_{proc}\left(1 + \lambda\int_0^t K(t-s)ds\right)
    \end{equation}

\noindent where $c_{proc}$ is an effective (model-dependent) speed-accuracy constant.

\subsubsection{Energy-Precision Trade-off}
Energy \textbf{cost per collective operation} becomes
    \begin{equation*}
        E_{op}(T) = T\log(2)\left(\frac{1}{\epsilon(T)} + \gamma(T)\sum_{\{i,j\}\in E^{(2)}(t)} w_{ij}(t)\,(\phi_i(t)-\phi_j(t))^2\right)
    \end{equation*}

\noindent leading to the fundamental relation
    \begin{equation}
        E_{op}(T)\tau_{proc}(T) \geq T\log(2)c_{proc}
    \end{equation}

\subsubsection{Parallel Processing Efficiency}
For $M$ parallel processes at temperature $T$, collective efficiency scales as
    \begin{equation*}
        \eta_{parallel}(T) = \frac{\text{throughput}(M,T)}{\text{throughput}(1,T)} \leq M^{\alpha(T)}
    \end{equation*}

\noindent where $\alpha(T) < 1$ reflects temperature-dependent communication overhead
    \begin{equation*}
        \alpha(T) = 1 - \beta(T)\frac{\log\log M}{\log M}
    \end{equation*}

\noindent Therefore, \textbf{total computational efficiency} becomes
    \begin{equation}
        \eta_{total}(T) = \eta_{parallel}(T)\cdot\eta_{memory}(T)\cdot\eta_{thermal}(T)
    \end{equation}

\noindent where component efficiencies satisfy
    \begin{align*}
        \eta_{memory}(T) &\leq \exp(-P(T)/P_{max}(T)) \\
        \eta_{thermal}(T) &\leq 1 - T/T_c
    \end{align*}


\section{Stability Analysis}
\label{sec:stability}

Ensuring stability is \textbf{crucial for any system to perform reliable computation and cognitive processing}. Unlike traditional computational architectures with 
fixed stability properties, {\cogent{COGENT$^{\textbf{3}}$}} exhibits rich stability characteristics that emerge from the interplay of pattern formation, group dynamics, 
and memory effects. This analysis provides a comprehensive understanding of how the system maintains computational reliability while preserving adaptability.

\subsection{Complete Stability Matrix}
\label{subsec:stabMatrix}

The stability of computational processes in {\cogent{COGENT$^{\textbf{3}}$}} requires analysis of interactions across all system components. 

\newpage

\noindent The complete stability matrix on $\Gamma$, defined as 

    \begin{equation}
        \mathbf{J} = \begin{pmatrix}
            \dfrac{\partial \dot{\phi}}{\partial \phi} & \dfrac{\partial \dot{\phi}}{\partial r} & \dfrac{\partial \dot{\phi}}{\partial T} & \dfrac{\partial \dot{\phi}}{\partial M} \\
            \dfrac{\partial \dot{r}}{\partial \phi} & \dfrac{\partial \dot{r}}{\partial r} & \dfrac{\partial \dot{r}}{\partial T} & \dfrac{\partial \dot{r}}{\partial M} \\
            \dfrac{\partial \dot{T}}{\partial \phi} & \dfrac{\partial \dot{T}}{\partial r} & \dfrac{\partial \dot{T}}{\partial T} & \dfrac{\partial \dot{T}}{\partial M} \\
            \dfrac{\partial \dot{M}}{\partial \phi} & \dfrac{\partial \dot{M}}{\partial r} & \dfrac{\partial \dot{M}}{\partial T} & \dfrac{\partial \dot{M}}{\partial M}
        \end{pmatrix}
    \end{equation}

\noindent provides a structured decomposition revealing how different aspects of the system influence each other, mirroring key elements of cognitive and 
computational systems.

\begin{itemize}
    \item \textbf{Pattern Formation ($\phi$)} captures the emergence of coherent computational structures from local interactions, analogous to neural activation 
    patterns.
    \item \textbf{Role Dynamics ($r$)} represents functional specialization, akin to modular roles in multi-agent systems or neural circuits.
    \item \textbf{Temperature Effects ($T$)} models energy or resource constraints, critical for balancing processing speed and accuracy.
    \item \textbf{Memory ($M$)} encodes the influence of past states on current stability, providing a mechanism for incorporating learning and historical context.
\end{itemize}

\noindent The above decomposition provides a structured approach to analyzing how these components interact to maintain computational stability while allowing for adaptive 
flexibility.

\subsubsection{Memory Operator Stability}
The memory operator $\mathcal{M}$ must satisfy explicit stability bounds, given by
    \begin{equation}
        \|\mathcal{M}\| = \left\|\int_0^t K(t-s)\hat{\mathfrak{L}}(s)ds\right\| \leq \gamma_c
    \end{equation}

\noindent where the critical coupling $\gamma_c$ is determined by
    \begin{equation*}
        \gamma_c = \sup\{\gamma: \text{Re}(\lambda_i(\hat{\mathfrak{L}} + \gamma\mathcal{M})) < 0 \quad \forall i\}
    \end{equation*}

\noindent and memory stability conditions
    \begin{align*}
        \textbf{Spectral bound} \rightarrow \; &\|\mathcal{M}\|_{sp} < 1/\|\hat{\mathfrak{L}}^{-1}\|; \\[0.2em]
        \textbf{Trace condition} \rightarrow \; &\text{tr}(\mathcal{M}) < -\text{tr}(\hat{\mathfrak{L}}); \, \text{and} \\[0.2em]
        \textbf{Lyapunov bound} \rightarrow \; &\mathcal{M}^T\mathbf{P} + \mathbf{P}\mathcal{M} < 0.
    \end{align*}

The memory operator bounds represent \textbf{fundamental limits on information retention} while maintaining system stability. These bounds have direct computational 
implications: if $\|\mathcal{M}\|$ exceeds $\gamma_c$, the system loses its ability to maintain stable memory states, compromising computational reliability.


\subsection{Local Stability Analysis}
\label{subsec:localStab}

While global stability ensures overall system coherence, local stability analysis reveals \textbf{how individual computational elements maintain reliable processing in 
the presence of perturbations}.

\noindent The linearized dynamics near fixed points 
    \begin{equation}
        \frac{d}{dt}\delta\mathbf{x} = \mathbf{J}(\mathbf{x}^*)\delta\mathbf{x}
    \end{equation}

\noindent define the system's response to small disturbances, with characteristic equation
    \begin{equation*}
        \det(\mathbf{J} - \lambda\mathbf{I}) = \lambda^4 + a_1\lambda^3 + a_2\lambda^2 + a_3\lambda + a_4 = 0
    \end{equation*}

\subsubsection{Coupled Stability}
The coupled system stability matrix
    \begin{equation}
        \mathbf{J}_{coupled} = \begin{pmatrix}
        \mathbf{J}_{form} & \mathbf{J}_{GV} & \mathbf{J}_{GM} \\[0.4em]
        \mathbf{J}_{VG} & \mathbf{J}_{group} & \mathbf{J}_{VM} \\[0.4em]
        \mathbf{J}_{MG} & \mathbf{J}_{MV} & \mathbf{J}_{mem}
        \end{pmatrix}
    \end{equation}

\noindent shows how \textbf{different computational processes interact to maintain stability}. Pattern formation (\textit{form}), group dynamics (\textit{group}), 
and memory processes (\textit{mem}) \textbf{must satisfy collective stability conditions for reliable computation}.

\noindent Stability requires all eigenvalues $\lambda$ to satisfy
    \begin{equation*}
        \text{Re}(\lambda_i(\mathbf{J}_{coupled})) < 0
    \end{equation*}

\noindent and the coupling matrices to satisfy
    \begin{align*}
        \|\mathbf{J}_{GV}\| &\leq \alpha_{GV}\|\mathbf{J}_{form}\| \\
        \|\mathbf{J}_{GM}\| &\leq \alpha_{GM}\|\mathbf{J}_{form}\| \\
        \|\mathbf{J}_{VM}\| &\leq \alpha_{VM}\|\mathbf{J}_{group}\|
    \end{align*}

\noindent Here,
\begin{itemize}
    \item \(\mathbf{J}_{GV}\): Reflects how group dynamics influence the stability of emergent patterns, akin to how collaborative problem-solving stabilizes 
    group performance.
    \item \(\mathbf{J}_{GM}\): Captures the influence of memory on pattern formation, enabling the recall of learned structures during computation.
\end{itemize}
\noindent These interactions mirror the interdependence of cognitive processes, where memory, perception, and decision-making dynamically reinforce one another 
to achieve robust performance.
    

\subsection{Stability Conditions}
\label{subsec:stabCond}

Global stability requires
    \begin{enumerate}
        \item $\text{Re}(\lambda_i) < 0$ for all eigenvalues $\lambda_i$ of $\mathbf{J}$;
        \item $\|\mathbf{M}\| < 1$ for memory operator $\mathbf{M}$; and
        \item $T_{min} < T < T_{max}$ for all local temperatures,
    \end{enumerate}
\noindent with Routh-Hurwitz conditions
    \begin{align*}
        a_1 &> 0 \\
        a_1a_2 - a_3 &> 0 \\
        a_3(a_1a_2 - a_3) - a_1^2a_4 &> 0 \\
        a_4 &> 0
    \end{align*}


\subsection{Structural Stability}
\label{subsec:structStab}

Beyond instantaneous stability, {\cogent{COGENT$^{\textbf{3}}$}} must be also able to \textbf{maintain computational reliability under structural perturbations}. 

This robustness is essential for cognitive-like processing in noisy environments. The perturbation analysis becomes
    \begin{equation}
        \mathbf{J}_\epsilon = \mathbf{J}_0 + \epsilon\delta\mathbf{J}
    \end{equation}

\noindent and the stability margin
    \begin{equation*}
        \Delta = \min_{\|\delta\mathbf{J}\|=1}\{\epsilon: \mathbf{J}_\epsilon \text{ unstable}\}
    \end{equation*}

\subsubsection{Robust Stability Analysis}
The robust stability margin is defined as
    \begin{equation}
        \mu_{\Delta}(\mathbf{J}) = \left(\min_{\|\Delta\| \leq 1}\{\sigma: \det(I - \sigma\mathbf{J}\Delta) = 0\}\right)^{-1}
    \end{equation}

\noindent quantifying the \textbf{tolerance to uncertainties} in different computational components. This structured approach allows us to separately consider 
uncertainties in pattern formation ($\Delta_G$), group dynamics ($\Delta_V$), and memory processes ($\Delta_M$), reflecting how \textbf{different aspects of computation} 
may be subject to \textbf{different types of perturbation}.

\noindent Uncertainties in our system can be decomposed into a block-diagonal structure 
    \begin{equation*}
        \Delta = \begin{pmatrix}
        \Delta_G & 0 & 0 \\
        0 & \Delta_V & 0 \\
        0 & 0 & \Delta_M
        \end{pmatrix} \quad \|\Delta_i\| \leq 1
    \end{equation*}

\noindent where \textbf{pattern formation uncertainties} ($\Delta_G$) capture variations in learning and adaptation processes, \textbf{group dynamic uncertainties} 
($\Delta_V$) represent variations in collective behavior, while \textbf{memory uncertainties} ($\Delta_M$) model fluctuations in information storage and retrieval. 

The bound $\|\Delta_i\| \leq 1$ normalizes these uncertainties to a common scale while preserving their relative independence.

Robust stability, in turn, requires that
    \begin{equation*}
        \mu_{\Delta}(\mathbf{J}) < 1
    \end{equation*}

\noindent When this condition is satisfied, {\cogent{COGENT$^{\textbf{3}}$}} \textbf{maintains computational integrity despite worst-case combinations of uncertainties} in its different 
components. This provides a crucial guarantee for reliable cognitive-like processing in real-world implementations where perfect modeling is impossible.

Structured singular value bounds
    \begin{equation*}
        \max_{\omega}\rho(\mathbf{M}(j\omega)) \leq \mu_{\Delta}(\mathbf{J}) \leq \min_{D \in \mathcal{D}}\|\mathbf{D}\mathbf{J}\mathbf{D}^{-1}\|
    \end{equation*}

\noindent provide computationally tractable ways to verify robust stability. The lower bound, based on the spectral radius $\rho(\mathbf{M}(j\omega))$, gives a 
necessary condition for stability, while the upper bound involving scaling matrices $D$ provides a sufficient condition. The gap between these bounds represents the 
trade-off between computational efficiency and conservative stability guarantees.

Margin optimization
    \begin{equation*}
        \max_{\theta \in \Theta} \min_{\omega} \sigma_{min}(I + \mathbf{G}(j\omega,\theta))
    \end{equation*}

\noindent balances maximum robustness (through $\theta$ parameters) against worst-case frequency response (minimization over $\omega$). This optimization reflects 
the practical challenge of designing systems that are both robust and computationally efficient. It is subject to a nominal stability constraint
    \begin{equation*}
        \text{Re}(\lambda_i(\mathbf{J}(\theta))) < 0 \quad \forall i
    \end{equation*}

\noindent ensuring that optimizing for robustness doesn't compromise the system's basic computational functionality. Together, these conditions provide a 
comprehensive framework for designing a reliable computational system that maintains performance even under uncertainty.


\subsection{Memory-Induced Stability}
\label{subsec:memIndStab}

Memory effects in {\cogent{COGENT$^{\textbf{3}}$}} don't just store information: \textbf{they actively contribute to computational stability}. The memory stability 
operator captures how past states influence current computational processes.

Define the memory stability operator
    \begin{equation}
        \mathcal{S}_{mem}(t) = \exp\left(\int_0^t K(t-s)\mathbf{J}(s)\,ds\right)
    \end{equation}

\noindent with memory stability condition
    \begin{equation*}
        \|\mathcal{S}_{mem}(t)\| < 1 \quad \forall t > t^*
    \end{equation*}

\noindent This stability condition ensures that memory effects enhance, rather than disrupt, ongoing computation. When $\|\mathcal{S}(t)\| < 1$, memory reinforces 
stable computational states while still allowing for adaptive changes when needed.

Memory effects are particularly critical in cognitive systems, where past experiences influence decision-making and adaptability. For example,
\begin{itemize}
    \item stable memory ensures the \textbf{persistence of learned patterns}, preventing computational drift during complex tasks; and
    \item properly bounded memory effects \textbf{allow the system to remain adaptable}, enabling the incorporation of new information without compromising stability.
\end{itemize}

\noindent This dual role of memory (enhancing stability while preserving adaptability) is a hallmark of cognitive processes such as learning, where past knowledge 
informs future behavior.


\subsection{Temperature-Role Coupling}
\label{subsec:tempCoupling}

The interaction between temperature and role dynamics is crucial for \textbf{balancing computational stability} with \textbf{adaptability}. 

The coupled stability matrix
    \begin{equation}
        \mathbf{J}_{TR} = \begin{pmatrix}
            \dfrac{\partial \dot{T}}{\partial T} & \dfrac{\partial \dot{T}}{\partial r} \\[1em]
            \dfrac{\partial \dot{r}}{\partial T} & \dfrac{\partial \dot{r}}{\partial r}
        \end{pmatrix}
    \end{equation}

\noindent describes how temperature changes affect role assignments, and vice versa, with stability criterion defined by
    \begin{equation*}
        \text{tr}(\mathbf{J}_{TR}) < 0 \quad \text{and} \quad \det(\mathbf{J}_{TR}) > 0
    \end{equation*}

\noindent These conditions ensure that \textbf{role transitions occur smoothly} without disrupting ongoing computations. When $\text{tr}(\mathbf{J}_{TR}) < 0$ and 
$\det(\mathbf{J}_{TR}) > 0$ the system maintains stable functional organization while allowing for temperature-mediated role adaptation.

The ability of {\cogent{COGENT$^{\textbf{3}}$}} to adapt its functional organization to environmental demands is a result of temperature-role coupling. 
\begin{itemize}
    \item Higher temperatures promote \textbf{exploration}, allowing dynamic role-switching and adaptation to novel tasks.
    \item Lower temperatures \textbf{stabilize functional roles}, ensuring efficient execution of well-learned tasks.
\end{itemize}

As noted earlier, this balance between exploration and exploitation mirrors (in a machine computation setup) the adaptive strategies of biological systems, where 
cognitive flexibility is essential for learning and problem-solving.


\subsection{Lyapunov Analysis}
\label{subsec:lyapunovAna}

Lyapunov stability provides a rigorous framework for understanding how {\cogent{COGENT$^{\textbf{3}}$}} maintains computational coherence over time. The Lyapunov 
function candidate
    \begin{equation}
        V(\mathbf{x}) = \mathbf{x}^T\mathbf{P}\mathbf{x} + \int_0^t K(t-s)\|\mathbf{x}(s)\|^2ds
    \end{equation}

\noindent incorporates both instantaneous state information and memory effects. The stability condition
    \begin{equation*}
        \dot{V}(\mathbf{x}) = \mathbf{x}^T(\mathbf{J}^T\mathbf{P} + \mathbf{P}\mathbf{J})\mathbf{x} + \|\mathbf{x}\|^2 - \int_0^t K(t-s)\|\mathbf{x}(s)\|^2ds < 0
    \end{equation*}

\noindent ensures that computational state converges to stable configurations. The memory integral term captures how historical information influences 
this convergence, reflecting {\cogent{COGENT$^{\textbf{3}}$}} ability to learn from past states while maintaining stability.


\subsection{Orbital Stability}
\label{subsec:orbitalStab}

Beyond fixed point stability, {\cogent{COGENT$^{\textbf{3}}$}} exhibits \textbf{periodic computational states} whose stability is characterized through 
\textbf{orbital analysis}. 

These periodic structures are essential to both cognitive and computational systems, enabling rhythmic processes like attention cycles 
and synchronized information processing (see Section \ref{subsec:univClass}).
\begin{itemize}
    \item In neural systems, rhythmic oscillations enable \textbf{attention cycles}, ensuring periodic engagement with relevant stimuli.
    \item Periodic computational patterns support \textbf{time-sensitive tasks}, such as synchronized data processing in distributed systems.
\end{itemize}

\noindent For periodic solutions, we define
    \begin{equation}
        \frac{d}{dt}\delta\mathbf{x}_\perp = (\mathbf{I} - \hat{\mathbf{v}}\hat{\mathbf{v}}^T)\mathbf{J}\delta\mathbf{x}_\perp
    \end{equation}

\noindent with Floquet multipliers
    \begin{equation*}
        \det(\mathbf{M}(T) - \mu\mathbf{I}) = 0
    \end{equation*}

The Floquet multipliers determine whether periodic computational patterns remain stable under perturbations. In that way, orbital stability guarantees the 
reliability of rhythmic processes, critical for artificial cognitive systems.\footnote{In a similar way as in biological systems.}


\subsection{Network Stability}
\label{subsec:networkStab}

The distributed nature of {\cogent{COGENT$^{\textbf{3}}$}} requires careful analysis of network-wide stability properties. The master stability function provides a 
unified approach to analyzing stability across the network
    \begin{equation}
        \Lambda(\alpha) = \max_{\lambda\in\text{spec}(\mathbf{L})}\text{Re}(\lambda_\text{max}(\mathbf{J} + \alpha\mathbf{L}))
    \end{equation}

\noindent with synchronization condition
    \begin{equation*}
        \Lambda(\alpha) < 0 \quad \forall \alpha \in \text{spec}(\mathbf{L})
    \end{equation*}

\noindent ensuring that distributed computational processes remain coordinated across the network. 

\noindent As a result, the master stability function
\begin{itemize}
    \item ensures that \textbf{local computations remain synchronized with global objectives}, similar to how neural networks achieve coherent cognitive states; and
    \item allows the system to \textbf{respond dynamically to changing inputs} while maintaining overall coordination, critical for tasks such as large-scale 
    decision-making.
\end{itemize}

\noindent This ability to balance local adaptability with global coherence is a key feature of both biological and artificial intelligence.


\subsection{Robust Stability}
\label{subsec:robustStab}
Practical implementation of our framework requires stability guarantees that hold under uncertainty and perturbation. 
The H-infinity norm bound
    \begin{equation}
        \|\mathcal{T}_{zw}\|_\infty = \sup_{\omega}\bar{\sigma}(\mathcal{T}_{zw}(j\omega)) < \gamma
    \end{equation}

\noindent quantifies the system's robustness, with stability margin
    \begin{equation*}
        \mu_{\Delta}(\mathbf{J}) < 1
    \end{equation*}
\noindent 

These robust stability conditions ensure reliable computation even in the presence of modeling uncertainties and external disturbances. The stability margin 
$\mu_{\Delta}(\mathbf{J}) < 1$ provides a quantitative measure of how much uncertainty the system can tolerate while maintaining computational integrity.


\section{Conclusions}
\label{sec:conclusions}

{\cogent{COGENT$^{\textbf{3}}$}} offers a principled path toward artificial systems that better captures the flexibility and context-sensitivity of human cognition. By 
integrating pattern formation with thermodynamic principles, we demonstrate how stable computational elements can dynamically reorganize to support adaptive processing 
across multiple scales.

Some key features of {\cogent{COGENT$^{\textbf{3}}$}} are
\begin{itemize}
    \item a mathematically rigorous framework for \textbf{emergent cognitive structures} through dynamic three-agent interactions;
    \item \textbf{temperature-modulated transitions} between exploratory and exploitative processing regimes;
    \newpage
    \item \textbf{integration of memory effects} that enable context-dependent adaptation while maintaining computational stability; and
    \item scale-bridging mechanisms that \textbf{connect local pattern formation to global computational capabilities}.
\end{itemize}

\noindent The strength of our approach lies in its ability to naturally express cognitive phenomena, a quest that, to date, has been a challenge for traditional 
architectures. Just as human experts can fluidly reconfigure basic knowledge elements to solve novel problems, {\cogent{COGENT$^{\textbf{3}}$}} enables computational 
structures to emerge and dissolve based on task demands. 

By resorting to temperature-dependent dynamics, our system provides a natural mechanism for balancing the exploration of new possibilities with the exploitation of 
established patterns: \textbf{a fundamental feature of adaptive intelligence}.

Perhaps more crucially, {\cogent{COGENT$^{\textbf{3}}$}} bridges theory with practical utility. The mathematical structures developed here (from pattern 
formation fields to memory kernels) have \textbf{clear physical interpretations} and \textbf{computational implementations}. It grounds abstract principles in 
concrete mechanisms while maintaining the flexibility needed for cognitive-like processing.

Looking ahead, {\cogent{COGENT$^{\textbf{3}}$}} opens new directions for developing artificial systems that can engage in many tasks elusive to current models. In 
general terms, we are thinking of artificial computational systems that can perform some of (or all) the tasks outlined below.
\begin{itemize}
    \item Adapt their processing strategies based on context and task demands.
    \item Form and dissolve computational structures dynamically.
    \item Balance stability with flexibility through temperature modulation.
    \item Learn from experience while maintaining reliable operation.
\end{itemize}

\noindent These capabilities move us \textbf{a bit closer to artificial systems} that can \textbf{engage with the world as fluidly and adaptively as biological 
intelligence}. By grounding emergent cognition in rigorous physical principles, we provide both a theoretical framework for understanding adaptive intelligence 
and practical tools for implementing it.

\newpage

The road ahead involves extending these principles to large-scale (and possibly distributed) systems, developing efficient implementations, and exploring the full 
range of cognitive capabilities {\cogent{COGENT$^{\textbf{3}}$}} can support. We hope it represents a significant step toward artificial systems that can be a small 
step closer to the flexibility, adaptability, and robustness of biological cognition.

\vspace*{14cm}
\noindent{\cogent{In memory to an exceptional colleague and scholar, Oscar Enrique Cornblit (1927-2010). I am grateful for our friendship, the countless conversations 
that influenced much of the work leading to this paper, and our equally spirited debates about opera.}}



\end{document}